\PassOptionsToPackage{table,dvipsnames}{xcolor}
\documentclass[a4paper,fleqn]{cas-sc}

\usepackage[numbers,sort&compress]{natbib}
\setcitestyle{numbers,square,comma,sort&compress}
\usepackage{graphicx}
\usepackage{dsfont}
\usepackage{subfig}
\usepackage{booktabs}
\usepackage{tabularx}
\usepackage{multirow}
\usepackage{makecell}
\usepackage{array}
\usepackage{threeparttable}
\usepackage{stfloats}
\usepackage{amssymb}
\usepackage{amsmath}
\usepackage{siunitx}
\usepackage{lineno}
\usepackage{longtable}
\usepackage{ragged2e}

\usepackage{xcolor}
\usepackage{pifont}
\usepackage{algorithm}
\usepackage{algpseudocode}

\definecolor{ElsevierBlue}{HTML}{00AEEF}

\definecolor{boxaccent}{RGB}{84,100,123}
\definecolor{maskaccent}{RGB}{96,119,105}
\definecolor{contaccent}{RGB}{128,98,76}

\definecolor{bestbg}{RGB}{217,231,249}      
\definecolor{bestfg}{RGB}{24,56,92}
\definecolor{secondbg}{RGB}{236,240,244}
\definecolor{secondfg}{RGB}{88,96,108}
\definecolor{oursbg}{RGB}{236,244,234}
\definecolor{summarygray}{RGB}{245,245,245}
\definecolor{tableblue}{RGB}{232,240,255}
\definecolor{gaincolor}{RGB}{0,110,0}

\definecolor{shapebestbg}{RGB}{246,223,190}
\definecolor{shapebestfg}{RGB}{115,68,20}

\definecolor{geombestbg}{RGB}{235,231,244}
\definecolor{geombestfg}{RGB}{86,74,120}

\definecolor{SoftBlue}{RGB}{50, 90, 140}
\definecolor{SoftRed}{RGB}{180, 90, 80}
\definecolor{SoftGray}{RGB}{113, 128, 150}

\newcommand{\shapebestnum}[1]{%
  {\setlength{\fboxsep}{0.20pt}%
   \colorbox{shapebestbg}{\textcolor{shapebestfg}{\strut\textbf{#1}}}}%
}

\newcommand{\geombestnum}[1]{%
  {\setlength{\fboxsep}{0.20pt}%
   \colorbox{geombestbg}{\textcolor{geombestfg}{\strut\textbf{#1}}}}%
}

\newcommand{\na}{--}

\algrenewcommand\algorithmiccomment[1]{\hfill\textcolor{SoftGray}{// #1}}

\newcommand{\abyes}{\raisebox{0.10ex}{\scalebox{0.92}{\textcolor{ForestGreen!70!black}{\ding{51}}}}}
\newcommand{\abno}{\raisebox{0.10ex}{\scalebox{0.92}{\textcolor{BrickRed!80!black}{\ding{55}}}}}

\newcommand{\abbest}[1]{\textbf{#1}}

\newcommand{\bestnum}[1]{%
  {\setlength{\fboxsep}{0.20pt}%
   \colorbox{bestbg}{\textcolor{bestfg}{\strut\textbf{#1}}}}%
}

\newcommand{\routebox}{\raisebox{0.15ex}{\textcolor{boxaccent}{\rule{0.78ex}{0.78ex}}}}
\newcommand{\routemask}{\raisebox{0.15ex}{\textcolor{maskaccent}{\rule{0.78ex}{0.78ex}}}}
\newcommand{\routecont}{\raisebox{0.15ex}{\textcolor{contaccent}{\rule{0.78ex}{0.78ex}}}}

\AtBeginDocument{%
  \hypersetup{
    colorlinks=true,
    linkcolor=ElsevierBlue,
    citecolor=ElsevierBlue,
    urlcolor=ElsevierBlue,
  }%
}

\begin{document}

\shortauthors{J.Liu \& W.Wang et~al.}
\shorttitle{Frequency-supervised Fourier Series Detection (FS-FSD)}

\title [mode = title]{Contour-Native Bridge Defect Detection and Compact Digital Archiving with Frequency-Supervised Fourier Contours}                      

\author[1]{Jin Liu}
\author[2]{Wang Wang}
\ead{wangw00821@gmail.com}
\cormark[1]
\author[2]{Hongxu Pu}
\author[1]{Zhen Cao}
\author[3]{Yasong Wang}
\author[4]{Hu Wang}
\author[5]{Kunming Luo}

\affiliation[1]{organization={State Key Laboratory of Information Engineering in Surveying, Mapping and Remote Sensing, Wuhan University},
                city={Wuhan},
                postcode={430072}, 
                country={China}}
                
\affiliation[2]{organization={Sustainability X-Lab, The University of Hong Kong},
                city={Hong Kong},
                country={China}}
                
\affiliation[3]{organization={Department of Cyber Security, Southeast University},
                addressline={No. 2, Southeast University Road}, 
                city={Nanjing},
                postcode={211102}, 
                country={China}}

\affiliation[4]{organization={School of Computer Science and Engineering, University of Electronic Science and Technology of China},
                addressline={No. 2006, Xiyuan Avenue, West Hi-Tech Zone}, 
                city={Chengdu},
                postcode={611731}, 
                country={China}}

\affiliation[5]{organization={Department of Electronic and Computer Engineering, The Hong Kong University of Science and Technology},
                addressline={Clear Water Bay, Kowloon},
                city={Hong Kong},
                country={China}}

\begin{abstract}
AI-assisted bridge defect inspection often produces bounding boxes with crude geometry or raster masks that are costly to store, transmit, and reuse.
This study investigates how detected defects can be represented as compact, recoverable contour-level vector records in image space.
We propose Frequency-Supervised Fourier Series Detection (FS-FSD), which directly regresses Fourier contour descriptors and evaluates boxes, masks, and contours under a unified polygon-space protocol.
On 3,767 UAV-collected bridge images with 42,346 defect instances, FS-FSD achieves higher polygon-space accuracy and better matched-TP geometric quality than representative detection, segmentation, and contour baselines.
These results show that, compared with bounding boxes and raster masks, Fourier contour records preserve defect-boundary geometry in a more compact, recoverable, and shareable form for engineering review and downstream information workflows.
Future work will study the modeling of multi-region, fragmented, and adjacent bridge-defect boundaries and extend the framework toward long-term bridge-defect tracking and lifecycle-oriented management.
\end{abstract}


\maketitle

\section{Introduction}

Bridges are critical components of transportation infrastructure, and their safety and serviceability directly affect traffic efficiency, economic activity, and public safety \cite{moghtadernejad2025digitalizing,dorafshan2018bridge}. As bridge networks continue to age, structural health monitoring and periodic condition assessment have become increasingly important in both research and engineering practice \cite{dorafshan2018bridge,duque2018synthesis,lyu2025uav}. In recent years, unmanned aerial vehicles (UAVs) and deep-learning-based vision systems have substantially improved the efficiency of bridge inspection, making the automated recognition of common visible defect categories, such as cracks, spalling, seepage, efflorescence, and corrosion, increasingly feasible \cite{seo2018drone,duque2018synthesis,agnisarman2019humanloop,lyu2025uav,chen2026improving,li2025gyudet,mundt2019codebrim}. However, image-level defect detection is only an initial step in practical bridge maintenance. Greater engineering value lies in converting inspection outputs into geometrically explicit, traceable, and structured records that can be archived, queried, and reused in maintenance-oriented information workflows and longer-term asset assessment \cite{moghtadernejad2025digitalizing,huthwohl2018integrating,broo2021digitaltwins,fuller2020digitaltwin,kaveh2025advancing,wang2026unified}.

From this engineering perspective, a clear representational mismatch exists between mainstream computer-vision outputs and the data requirements of bridge inspection practice \cite{moghtadernejad2025digitalizing,huthwohl2018integrating,adhikari2014imagebridge}. Conventional object detectors typically represent targets using horizontal or oriented bounding boxes. Although effective for coarse localization, such representations remain structurally limited for bridge defects, whose annotated extents are often slender, fragmented, irregular, and arbitrarily shaped \cite{adhikari2014imagebridge,lyu2025uav,chen2026improving}. A bounding box inevitably introduces substantial background redundancy and spatial ambiguity, and it does not directly preserve faithful contour information or support the geometric attributes commonly required by downstream records, such as area, perimeter, centroid, orientation, and shape descriptors \cite{huthwohl2018integrating,tang2023novel}. Box-based detection is therefore more suitable for indicating where a defect may be located than for serving as a native representation for defect characterization, structured archival, or long-term temporal comparison \cite{moghtadernejad2025digitalizing,wang2026unified}.

Instance segmentation provides a more detailed alternative through dense pixel-wise raster masks, yet important limitations remain in bridge inspection scenarios \cite{cha2017deep,song2024pixel}. Bridge surfaces are frequently affected by uneven illumination, shadows, contamination, water stains, rough material textures, and complex background interference, all of which can degrade mask quality and produce fragmented regions or unstable boundaries \cite{agnisarman2019humanloop,lyu2025uav,chen2026improving}. More importantly, from the perspective of engineering informatics, a mask remains a rasterized geometric representation whose storage complexity scales with image resolution \cite{moghtadernejad2025digitalizing,huthwohl2018integrating}. When defect data must be managed across large bridge inventories and updated repeatedly in routine inspections, dense mask archival can impose a considerable burden on storage, transmission, and retrieval, particularly in edge-to-cloud inspection pipelines \cite{moghtadernejad2025digitalizing,broo2021digitaltwins,fuller2020digitaltwin,kaveh2025advancing,wang2026unified}. Moreover, raster masks are not naturally aligned with compact vectorized storage, geometry-oriented querying, or resolution-independent rendering in infrastructure information systems \cite{huthwohl2018integrating,wang2026unified}.

Existing studies have shown that contour-native Fourier parameterization provides a feasible pathway for modeling closed object boundaries \cite{zahn1972fourier,kuhl1982elliptic,peng2020deepsnake,liu2025fourierseries}. In particular, \textbf{Fourier Series Detection (FSD)} has demonstrated that contour-native prediction can serve as an alternative output paradigm to boxes and raster masks. Nevertheless, prior work has focused primarily on detection performance or contour regression feasibility, while an important question remains insufficiently examined in bridge defect inspection and archival workflows: when predicted defects are expected to enter databases, browser-side review environments, and broader asset-information workflows, what geometric form should serve as the native engineering record, and how can this representation remain learnable and stable in complex bridge defect scenes?

Motivated by this gap, \textbf{Frequency-Supervised Fourier Series Detection (FS-FSD)} is introduced as a contour-native framework for bridge defect analysis rather than as a detector refinement study alone. By extending the contour-native idea of FSD, the framework incorporates a representation-aligned learning design for bridge defect scenes, including direct frequency-domain supervision, order-aware coefficient normalization, and closed-form harmonic phase alignment. Through this formulation, defect boundaries are represented directly as finite-order Fourier contours instead of being described by bounding boxes or dense raster masks. Each defect instance is thereby encoded as a compact, continuous, and directly reconstructable geometric record that is better aligned with subsequent geometric characterization, archival, and reuse requirements.

On this basis, the scope of the study is not confined to detection accuracy alone, but is extended to a broader evidence chain centered on representational quality and workflow relevance. To compare heterogeneous outputs within a shared geometric domain, a representation-aware polygon-space evaluation protocol is established so that assessment can focus on the geometric quality of defect boundaries rather than on the native output form alone. In addition, database-oriented archival and recovery validation, browser-side visualization, and edge-side deployment are incorporated into the same framework to examine the practical implications of this representation in bridge inspection information workflows. The central concern of the paper is therefore not only whether a defect can be detected, but also whether the resulting output can function as a compact, geometry-preserving, retrievable, and reusable inspection record in downstream use.

This study focuses on the representation layer between visual recognition and downstream bridge-inspection information systems. Unless external calibration or registration metadata are available, all geometric outputs are treated as image-space or normalized-image-space records. The workflow experiments therefore evaluate compactness, recoverability, visualization, and on-device executability of the proposed representation, rather than physical-scale metrology or full asset-management deployment.

The main contributions are summarized as follows.
\begin{enumerate}
    \item \textbf{Frequency-Supervised Fourier Series Detection (FS-FSD)} is established as a contour-native framework for bridge defect analysis, in which each defect is represented directly as a finite-order Fourier contour rather than as a bounding box or a dense raster mask. To improve the learnability of this representation in bridge inspection scenes, the framework is coupled with a representation-aligned training formulation, including grid-unit frequency-domain supervision, order-aware coefficient normalization, and closed-form harmonic phase alignment. Bridge defects are thus modeled as compact, continuous, and directly reconstructable geometric records, while Fourier contour learning remains stable and effective within a multi-scale detection architecture.

    \item A representation-aware polygon-space evaluation protocol is introduced to compare box-based, raster-mask-based, and contour-based outputs in a unified geometric domain. This protocol is not intended to replace native task metrics, but to provide a complementary geometry-oriented perspective for cross-paradigm assessment. Under this setting, FS-FSD demonstrates consistently stronger polygon-space geometric fidelity than representative detection and segmentation baselines, especially for irregular, elongated, and fragmented bridge defects, indicating the advantage of contour-native representation for geometry-oriented defect characterization.

    \item The analysis is further extended from geometric prediction to representation-level workflow behavior. Through database-oriented archival and recovery benchmarking, browser-side visualization, and edge-side deployment experiments, contour-native Fourier encoding is evaluated in terms of compact storage, efficient recovery, interactive image-space review, and on-device executability. These component-level validations show that FS-FSD can function as a compact and recoverable image-space defect record, providing a practical representation layer for downstream bridge-inspection data handling.
\end{enumerate}

\section{Related Work}
\label{sec:related_work}

\begin{table}[t]
\centering
\caption{\textbf{Positioning of this study relative to the two dominant research streams in intelligent bridge defect inspection.}
The proposed framework focuses on the underdeveloped representational layer between vision-based defect recognition and AEC information workflows.}
\label{tab:gap_positioning}

\fontsize{6.35}{7.15}\selectfont
\setlength{\tabcolsep}{2.7pt}
\renewcommand{\arraystretch}{1.12}
\setlength{\aboverulesep}{0.34ex}
\setlength{\belowrulesep}{0.34ex}

\begin{threeparttable}
\begin{tabularx}{\linewidth}{@{}>{\RaggedRight\arraybackslash}p{0.19\linewidth}
                            >{\RaggedRight\arraybackslash}p{0.24\linewidth}
                            >{\RaggedRight\arraybackslash}p{0.23\linewidth}
                            >{\RaggedRight\arraybackslash}X@{}}
\toprule
\makecell[c]{\textbf{Research}\\[-1pt]\textbf{stream}} &
\makecell[c]{\textbf{Typical}\\[-1pt]\textbf{focus}} &
\makecell[c]{\textbf{Representative}\\[-1pt]\textbf{output}} &
\makecell[c]{\textbf{Remaining}\\[-1pt]\textbf{limitation}} \\
\midrule

Vision-based recognition
& Detection, segmentation, and boundary delineation of visible defects
& Bounding boxes, raster masks, or contour predictions
& Substantial progress has been achieved in perception, yet the native geometric form of the defect record for long-term archival and downstream reuse remains insufficiently examined. \\

AEC information workflows
& Storage, visualization, semantic linking, and lifecycle-oriented organization of inspection information
& Images, reports, metadata, and inspection records linked to BIM, GIS, or IFC contexts
& Substantial progress has been made in platform-level information management, but defect instances are still often stored as images or raster annotations rather than compact and computable geometric records. \\

\rowcolor{oursbg}
\textbf{This study}
& Representation design between recognition outputs and information workflows
& Fourier contour defect records
& Examines the underdeveloped representational layer by connecting learnable contour outputs with polygon-space evaluation and proof-of-concept archival, recovery, visualization, and deployment validations. \\
\bottomrule
\end{tabularx}

\vspace{1.4pt}

\begin{tablenotes}[flushleft]
\fontsize{5.35}{6.05}\selectfont
\item \textbf{Note.}
BIM/GIS/IFC are listed as representative downstream information contexts and formats.
The present study evaluates a compact geometric representation layer through proof-of-concept workflow validations.
\end{tablenotes}
\end{threeparttable}
\end{table}

Research on intelligent bridge defect inspection has progressively evolved from damage detection toward making damage information more usable in asset management workflows \cite{duque2018synthesis,lyu2025uav,huthwohl2018integrating,moghtadernejad2025digitalizing,boje2020semanticdt,broo2021digitaltwins,fuller2020digitaltwin,wang2026unified,kaveh2025advancing}. Existing studies can be broadly organized into two interconnected streams. The first stream focuses on visual recognition and aims to improve the automatic localization, boundary delineation, and image-based characterization of visible bridge defects. The second stream focuses on information integration and examines how inspection outputs can be organized or linked within BIM, GIS, IFC, and digital twin oriented AEC environments to support asset information management, maintenance planning support, and lifecycle oriented tracking. While the former stream has substantially improved the reliability of defect recognition, and the latter has advanced the digital organization of inspection information, a critical layer remains insufficiently addressed, namely the geometric form that a detected defect should take as a native inspection record for storage, comparison, and reuse. This representational layer constitutes the primary entry point of the present study.

\subsection{Vision-based recognition of bridge defects}

Early work on bridge defect inspection primarily aimed to reduce the subjectivity, labor intensity, and safety risks of manual visual surveys \cite{seo2018drone,duque2018synthesis,agnisarman2019humanloop}. With the emergence of deep learning, bridge defect analysis rapidly shifted from thresholding, edge, and texture based image processing to data-driven visual recognition. Vision-based crack detection studies first demonstrated that visible bridge damage could be automatically identified from images with increasing reliability \cite{yeumdyke2015,cha2017deep}. Subsequent work extended this line of research from crack-only recognition to multi-defect concrete inspection, showing that deep models could detect and distinguish heterogeneous surface anomalies under realistic inspection conditions \cite{adhikari2014imagebridge,lyu2025uav,chen2026improving,mundt2019codebrim,li2025gyudet}.

In terms of localization mechanisms, current bridge defect detectors largely inherit the evolution of generic object detection. Two-stage proposal-based detectors such as Faster Region-based Convolutional Neural Network (Faster R-CNN) established robust region-based localization pipelines \cite{ren2015faster}, while one-stage detectors such as RetinaNet and You Only Look Once (YOLO) improved efficiency and deployment feasibility for large-scale inspection scenarios \cite{lin2017focal,redmonfarhadi2018yolov3}. For elongated or direction-sensitive targets, oriented and rotated-object formulations were further introduced, including the Rotated Region Proposal Network (RRPN), Region of Interest (RoI) Transformer, and Refined Single Stage Detector with Feature Refinement for Rotating Object Detection (R3Det), which explicitly model target orientation in the localization head \cite{ma2018rrpn,ding2019roitransformer,yang2021r3det}. In parallel, anchor-free and keypoint-based formulations such as Fully Convolutional One Stage Object Detection (FCOS) and CenterNet reduced the dependence on predefined anchors and provided more flexible detection heads \cite{tian2019fcos,zhou2019objects}. These paradigms have been adapted to bridge defect localization and supported by bridge-related datasets and benchmarks such as CODEBRIM and GYU-DET \cite{mundt2019codebrim,li2025gyudet}. Their common strength lies in efficient large-area screening and coarse localization.

As the focus moved from object-level localization to boundary-level delineation, semantic and instance segmentation became more widely adopted. Pixel-level damage interpretation has already been demonstrated for multiple concrete defects \cite{cha2017deep,song2024pixel}, and subsequent work has applied U-Net variants, Mask Region-based Convolutional Neural Network (Mask R-CNN), and related dense prediction models to bridge cracks, spalling, corrosion, and exposed rebar \cite{ronneberger2015u,he2017maskrcnn,song2024pixel}. Compared with detection boxes, segmentation masks are more suitable for capturing irregular defect regions and can provide an image-space basis for estimating region extent and boundary-related descriptors. When external scale calibration is available, such masks may further support physical quantity estimation, although the reliability of this step remains sensitive to segmentation stability, calibration quality, and viewing geometry. At the same time, the evaluation focus has gradually expanded from benchmark accuracy under ideal conditions to robustness under realistic inspection environments, because shadows, moisture traces, stains, rough textures, blur, and background clutter can significantly destabilize boundary extraction \cite{agnisarman2019humanloop,lyu2025uav,song2024pixel,chen2026improving}.

More recent studies have begun to move beyond raster masks by parameterizing boundaries more directly. Deep Snake models instance contours through polygon deformation, and Liu et al. further showed that Fourier series can be incorporated into object detection through compact closed-curve parameterization \cite{peng2020deepsnake,liu2025fourierseries}. These studies indicate that contours need not be recovered only through mask post-processing, but can instead be learned as native outputs. Nevertheless, the existing emphasis has remained largely on contour regression feasibility and detection performance. The extent to which contour-native representations can serve as durable geometric records for bridge defects, particularly for archival, comparison, and cross-platform reuse, remains insufficiently explored.

\subsection{Defect information representation, archival, and AEC information workflows}

If visual-recognition studies primarily address whether defects can be reliably perceived by machines, AEC information-modeling studies are more concerned with how such results can be systematically managed after perception. In conventional bridge management practice, defect records remain largely image and metadata centric, typically consisting of photographs, inspection dates, bridge and component identifiers, defect categories, textual descriptions, and, in some cases, manually or automatically generated boxes, contours, or masks. This archival pattern has clear practical value because it preserves direct visual evidence, supports manual verification, and aligns well with current inspection routines and reporting practices. For traceability and auditability, image-centered records remain indispensable.

From the viewpoint of AEC automation, however, the core limitation of this archival pattern is that what is persistently stored is usually the image itself or a raster annotation attached to it, rather than a stable, computable, and reusable defect record. This limitation has already been recognized in infrastructure information-modeling research. Inspection information has been progressively moved from reports and forms into structured workflows, for example through BIM and IFC oriented models that explicitly connect inspection tasks, bridge elements, and inspection outcomes in machine-readable form \cite{huthwohl2018integrating,moghtadernejad2025digitalizing,chen2023roboticsbim,chen2023aligningbim}. Bridge studies have further shown that defect geometry, spatial placement, and semantic relations can be described parametrically rather than merely attached as external evidence \cite{huthwohl2018integrating,moghtadernejad2025digitalizing,chen2023aligningbim}. This marks an important shift from documenting defects to modeling defects as information objects.

Recent work has further extended this direction toward semantic enrichment and platform integration. Defect ontologies and defect-aware information models have started to formalize the relationships among defect categories, diagnosis, causes, and repair actions, thereby improving the machine interpretability of inspection knowledge. In parallel, bridge and building studies have explored BIM-based storage and visualization of defect information, as well as digital workflows that connect inspection data with analytics and decision support oriented tools \cite{huthwohl2018integrating,moghtadernejad2025digitalizing,chen2021georegistering,chen2023gisfacade,chen2023roboticsbim,chen2023aligningbim,nair2025gisbimreview,wang2026unified}. In the broader information chain, BIM and IFC can provide component-level semantic anchors, GIS, geospatial BIM (GeoBIM), and web-based GIS (WebGIS) can support spatial indexing and cross-asset querying, and digital twin frameworks are increasingly used to organize links among geometry, semantics, time-varying observations, and operational status \cite{huthwohl2018integrating,moghtadernejad2025digitalizing,chen2023aligningbim,mohammed2026bibliometric,chen2021georegistering,chen2023gisfacade,boje2020semanticdt,broo2021digitaltwins,fuller2020digitaltwin,kong2023photogrammetrydt,wang2026unified,kaveh2025advancing}. 

Despite these advances, one fundamental issue remains unresolved. Many existing AEC information-modeling studies focus more strongly on where defect information can be stored and how it can be linked within BIM, GIS, or digital twin oriented systems than on what geometric form the defect itself should take as the native archival unit. This distinction becomes particularly important in repeated inspection settings. When the same defect is recorded across multiple inspections, image-centric and raster-centric archival typically still requires additional registration, segmentation refinement, pixel-level comparison, or manual interpretation before temporal changes can be assessed. In other words, the information platform has advanced, but the defect record itself has not necessarily been upgraded from visual evidence to a compact and computable geometric entity.

\subsection{Research gap and positioning of this study}

Taken together, existing studies have advanced two essential dimensions of intelligent bridge inspection: how defects are recognized and how inspection information is managed. What remains comparatively underdeveloped is the representational layer that would allow results to move effectively between these two dimensions. Current recognition methods typically output boxes or raster masks. Existing AEC information platforms and models can store and link inspection information, yet they still mostly receive images, text, or loosely structured geometries rather than compact, stable, and reusable defect records. A sufficiently lightweight and geometry-preserving representation is therefore still missing from the automation chain.

The present study addresses this gap not primarily by proposing another detector solely for higher recognition accuracy, and not by introducing another platform solely for defect organization, but by upgrading the output of bridge defect recognition into a geometry-preserving, queryable, and archive-ready vectorized inspection record. The central objective is to elevate bridge defect outputs from transient visual evidence to persistent geometric records. To this end, representation learning, polygon-space evaluation, and workflow-oriented proof-of-concept validation, including archival and recovery, browser-side visualization, and edge deployment, are examined within a unified framework. The resulting focus is not only whether a defect can be detected, but whether the resulting output can be produced, stored, recovered, and reused as a compact geometric record in downstream bridge inspection information workflows.

To clarify the positioning of the study, Table~\ref{tab:gap_positioning} summarizes the two dominant research streams and highlights the underdeveloped representational layer addressed in this work.

\section{Methodology}
\label{sec:methodology}

To upgrade bridge defect outputs from image-centric visual evidence to compact, recoverable, and reusable inspection records, FS-FSD builds a contour-native Fourier output on top of a one-stage detection framework. The inherited detector provides efficient multi-scale feature extraction, candidate localization, and category discrimination, while the proposed Fourier branch directly predicts a fixed-dimensional contour descriptor for each defect instance. The emphasis of this section is therefore not on conventional box-level localization, but on how a detected defect is represented, supervised, decoded, and organized as a compact geometric record.

\subsection{Overview of FS-FSD}
\label{subsec:overview}

FS-FSD consists of a backbone, a feature-fusion neck, and Fourier-oriented prediction heads. A medium-scale YOLO11-style realization is adopted. Three detection levels are produced with strides of 8, 16, and 32, corresponding to $P3$, $P4$, and $P5$. These levels mainly support small-, medium-, and large-scale defect instances, respectively. The backbone extracts hierarchical visual features from UAV bridge images, while the neck fuses shallow spatial details and deep semantic cues through a PAN-FPN-style multi-scale structure. This shared detection skeleton provides the feature basis for both category prediction and contour prediction.

\begin{figure}[pos=htbp]
    \centering
    \includegraphics[width=\linewidth]{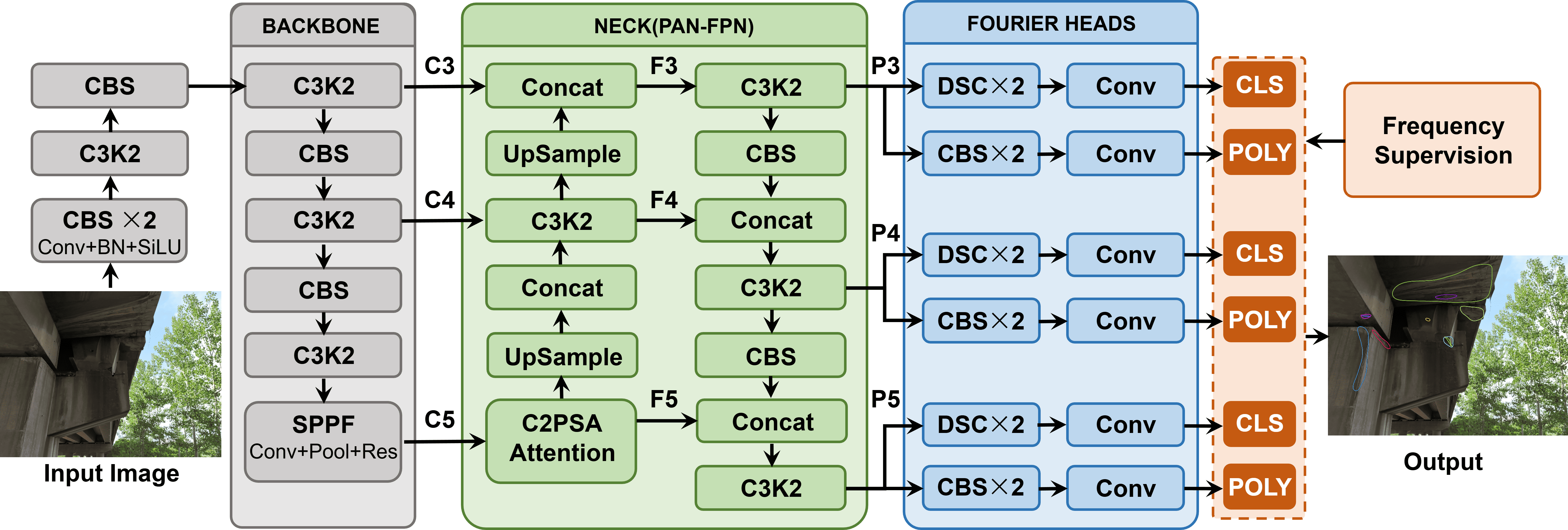}
    \caption{Architecture of FS-FSD.}
    \label{fig:fsfsd_arch}
\end{figure}

The overall architecture is shown in Fig.~\ref{fig:fsfsd_arch}. In the figure, CBS denotes a convolutional block composed of

\[
\mathrm{CBS}=\mathrm{Conv}(2\times2,\ \mathrm{stride}=2)+\mathrm{BN}+\mathrm{SiLU},
\]
where BN denotes batch normalization and SiLU denotes the sigmoid linear unit activation. DSC stands for Depthwise Separable Convolution. The CLS branch predicts defect categories, whereas the POLY branch predicts Fourier contour coefficients. The frequency supervision module is applied to the Fourier contour output during training so that the predicted contour descriptor is learned directly in the frequency domain.

The distinction between FS-FSD and the FSD baseline lies primarily in the representation-learning strategy rather than in the use of a Fourier contour basis alone. Both methods adopt a contour-native Fourier descriptor. However, FS-FSD further adapts the descriptor learning process to multi-scale bridge defect scenes. Specifically, unit-grid direct frequency-domain supervision (U-DFS) is used to reduce scale-induced coefficient imbalance, order-aware coefficient normalization (FT norm) is introduced to rebalance different harmonic orders, and closed-form harmonic phase alignment (CHPA) is adopted to reduce the starting-point ambiguity of closed contours. These components are designed to make Fourier contour regression more stable for irregular bridge defects.

To clarify the methodological positioning of FS-FSD relative to FSD, Table~\ref{tab:fsd_vs_fsfsd} summarizes their main differences at the representation-learning level. The purpose of this comparison is not to reproduce all implementation details of FSD, but to highlight the extensions introduced in this study for bridge defect contour learning.

\begin{table}[htbp]
\centering
\caption{Representation-learning distinction between FSD and FS-FSD.}
\label{tab:fsd_vs_fsfsd}
\fontsize{8.2}{9.4}\selectfont
\setlength{\tabcolsep}{4.5pt}
\renewcommand{\arraystretch}{1.12}
\begin{tabularx}{\linewidth}{@{}p{0.32\linewidth}p{0.25\linewidth}X@{}}
\toprule
\textbf{Item} & \textbf{FSD baseline} & \textbf{FS-FSD} \\
\midrule
Backbone and neck & YOLOv5-style realization & YOLO11-style lightweight realization \\
Native contour basis & Fourier contour & Fourier contour \\
Scale handling of Fourier coefficients & No grid-unit frequency supervision & Unit-grid direct frequency-domain supervision \\
Harmonic-order balancing & No order-aware normalization & Order-aware coefficient normalization \\
Starting-point ambiguity in closed contours & No closed-form harmonic phase alignment & Closed-form harmonic phase alignment \\
Representation-learning role & Contour-native detection baseline & Representation-aligned Fourier contour learning for bridge defects \\
\bottomrule
\end{tabularx}
\end{table}

\subsection{Contour-native Fourier representation and prediction head}
\label{subsec:fourier_head}

For a closed defect contour represented by a finite Fourier order $n_f$, the Fourier parameter vector is defined as
\begin{equation}
\mathbf{f}
=
[a_0,\; c_0,\; a_1,\; b_1,\; c_1,\; d_1,\; \ldots,\; a_{n_f},\; b_{n_f},\; c_{n_f},\; d_{n_f}]^\top
\in
\mathbb{R}^{M},
\qquad
M = 2 + 4n_f .
\label{eq:fourier_vector}
\end{equation}
Here, $(a_0,c_0)$ are the zeroth-order Fourier coefficients and correspond to the contour center. This notation is equivalent to writing $(x_c,y_c)=(a_0,c_0)$, but it makes the channel layout of the Fourier head explicit. The remaining $4n_f$ terms are the harmonic coefficients used to describe the contour shape.

The corresponding closed contour is reconstructed as
\begin{equation}
x(t)
=
a_0
+
\sum_{k=1}^{n_f}
\left(
a_k\cos kt + b_k\sin kt
\right),
\label{eq:fourier_x}
\end{equation}
\begin{equation}
y(t)
=
c_0
+
\sum_{k=1}^{n_f}
\left(
c_k\cos kt + d_k\sin kt
\right),
\qquad
t\in[0,2\pi].
\label{eq:fourier_y}
\end{equation}
In this formulation, the zeroth-order coefficients determine the center location, low-order harmonics mainly describe the global contour shape, and higher-order harmonics progressively encode local boundary details.

For ordered contour annotations or resampled polygon boundaries, the coefficient vector is obtained through a discrete Fourier expansion. Let the uniformly resampled closed boundary points be denoted by $\{(x_q,y_q)\}_{q=0}^{Q-1}$, with

\[
t_q=\frac{2\pi q}{Q}.
\]
The zeroth-order coefficients are computed as
\begin{equation}
a_0 = \frac{1}{Q}\sum_{q=0}^{Q-1}x_q,
\qquad
c_0 = \frac{1}{Q}\sum_{q=0}^{Q-1}y_q .
\label{eq:center_discrete}
\end{equation}
For $k=1,\ldots,n_f$, the harmonic coefficients are computed as
\begin{equation}
a_k = \frac{2}{Q}\sum_{q=0}^{Q-1}x_q\cos(kt_q),
\qquad
b_k = \frac{2}{Q}\sum_{q=0}^{Q-1}x_q\sin(kt_q),
\label{eq:x_coeff_discrete}
\end{equation}
\begin{equation}
c_k = \frac{2}{Q}\sum_{q=0}^{Q-1}y_q\cos(kt_q),
\qquad
d_k = \frac{2}{Q}\sum_{q=0}^{Q-1}y_q\sin(kt_q).
\label{eq:y_coeff_discrete}
\end{equation}
Through this parameterization, an arbitrarily long contour-point sequence is compressed into a fixed-dimensional vector of length $2+4n_f$. The defect boundary can then be predicted, stored, reconstructed, and reused through this compact vector instead of through a dense raster mask.

\begin{figure}[pos=htbp]
    \centering
    \includegraphics[width=\linewidth]{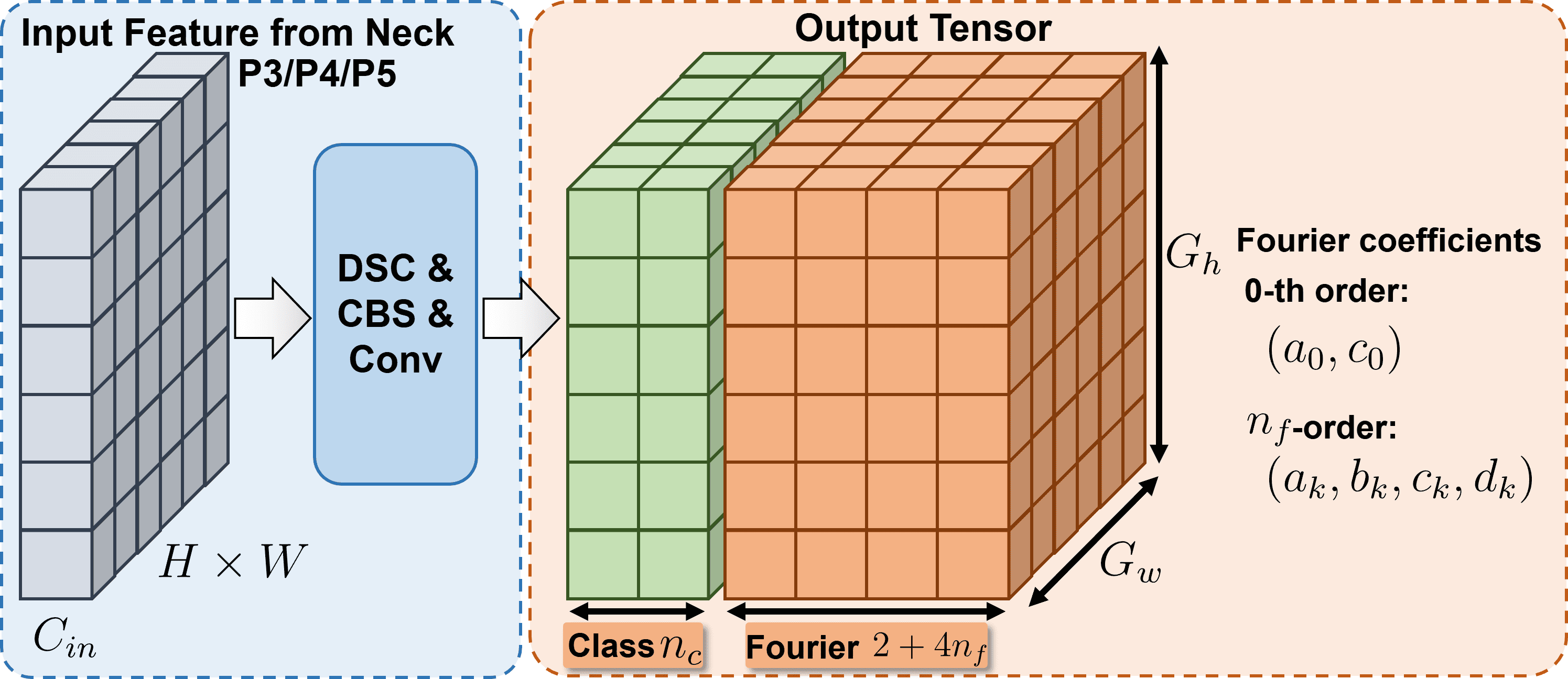}
    \caption{Fourier prediction head.}
    \label{fig:fsfsd_head}
\end{figure}

At each detection level, FS-FSD employs a Fourier prediction head to produce category scores and contour coefficients. As illustrated in Fig.~\ref{fig:fsfsd_head}, the head receives the feature map from the neck at level $P3$, $P4$, or $P5$. The classification branch outputs $n_c$ category channels, where $n_c$ is the number of defect classes. The contour branch outputs the Fourier coefficient channels.

For detection level $l$ and grid location $(u,v)$, the prediction tensor is written as
\begin{equation}
\mathbf{o}_{l,u,v}
=
\left[
\hat{\mathbf{s}}_{l,u,v},
\;
\hat{\mathbf{f}}^{grid}_{l,u,v}
\right],
\label{eq:head_output}
\end{equation}
where
\begin{equation}
\hat{\mathbf{s}}_{l,u,v}\in\mathbb{R}^{n_c},
\qquad
\hat{\mathbf{f}}^{grid}_{l,u,v}\in\mathbb{R}^{2+4n_f}.
\label{eq:cls_fourier_output}
\end{equation}
The Fourier part is organized as
\begin{equation}
\hat{\mathbf{f}}^{grid}_{l,u,v}
=
[
\hat{a}^{grid}_0,\;
\hat{c}^{grid}_0,\;
\hat{a}^{grid}_1,\;
\hat{b}^{grid}_1,\;
\hat{c}^{grid}_1,\;
\hat{d}^{grid}_1,\;
\ldots,\;
\hat{a}^{grid}_{n_f},\;
\hat{b}^{grid}_{n_f},\;
\hat{c}^{grid}_{n_f},\;
\hat{d}^{grid}_{n_f}
]^\top .
\label{eq:grid_fourier_vector}
\end{equation}

The channel layout of the Fourier branch can be summarized as
\begin{equation}
\mathrm{Fourier\ channels}=2+4n_f .
\label{eq:fourier_channels}
\end{equation}
Here, the first two channels are the zeroth-order Fourier coefficients $(a_0,c_0)$, which correspond to the object center. The remaining $4n_f$ channels are occupied by the harmonic Fourier coefficients $(a_k,b_k,c_k,d_k)$ for $k=1,\ldots,n_f$.

The Fourier branch directly regresses continuous grid-unit coefficients. The first two Fourier channels represent the local center terms relative to the responsible grid location, and the remaining channels represent the grid-unit harmonic coefficients. Therefore, a reconstructable contour descriptor is obtained directly from the prediction head, without first generating a dense mask and then extracting a contour through post-processing.

For detection level $l$ with stride $s_l$, the grid-unit Fourier coefficients are decoded into image-space coefficients as
\begin{equation}
\hat{a}_0
=
(u+\hat{a}^{grid}_0)s_l,
\qquad
\hat{c}_0
=
(v+\hat{c}^{grid}_0)s_l,
\label{eq:center_decode}
\end{equation}
\begin{equation}
\hat{a}_k = \hat{a}^{grid}_k s_l,
\qquad
\hat{b}_k = \hat{b}^{grid}_k s_l,
\qquad
\hat{c}_k = \hat{c}^{grid}_k s_l,
\qquad
\hat{d}_k = \hat{d}^{grid}_k s_l,
\qquad
k=1,\ldots,n_f .
\label{eq:coef_decode}
\end{equation}
The decoded image-space Fourier vector is therefore
\begin{equation}
\hat{\mathbf{f}}
=
[
\hat{a}_0,\;
\hat{c}_0,\;
\hat{a}_1,\;
\hat{b}_1,\;
\hat{c}_1,\;
\hat{d}_1,\;
\ldots,\;
\hat{a}_{n_f},\;
\hat{b}_{n_f},\;
\hat{c}_{n_f},\;
\hat{d}_{n_f}
]^\top .
\label{eq:decoded_fourier_vector}
\end{equation}
The reconstructed image-space contour is then obtained by substituting $\hat{\mathbf{f}}$ into Eqs.~\eqref{eq:fourier_x} and \eqref{eq:fourier_y}. This decoding process provides a consistent basis for contour reconstruction, image-space geometric characterization, compact archival, and later reuse.

\subsection{Unit-grid direct frequency-domain supervision}
\label{subsec:udfs}

Although the final contour is reconstructed in image coordinates, supervision is defined on multi-scale feature grids during training. If Fourier coefficients are supervised directly in pixel space, large defects typically produce coefficients with much larger magnitudes than small defects, making optimization sensitive to object scale. This issue is particularly relevant to bridge inspection images, where small cracks, elongated seepage regions, and larger spalling areas may coexist in the same dataset. U-DFS addresses this problem by transforming all Fourier coefficients into the grid coordinate system of the responsible detection level before supervision is imposed.

Let the input image size be $(W,H)$ and let the stride of the $l$-th detection level be $s_l$. The corresponding grid size is
\begin{equation}
G_l^w=\frac{W}{s_l},
\qquad
G_l^h=\frac{H}{s_l}.
\label{eq:grid_size}
\end{equation}

If the annotation is given in normalized image coordinates, the contour center and harmonic coefficients can be mapped to grid units at level $l$. The local supervision for the contour center is defined as
\begin{equation}
\Delta x_c^{(l)} = G_l^w \tilde{x}_c - u,
\qquad
\Delta y_c^{(l)} = G_l^h \tilde{y}_c - v,
\label{eq:center_grid_target}
\end{equation}
and the harmonic coefficients are supervised as
\begin{equation}
a_n^{(l)}=G_l^w \tilde{a}_n,\qquad
b_n^{(l)}=G_l^w \tilde{b}_n,
\label{eq:x_grid_target}
\end{equation}
\begin{equation}
c_n^{(l)}=G_l^h \tilde{c}_n,\qquad
d_n^{(l)}=G_l^h \tilde{d}_n,
\qquad n=1,\ldots,N.
\label{eq:y_grid_target}
\end{equation}

Through this transformation, large defects are normalized on coarse grids and small defects are normalized on fine grids. Fourier regression targets of different object scales are therefore kept within more comparable numerical ranges, which alleviates scale-induced optimization imbalance. Unlike the baseline FSD formulation, where the Fourier representation is mainly introduced as a contour output, U-DFS explicitly adapts frequency-domain supervision to the multi-scale grid structure of the detector. At the same time, supervision remains entirely in the frequency domain and does not require repeated inverse contour transforms or dense resampling during training.

\subsection{Order-aware coefficient normalization}
\label{subsec:ftnorm}

A practical issue of Fourier regression is that different harmonic orders usually exhibit markedly different value distributions. Low-order terms mainly determine the global contour shape and often have relatively large magnitudes, whereas high-order terms mainly encode local details and are usually much closer to zero. If all orders are optimized on the same numerical scale, low-order terms tend to dominate the regression process, while high-order detail learning is weakened. FT norm is introduced to rebalance the relative contribution of different harmonic orders during training.

Let the coefficient group of the $n$-th harmonic be
\begin{equation}
\mathbf q_n=[a_n,b_n,c_n,d_n]^\top.
\label{eq:harmonic_group}
\end{equation}
Let $\omega_n$ denote the normalization weight of the $n$-th harmonic. The Fourier coefficient regression loss is then defined as
\begin{equation}
L_{coef}
=
\frac{1}{S}
\sum_i
w_i
\sum_{n=1}^{N}
\omega_n\,
\mathrm{SL1}\!\left(
\hat{\mathbf q}_{i,n},
\mathbf q_{i,n}^{*}
\right),
\label{eq:coef_loss}
\end{equation}
where $\mathrm{SL1}(\cdot,\cdot)$ denotes element-wise Smooth L1 loss, $w_i$ denotes the positive-sample weight, $S$ is the normalization factor, and $\mathbf q_{i,n}^{*}$ denotes the phase-aligned target coefficients. The weight $\omega_n$ reflects the typical magnitude scale of the $n$-th harmonic in the training data and prevents low-order terms from dominating optimization simply because of their larger numerical range, while still ensuring that high-order terms receive sufficient supervision for local boundary detail learning.

From a representational perspective, FT norm does not alter the Fourier contour itself. Its role is to correct the unequal optimization influence of different harmonic orders. Since higher-order harmonics are mainly responsible for local boundary detail, the absence of such order-aware normalization would make the optimization process more likely to favor low-order global-shape fitting while under-learning local curvature and boundary perturbations. FT norm therefore acts as a representation-aligned numerical rescaling mechanism, allowing different harmonic orders to participate in training more evenly within a unified optimization framework. This mechanism further differentiates FS-FSD from FSD, where harmonic-order imbalance is not explicitly handled during coefficient learning.

\subsection{Closed-form harmonic phase alignment}
\label{subsec:chpa}

\begin{figure}[pos=h]
    \centering
    \includegraphics[width=0.8\linewidth]{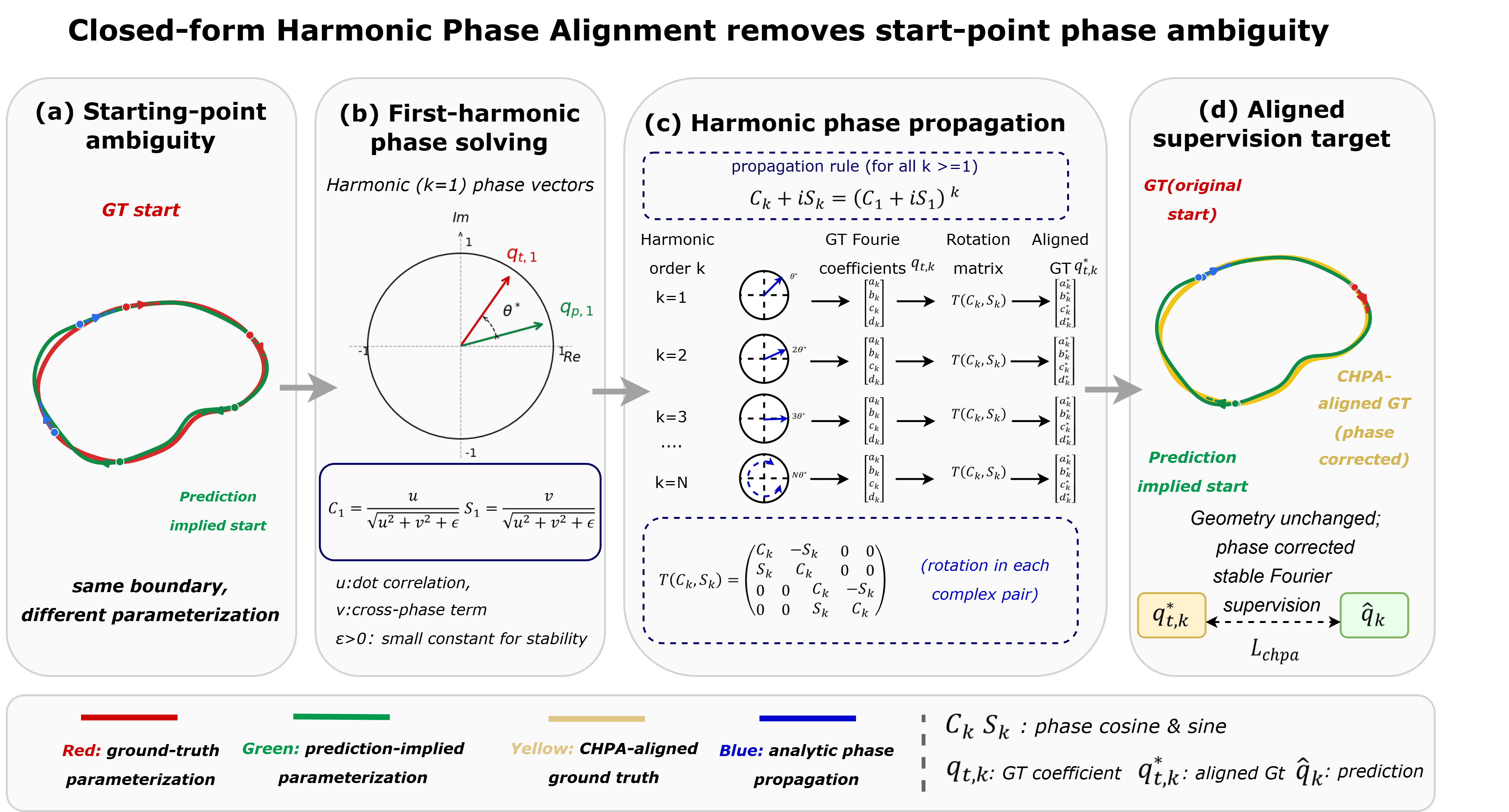}
    \caption{Analytical principle of CHPA. The procedure resolves starting-point ambiguity by solving the first-harmonic phase, propagating the phase to higher harmonic orders, and rotating the ground truth coefficients into an aligned supervision target.}
    \label{fig:chpa_principle}
\end{figure}

A fundamental ambiguity of closed contours is that the same geometric boundary can be parameterized from arbitrary starting points. Consequently, even if two contours are geometrically identical, their Fourier coefficients may differ because of a phase shift in the underlying parameterization. CHPA is adopted to remove this supervision inconsistency.

\begin{figure}[pos=h]
    \centering
    \includegraphics[width=0.8\linewidth]{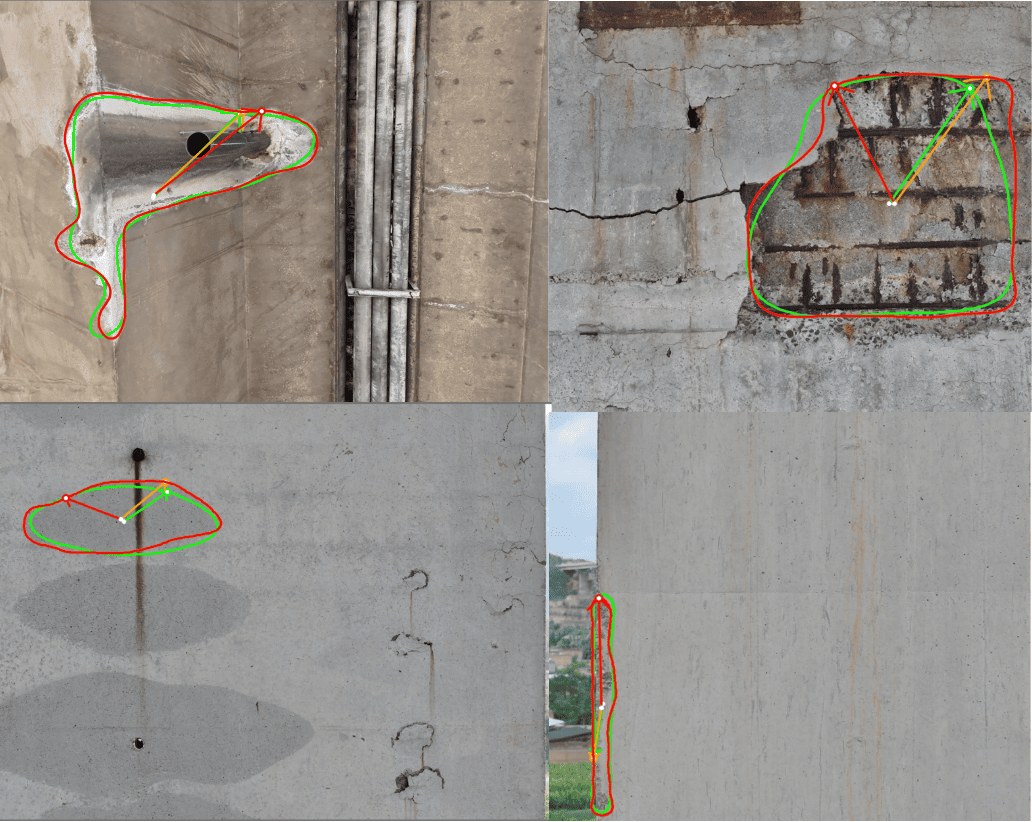}
    \caption{Qualitative visualization of CHPA in bridge defect images. Red indicates the original ground truth parameterization, green indicates the prediction-implied parameterization, and yellow indicates the phase-aligned ground truth reference used to stabilize Fourier coefficient supervision.}
    \label{fig:chpa_visualization}
\end{figure}

The key observation is that the coefficient variation induced by different contour starting points is a phase rotation in harmonic space. Once the base phase is solved analytically, all higher-order harmonics can be aligned through harmonic phase propagation, without evaluating multiple candidate starting points.

Let the ground-truth and predicted coefficient vectors of the $k$-th harmonic be
\begin{equation}
\mathbf q_{t,k}=[a_{t,k},b_{t,k},c_{t,k},d_{t,k}]^\top,
\qquad
\hat{\mathbf q}_{k}=[a_{p,k},b_{p,k},c_{p,k},d_{p,k}]^\top.
\label{eq:qtk_qpk}
\end{equation}
For a phase shift $\theta$, the aligned ground-truth coefficients are written as
\begin{equation}
\mathbf q_{t,k}^{*}
=
\mathbf T(C_k,S_k)\,\mathbf q_{t,k},
\qquad
C_k=\cos(k\theta),\;
S_k=\sin(k\theta),
\label{eq:chpa_transform}
\end{equation}
where
\begin{equation}
\mathbf T(C_k,S_k)
=
\begin{bmatrix}
C_k & S_k & 0 & 0\\
-S_k & C_k & 0 & 0\\
0 & 0 & C_k & S_k\\
0 & 0 & -S_k & C_k
\end{bmatrix}.
\label{eq:chpa_matrix}
\end{equation}

The base phase is solved analytically from the first harmonic. Let
\begin{equation}
\mathbf q_{t,1}=[a_{t,1},b_{t,1},c_{t,1},d_{t,1}]^\top,
\qquad
\hat{\mathbf q}_{1}=[a_{p,1},b_{p,1},c_{p,1},d_{p,1}]^\top.
\label{eq:first_harmonic}
\end{equation}
Define
\begin{equation}
u
=
a_{t,1}a_{p,1}
+
b_{t,1}b_{p,1}
+
c_{t,1}c_{p,1}
+
d_{t,1}d_{p,1},
\label{eq:chpa_u}
\end{equation}
\begin{equation}
v
=
b_{t,1}a_{p,1}
-
a_{t,1}b_{p,1}
+
d_{t,1}c_{p,1}
-
c_{t,1}d_{p,1}.
\label{eq:chpa_v}
\end{equation}
Then the optimal base phase is obtained in closed form as
\begin{equation}
C_1=\cos\theta^{*}=\frac{u}{\sqrt{u^2+v^2+\varepsilon}},
\qquad
S_1=\sin\theta^{*}=\frac{v}{\sqrt{u^2+v^2+\varepsilon}},
\label{eq:chpa_base}
\end{equation}
where $\varepsilon$ is a small numerical stabilizer. Once $(C_1,S_1)$ has been determined, the phase terms of higher-order harmonics are generated analytically through harmonic phase propagation:
\begin{equation}
C_k + iS_k = (C_1 + iS_1)^k,
\qquad k=1,\ldots,N.
\label{eq:angle_multiple}
\end{equation}
These phase terms are then used to rotate the target coefficients of all orders into an aligned supervision target.

This procedure does not alter the underlying contour geometry. What is corrected is the parameterization phase associated with the contour starting point. CHPA therefore removes a parameterization inconsistency rather than a geometric discrepancy. Compared with search-based phase adjustment, this closed-form procedure avoids the additional computational burden of explicit rolling search and prevents phase quantization error induced by a finite candidate interval.

The analytical workflow of CHPA is illustrated in Fig.~\ref{fig:chpa_principle}. First, the same closed boundary may have different starting point parameterizations, which leads to inconsistent Fourier coefficients for geometrically equivalent contours. Second, the base phase is solved from the first harmonic by using the dot correlation term and the cross phase term. Third, the solved phase is propagated to higher harmonic orders through analytic harmonic phase propagation. Finally, the ground truth coefficients are rotated into a phase-aligned supervision target, so that coefficient regression is performed against a stable target in the Fourier domain.

Qualitative examples of CHPA in bridge defect scenes are provided in Fig.~\ref{fig:chpa_visualization}. In these examples, the red contour denotes the original ground truth parameterization, the green contour denotes the parameterization implied by the prediction, and the yellow line or marker denotes the analytically corrected phase relationship for the same ground truth contour. The yellow overlay should therefore be interpreted as a phase-aligned ground truth reference used for supervision, rather than as an additional prediction or a modified defect boundary.

\subsection{Training objective}
\label{subsec:loss}

The total training objective combines the inherited detector loss with Fourier-specific supervision. To keep the methodological focus on the proposed contour representation branch, the inherited detector loss is denoted collectively by $L_{det}$ and is not expanded here. The training objective adopted in this study is written as
\begin{equation}
L
=
L_{det}
+
\lambda_{xy} L_{xy}
+
\lambda_{coef} L_{coef},
\label{eq:total_loss}
\end{equation}
where $L_{xy}$ supervises the zeroth-order center terms $(a_0,c_0)$ in grid units, and $L_{coef}$ supervises the CHPA-aligned harmonic Fourier coefficients.

The center-term loss is defined as
\begin{equation}
L_{xy}
=
\frac{1}{S}
\sum_{i\in\mathcal{P}}
w_i\,
\mathrm{SL1}\!\left(
\hat{\Delta \mathbf z}_{0,i},
\Delta \mathbf z_{0,i}
\right),
\label{eq:center_loss}
\end{equation}
where $\mathcal{P}$ denotes the set of positive samples, $w_i$ is the positive-sample weight, and $S$ is the normalization factor. The predicted and target center-offset vectors are defined as
\begin{equation}
\hat{\Delta \mathbf z}_{0,i}
=
[
\hat{a}^{grid}_{0,i},
\hat{c}^{grid}_{0,i}
]^\top,
\qquad
\Delta \mathbf z_{0,i}
=
[
\Delta a^{(l_i)}_{0,i},
\Delta c^{(l_i)}_{0,i}
]^\top,
\label{eq:center_vector_loss}
\end{equation}
where $l_i$ denotes the responsible detection level of the $i$-th positive sample. Here, $\Delta a^{(l_i)}_{0,i}$ and $\Delta c^{(l_i)}_{0,i}$ are the grid-unit target offsets of the zeroth-order Fourier coefficients defined by U-DFS.

The coefficient loss follows Eq.~\eqref{eq:coef_loss}. It is applied to the continuous harmonic Fourier coefficients after CHPA-based phase alignment. Since the Fourier branch directly regresses continuous grid-unit Fourier parameters, the final optimization is centered on the accuracy of the zeroth-order center terms and the phase-aligned harmonic coefficient regression.

\subsection{Inference and archive-ready contour record}
\label{subsec:inference}

During inference, each retained defect candidate is associated with a responsible detection level $l_i$ and grid location $(u_i,v_i)$. The Fourier branch directly outputs continuous grid-unit Fourier coefficients
$\hat{\mathbf f}^{grid}_i\in\mathbb{R}^{2+4n_f}$. These coefficients are decoded into an image-space Fourier vector
$\hat{\mathbf f}_i\in\mathbb{R}^{2+4n_f}$ according to the stride $s_{l_i}$ of the responsible detection level, following the decoding rules in Eqs.~\eqref{eq:center_decode} and \eqref{eq:coef_decode}. Thus, each retained defect instance is represented by a compact fixed-dimensional Fourier descriptor.

Let the decoded image-space Fourier vector of the $i$-th defect instance be written as
\begin{equation}
\hat{\mathbf f}_i
=
[
\hat{a}_{0,i},\;
\hat{c}_{0,i},\;
\hat{a}_{1,i},\;
\hat{b}_{1,i},\;
\hat{c}_{1,i},\;
\hat{d}_{1,i},\;
\ldots,\;
\hat{a}_{n_f,i},\;
\hat{b}_{n_f,i},\;
\hat{c}_{n_f,i},\;
\hat{d}_{n_f,i}
]^\top .
\label{eq:inference_fourier_vector}
\end{equation}
At an arbitrary sampling density $T$, the closed contour is reconstructed by evaluating the Fourier series at uniformly sampled phase values
\begin{equation}
t_s = \frac{2\pi s}{T},
\qquad
s=0,\ldots,T-1 .
\label{eq:inference_sampling}
\end{equation}
The reconstructed contour is therefore obtained as
\begin{equation}
\hat{\mathcal C}_i
=
\left\{
\bigl(\hat{x}_i(t_s),\hat{y}_i(t_s)\bigr)
\right\}_{s=0}^{T-1},
\label{eq:inference_contour}
\end{equation}
where
\begin{equation}
\hat{x}_i(t_s)
=
\hat{a}_{0,i}
+
\sum_{k=1}^{n_f}
\left(
\hat{a}_{k,i}\cos kt_s
+
\hat{b}_{k,i}\sin kt_s
\right),
\label{eq:inference_x}
\end{equation}
\begin{equation}
\hat{y}_i(t_s)
=
\hat{c}_{0,i}
+
\sum_{k=1}^{n_f}
\left(
\hat{c}_{k,i}\cos kt_s
+
\hat{d}_{k,i}\sin kt_s
\right).
\label{eq:inference_y}
\end{equation}

Unlike conventional segmentation pipelines, this process does not require a dense raster mask to be generated and then converted into geometry through post-processing. Instead, the network directly provides a compact, reconstructable, and structurally meaningful contour descriptor. The resulting archive-ready defect record can therefore be written as
\begin{equation}
\mathcal R_i
=
\left\{
\hat{\ell}_i,\;
\hat{s}_i,\;
\hat{\mathbf f}_i,\;
\hat{\mathcal C}_i
\right\},
\label{eq:record}
\end{equation}
where $\hat{\ell}_i$ denotes the predicted defect category, $\hat{s}_i$ denotes the confidence score, $\hat{\mathbf f}_i$ denotes the decoded image-space Fourier coefficient vector, and $\hat{\mathcal C}_i$ denotes the reconstructed image-space contour. The inherited support region from the detector may still be retained as auxiliary localization metadata, but it is no longer treated as the native representation of the defect. FS-FSD therefore produces not only a detection result, but also a compact geometric inspection record that can support downstream archival, recovery, visualization, comparison, and, when external scale or registration information is available, measurement-oriented analysis.

\section{Experiments and Geometric Evaluation}
\label{sec:experiments}

\subsection{Experimental platform}
\label{subsec:experimental_platform}

All experiments were conducted on a workstation equipped with an AMD Ryzen 9 9950X central processing unit (CPU), 96 GB of random access memory (RAM), and an NVIDIA GeForce RTX 5090 graphics processing unit (GPU) with 32 GB of video random access memory (VRAM), running the Windows 11 operating system.

\subsection{Dataset acquisition, partitioning, and task-oriented supervision}
\label{subsec:dataset}

\begin{figure}[pos=h]
    \centering
    \includegraphics[width=\linewidth]{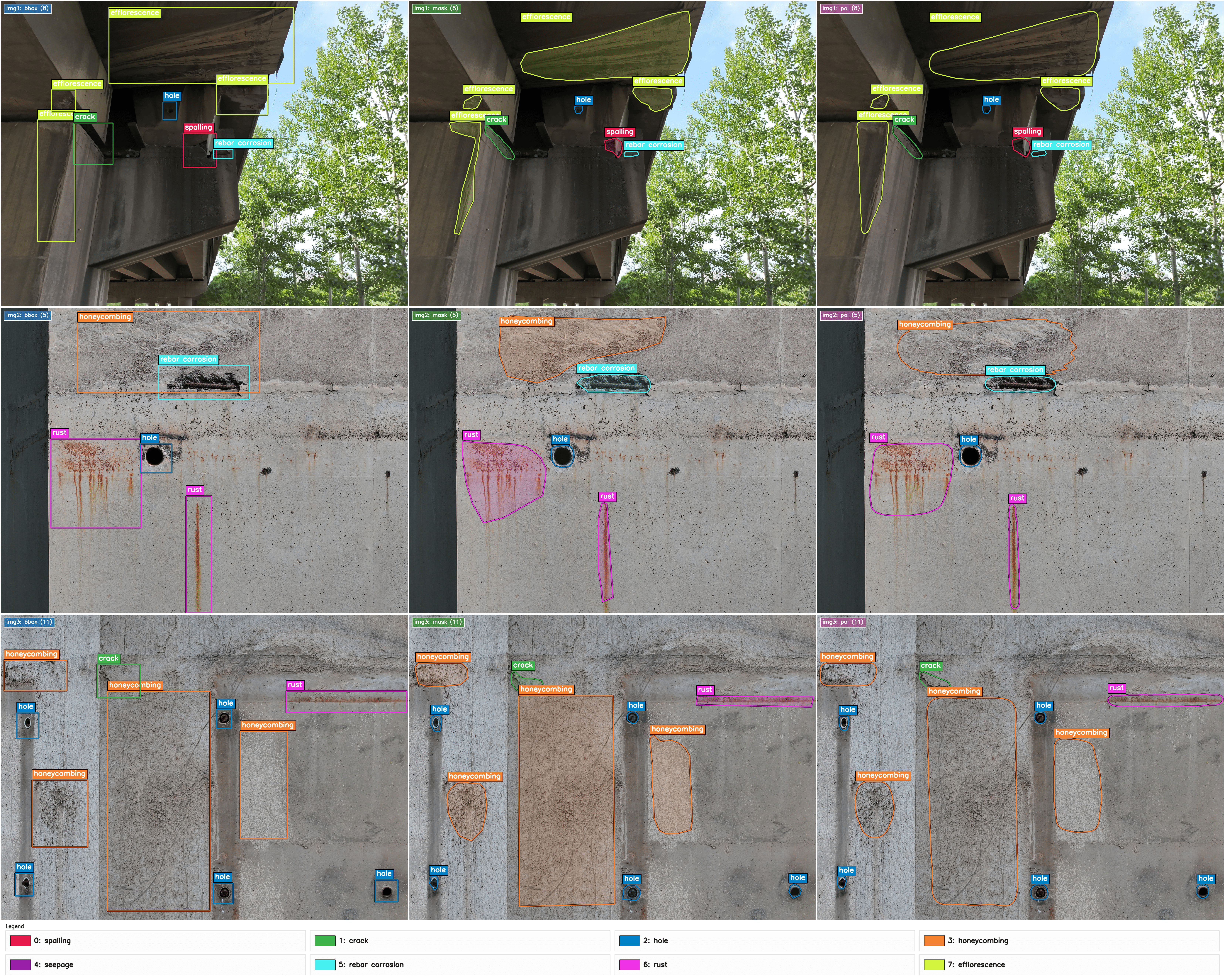}
    \caption{Representative bridge defect samples from the field dataset.}
    \label{fig:compare_3x3_exact}
\end{figure}

To support bridge defect detection, segmentation, and geometric representation learning, a field dataset was constructed from real bridge inspection scenarios. The acquisition campaign covered seven bridges and four representative inspection contexts, namely urban viaducts, mountainous bridges, railway bridges, and long-span highway bridges. This sampling design was intended to increase diversity in bridge type, background environment, illumination condition, imaging viewpoint, and defect morphology. All images were acquired using a UAV platform equipped with a DJI H30T imaging system, with an original spatial resolution of $4032 \times 3024$ pixels. Representative samples are shown in Fig.~\ref{fig:compare_3x3_exact}.

In total, the dataset contains 3,767 images, of which 2,895, 495, and 377 images were allocated to the training, validation, and test sets, respectively. To reduce the risk of data leakage caused by repeated acquisition, adjacent viewpoints, or highly similar scene content, the partition was organized to remain approximately balanced with respect to bridge source and inspection context. Duplicate image screening was also conducted before splitting, and no repeated images were shared across the training, validation, and test subsets. This partition strategy reduces performance inflation caused by repeated views or scene memorization and makes the subsequent evaluation more reflective of genuine defect-shape learning.

\begin{table}[pos=h]
\centering
\caption{Overview of the field bridge defect dataset.}
\label{tab:dataset_overview}
\fontsize{8.2}{9.4}\selectfont
\setlength{\tabcolsep}{4.5pt}
\renewcommand{\arraystretch}{1.12}
\begin{tabularx}{\linewidth}{@{}p{0.38\linewidth}X@{}}
\toprule
\textbf{Item} & \textbf{Content} \\
\midrule
Number of bridges & 7 \\
Inspection contexts & 4, including urban viaducts, mountainous bridges, railway bridges, and long-span highway bridges \\
Acquisition platform & UAV equipped with DJI H30T \\
Original image resolution & $4032 \times 3024$ pixels \\
Total number of images & 3,767 \\
Train / validation / test split & 2,895 / 495 / 377 \\
Repeated images across subsets & None after duplicate image screening \\
Defect categories & Spalling, Crack, Hole, Honeycombing, Seepage, Rebar Corrosion, Rust, Efflorescence \\
Annotation protocol & Minimum enclosing rectangle and closed polygon contour \\
\bottomrule
\end{tabularx}
\end{table}

The dataset defines eight categories of bridge surface defects: Spalling, Crack, Hole, Honeycombing, Seepage, Rebar Corrosion, Rust, and Efflorescence. All images were manually annotated under a unified annotation protocol. Specifically, each defect instance was assigned two complementary geometric labels: a minimum enclosing rectangle for object-level localization and a closed polygon contour for fine boundary representation. This dual-annotation design establishes a common data basis for detection, segmentation, and FS-FSD.

On top of this unified annotation source, three task-oriented supervision forms were constructed. For detection, the supervision target is the minimum enclosing rectangle, which supports the training of conventional detectors and rectangle-level evaluation. For segmentation, the supervision target is the closed polygon contour, which supports segmentation training and polygon-space boundary representation. For FS-FSD, the supervision target is obtained by deterministically reparameterizing the same closed polygon contour. More specifically, each polygon boundary is first normalized with respect to image size and regularized through duplicate vertex removal, closure consistency correction, and degenerate-contour filtering. A cyclically ordered closed contour sequence is then constructed while preserving boundary traversal continuity, from which the contour center and harmonic coefficient targets required by FS-FSD are derived. The resulting supervision preserves the original geometric semantics of the defect boundary while satisfying the parameterization requirements of Fourier contour learning.

It is important to emphasize that these three supervision forms do not originate from independent annotation systems. Rather, they are derived from the same defect instances under the same boundary semantics. In other words, FS-FSD does not benefit from additional manual annotation information. Its supervision targets are obtained solely through a representation-level remapping of the same polygon boundaries used by the other task formulations. As a result, the comparison among detection, segmentation, and FS-FSD is grounded in identical image sources, identical defect instances, and identical boundary semantics, thereby reducing evaluation bias introduced by supervision-source discrepancies and strengthening the fairness of the benchmark.

\begin{table}[pos=h]
\centering
\caption{Task-oriented supervision forms and class-wise instance distribution of the dataset. Panel A summarizes the geometric supervision used by different learning paradigms, while Panel B reports the class-wise instance distribution of the full dataset.}
\label{tab:supervision_and_distribution}

\fontsize{6.20}{7.00}\selectfont
\setlength{\tabcolsep}{2.6pt}
\renewcommand{\arraystretch}{1.09}
\setlength{\aboverulesep}{0.32ex}
\setlength{\belowrulesep}{0.32ex}

\begin{threeparttable}

\begin{minipage}{\linewidth}
\raggedright\textbf{Panel A. Task-oriented supervision forms}
\end{minipage}

\vspace{1.4pt}

\begin{tabularx}{\linewidth}{@{}>{\RaggedRight\arraybackslash}p{0.19\linewidth}
                            >{\RaggedRight\arraybackslash}p{0.35\linewidth}
                            >{\RaggedRight\arraybackslash}X@{}}
\toprule
\textbf{Supervision type} & \textbf{Geometric representation} & \textbf{Primary use} \\
\midrule
Detection
& Minimum enclosing rectangle
& Training and rectangle-level evaluation of conventional detectors. \\

Segmentation
& Closed polygon contour
& Training of segmentation models and polygon-space boundary representation. \\

\rowcolor{oursbg}
FS-FSD
& Normalized and cyclically ordered closed contour sequence mapped to Fourier parameters
& Frequency-domain contour learning and geometric supervision for FS-FSD. \\
\bottomrule
\end{tabularx}

\vspace{4.2pt}

\begin{minipage}{\linewidth}
\raggedright\textbf{Panel B. Class-wise instance distribution}
\end{minipage}

\vspace{1.4pt}

\begin{tabularx}{\linewidth}{@{}>{\centering\arraybackslash}p{0.13\linewidth}
                            >{\RaggedRight\arraybackslash}p{0.39\linewidth}
                            >{\RaggedLeft\arraybackslash}p{0.20\linewidth}
                            >{\RaggedLeft\arraybackslash}X@{}}
\toprule
\textbf{Class ID} & \textbf{Class name} & \textbf{Instance count} & \textbf{Proportion} \\
\midrule
0 & Spalling         & 9,172  & 21.66\% \\
1 & Crack            & 4,209  & 9.94\% \\
2 & Hole             & 11,663 & 27.54\% \\
3 & Honeycombing     & 1,492  & 3.53\% \\
4 & Seepage          & 1,986  & 4.69\% \\
5 & Rebar corrosion  & 4,511  & 10.65\% \\
6 & Rust             & 2,360  & 5.57\% \\
7 & Efflorescence    & 6,953  & 16.42\% \\
\midrule
& \textbf{Total} & \textbf{42,346} & \textbf{100.00\%} \\
\bottomrule
\end{tabularx}

\vspace{1.4pt}

\begin{tablenotes}[flushleft]
\fontsize{5.20}{5.90}\selectfont
\item \textbf{Note.}
Panel A describes the supervision format used by each task formulation. Panel B reports the instance-level class distribution over the full dataset.
\end{tablenotes}

\end{threeparttable}
\end{table}

All annotations were produced and managed using a self-developed annotation tool. The tool supports unified annotation of axis-aligned boxes, rotated boxes, and polygonal contours, and can export multiple formats compatible with YOLO/Ultralytics workflows. The tool has been publicly released as an open-source project:
\url{https://github.com/wangzai822/YOLO-format-annotation-tool-}

The finalized annotation corpus contains 42,346 defect instances in total. The class-wise counts and proportions are summarized in Table~\ref{tab:supervision_and_distribution}. A clear class imbalance is present: Hole, Spalling, and Efflorescence account for comparatively larger shares, whereas Honeycombing and Seepage are relatively limited. Such a distribution is consistent with real inspection conditions, where different defect types vary substantially in occurrence frequency, visual observability, and collectability.

The unified workflow covering acquisition, annotation, and supervision construction enables rectangle-level detection, polygon-space evaluation, and Fourier contour learning to be conducted on a common data basis. Therefore, comparisons among different models are grounded in the same image sources, defect instances, and geometric semantics.

\subsection{Evaluation protocol and metrics}
\label{sec:evaluation_metrics}

Bridge defect analysis is not only a recognition problem, but also a geometric quality problem. In the present study, conventional detectors, segmentation models, and FS-FSD produce different native outputs, namely bounding boxes, masks or polygonal regions, and parameterized closed contours, respectively. If these models are evaluated only in their own native output spaces, the resulting comparison may be influenced by representation mismatch rather than by the actual geometric quality of the recovered defect boundary. For this reason, the evaluation protocol adopted in this study is organized into three complementary layers: native task metrics, unified polygon-space metrics, and boundary and geometry metrics. This protocol is intended as a representation-aware polygon-space assessment for downstream geometric quality. It complements native task metrics rather than replacing them.

\paragraph{Native task metrics.}
Native task metrics are retained to describe how each model performs in the output space for which it was originally designed. For object detection, the standard metric is bounding-box mean Average Precision (mAP), based on the Intersection over Union (IoU) between a predicted box $\hat{B}$ and a ground-truth box $B$:
\begin{equation}
\mathrm{IoU}_{\mathrm{box}}(\hat{B}, B)
=
\frac{|\hat{B}\cap B|}{|\hat{B}\cup B|}.
\label{eq:eval_iou_box}
\end{equation}
For instance segmentation, the standard metric is mask mAP, based on the overlap between a predicted mask $\hat{M}$ and a ground-truth mask $M$:
\begin{equation}
\mathrm{IoU}_{\mathrm{mask}}(\hat{M}, M)
=
\frac{|\hat{M}\cap M|}{|\hat{M}\cup M|}.
\label{eq:eval_iou_mask}
\end{equation}
These native metrics are reported for reference, but they are not used as the principal criterion for cross-paradigm comparison, because box IoU, mask IoU, and contour-parameter regression errors do not lie in a common geometric space.

\paragraph{Unified polygon-space metrics.}
To compare heterogeneous outputs under common geometric semantics, all predictions are projected into polygon space and evaluated against polygon ground truth. In the main comparison tables, this unified polygon-space assessment is organized into two routes. The Rect-to-Polygon (R2P) route is used for box-producing models, where a predicted bounding box is converted into a rectangular polygon and then compared with the ground-truth polygon. The Shape-to-Polygon (S2P) route is used for models that already produce boundary-bearing shapes. For segmentation models, the native instance-mask output is converted into a polygon through the S2P-Mask route. For contour-native methods, including Deep Snake, FSD, and FS-FSD, the predicted contour or Fourier contour is reconstructed and discretized into a polygon through the S2P-Contour route before comparison. Detailed implementation-level information on the mask-to-polygon conversion, including the YOLO-seg and Mask2Former interfaces and the corresponding route assignment for boundary and geometry metrics, is provided in Appendix~\ref{sec:mask_to_polygon_yolo_mask2former}.

Let $\hat{P}$ denote the polygon obtained after route-specific representation conversion, and let $P$ denote the ground-truth polygon. The unified polygon IoU is defined as
\begin{equation}
\mathrm{IoU}_{\mathrm{poly}}(\hat{P}, P)
=
\frac{|\hat{P}\cap P|}{|\hat{P}\cup P|}.
\label{eq:eval_iou_poly}
\end{equation}
Formally, $\hat{P}$ is obtained as
\begin{equation}
\hat{P}
=
\begin{cases}
\mathcal{C}_{\mathrm{R2P}}(\hat{B}), & \text{for box-producing models},\\[4pt]
\mathcal{C}_{\mathrm{S2P}}(\hat{S}), & \text{for mask- or contour-producing models},
\end{cases}
\label{eq:eval_route_conversion}
\end{equation}
where $\hat{B}$ is the predicted bounding box, $\hat{S}$ is a native shape output such as a mask, polygon, explicit contour, or Fourier contour, and $\mathcal{C}_{\mathrm{R2P}}(\cdot)$ and $\mathcal{C}_{\mathrm{S2P}}(\cdot)$ are deterministic route-specific conversion operators. In practice, $\mathcal{C}_{\mathrm{S2P}}(\cdot)$ is instantiated as mask-to-polygon conversion for segmentation models and contour or Fourier-to-polygon conversion for contour-native models. The detailed mask-to-polygon implementation is not repeated in the main text, because it is specified in Appendix~\ref{sec:mask_to_polygon_yolo_mask2former}.

\begin{figure}[pos=h]
    \centering
    \includegraphics[width=0.70\linewidth]{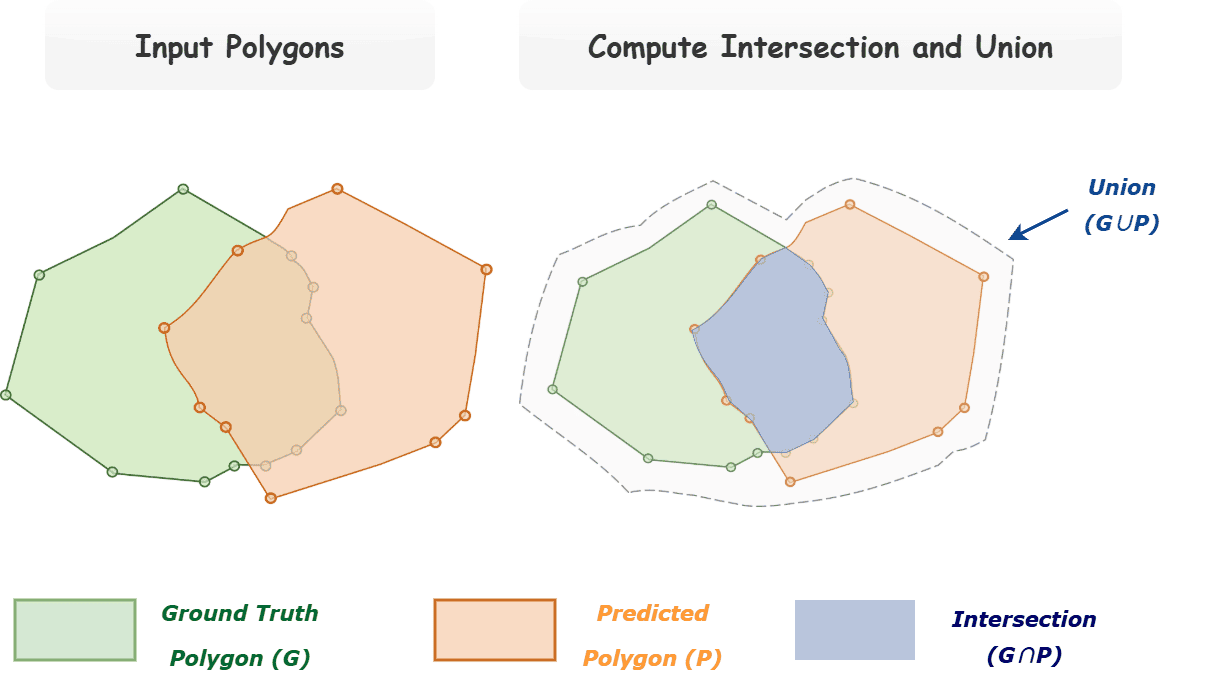}
    \caption{Polygon-space IoU computation. The predicted output, regardless of its native representation, is first converted into polygon space and then compared with the polygon ground truth through intersection and union areas.}
    \label{fig:polygon_iou}
\end{figure}

For FS-FSD, the Fourier contour is first reconstructed at a fixed sampling density $T$ as
\begin{equation}
\hat{\mathcal C}
=
\left\{
\bigl(x(t_s),y(t_s)\bigr)
\right\}_{s=0}^{T-1},
\qquad
t_s=\frac{2\pi s}{T},
\label{eq:eval_fourier_sampling}
\end{equation}
and the resulting closed point sequence is treated as the polygon $\hat{P}$ for subsequent geometric evaluation. In the reported experiments, the polygonization density for Fourier contours is fixed to $T=256$. As illustrated in Fig.~\ref{fig:polygon_iou}, polygon IoU is computed directly from the intersection and union areas between the predicted and ground-truth polygons after route-specific conversion.

For each class $c$, predictions are sorted in descending order of confidence and greedily matched to unmatched ground-truth polygons of the same class under a polygon IoU threshold $\tau$. The corresponding Average Precision (AP) is denoted by $AP_{c}^{r,\tau}$, where $r\in\{\mathrm{R2P},\mathrm{S2P}\}$ denotes the evaluation route. On this basis, the polygon-space metrics are defined as
\begin{equation}
\mathrm{mAP}^{r}_{\mathrm{poly}@50}
=
\frac{1}{C}
\sum_{c=1}^{C}
AP_{c}^{r,0.50},
\label{eq:eval_map_polygon_50}
\end{equation}
and
\begin{equation}
\mathrm{mAP}^{r}_{\mathrm{poly}@50:95}
=
\frac{1}{|\mathcal{T}|C}
\sum_{\tau\in\mathcal{T}}
\sum_{c=1}^{C}
AP_{c}^{r,\tau},
\qquad
\mathcal{T}=\{0.50,0.55,\ldots,0.95\},
\label{eq:eval_map_polygon_5095}
\end{equation}
where $C$ is the number of valid categories and categories without ground-truth instances are excluded from averaging. Under this formulation, R2P measures how well a rectangular support region approximates the actual defect boundary in polygon space, whereas S2P measures how well a boundary-bearing output recovers that boundary geometry.

This distinction is important for the present study. FS-FSD is designed not merely to indicate where a defect may be located, but to produce a contour-native representation for downstream geometric use. Native box or mask mAP alone cannot fully reveal the influence of output representation on boundary recovery. By contrast, R2P and S2P compare heterogeneous predictions in a common polygon space and therefore provide a more direct assessment of boundary quality.

\paragraph{Boundary and geometry metrics.}
Polygon-space mAP evaluates detection and matching quality in a unified geometric domain, but it is still insufficient to characterize local boundary adherence and global geometric consistency. For this reason, four additional boundary and geometry metrics are introduced: Boundary F-score (B-F1), Chamfer Distance (CD), Perimeter Error (P-Err), and Area Error (A-Err).

These metrics are computed only on matched true-positive pairs. Specifically, a prediction and a ground-truth polygon form a valid evaluation pair only if they belong to the same class and satisfy $\mathrm{IoU}_{\mathrm{poly}}\ge \tau_q$, where $\tau_q$ is a quality-matching threshold set to 0.50 in the experiments. All geometric distances are computed in the normalized image coordinate system $[0,1]^2$, which eliminates resolution-induced scale bias.

In the unified polygon-space evaluation adopted in this study, a distinction is made between the evaluation polygon geometry and the boundary sampling set. The former is used for overlap computation as well as for area- and perimeter-related quantities, whereas the latter is used only for local boundary-quality assessment. This distinction is important because, if all geometric quantities were computed directly from the original vertex list or from a fixed number of sampled points, the resulting metrics would become unnecessarily sensitive to route-specific discretization density, thereby weakening the stability and interpretability of cross-paradigm comparison.

More specifically, let the predicted and ground-truth polygons after route-specific conversion be denoted by

\[
\hat{P}=\{(\hat{x}_i,\hat{y}_i)\}_{i=1}^{K_{\hat{P}}},
\qquad
P=\{(x_i,y_i)\}_{i=1}^{K_P},
\]
where the vertices are ordered along the closed boundary, and $K_{\hat{P}}$ and $K_P$ need not be the same. For box-producing models, $\hat{P}$ is obtained by converting the predicted bounding box into a rectangular polygon. For segmentation models, $\hat{P}$ is extracted from the predicted mask boundary. For contour-native methods, including FS-FSD, the predicted explicit contour or reconstructed Fourier contour is discretized into a polygon before entering polygon-space evaluation. For notational simplicity, the definitions below are written for a single closed polygon. In implementation, if the evaluated object consists of multiple merged polygonal components, the same geometric quantities are computed on the resulting polygonal geometry by the geometry engine.

On this basis, both area and perimeter are computed from the final polygon geometry itself, rather than from the boundary resampling set. For an arbitrary valid polygon $P$, its perimeter is defined as the sum of Euclidean distances between consecutive vertices:
\begin{equation}
L(P)
=
\sum_{i=1}^{K_P}
\sqrt{
(x_{i+1}-x_i)^2+(y_{i+1}-y_i)^2
},
\qquad
(x_{K_P+1},y_{K_P+1})=(x_1,y_1).
\label{eq:eval_polygon_perimeter}
\end{equation}

Its area is computed by the shoelace formula:
\begin{equation}
A(P)
=
\frac{1}{2}
\left|
\sum_{i=1}^{K_P}
\left(
x_i y_{i+1}-x_{i+1} y_i
\right)
\right|,
\qquad
(x_{K_P+1},y_{K_P+1})=(x_1,y_1).
\label{eq:eval_polygon_area}
\end{equation}

The same definitions are applied to the predicted polygon $\hat{P}$, yielding $L(\hat{P})$ and $A(\hat{P})$. The relative absolute perimeter error and relative absolute area error are then defined as
\begin{equation}
\mathrm{P\text{-}Err}
=
\frac{|L(\hat{P})-L(P)|}{L(P)},
\qquad
\mathrm{A\text{-}Err}
=
\frac{|A(\hat{P})-A(P)|}{A(P)}.
\label{eq:eval_geom_errors}
\end{equation}
Degenerate ground-truth polygons with zero area or zero perimeter are excluded during annotation regularization and evaluation preparation, so the denominators in Eq.~\eqref{eq:eval_geom_errors} are well-defined.

It is important to emphasize that P-Err and A-Err evaluate the recovered polygon geometry itself. Accordingly, all methods are compared on the same geometric carrier, and FS-FSD does not receive any additional evaluation privilege from its Fourier parameterization. This definition is consistent with the cross-paradigm polygon-space benchmark adopted in this study and can support downstream image-space geometric summarization, archival, recovery, visualization, and comparison. Conversion to physical units would require external scale calibration or spatial registration and is not assumed in this evaluation.

By contrast, B-F1 and CD are intended to characterize local boundary adherence. For this reason, polygon boundaries are further resampled uniformly with respect to arc length. Let $\partial P$ denote the boundary of polygon $P$. The number of boundary samples is defined as
\begin{equation}
N(P)
=
\min\!\left(
N_{\max},
\max\!\left(
N_{\min},
\left\lceil \frac{L(P)}{\Delta s}\right\rceil
\right)
\right),
\label{eq:eval_boundary_sample_count}
\end{equation}
where $\Delta s$ is the arc-length sampling step, and $N_{\min}$ and $N_{\max}$ are the lower and upper bounds on the number of sampled points. The resulting uniformly sampled boundary point sets are denoted by $S_P$ and $S_{\hat{P}}$ for the ground-truth and predicted polygons, respectively. In the reported experiments,

\[
\Delta s = 0.002,\qquad N_{\min}=32,\qquad N_{\max}=512.
\]
Hence, the boundary sampling size is not fixed at 512 points. Instead, it varies adaptively with polygon perimeter and is capped at 512 to control computation on long boundaries.

Given a boundary tolerance $\epsilon_b$, boundary precision and boundary recall are defined as
\begin{equation}
\mathrm{B\text{-}Prec}
=
\frac{1}{|S_{\hat{P}}|}
\sum_{x\in S_{\hat{P}}}
\mathbf{1}
\left(
\min_{y\in S_P}\|x-y\|_2 \le \epsilon_b
\right),
\label{eq:eval_boundary_precision}
\end{equation}
\begin{equation}
\mathrm{B\text{-}Rec}
=
\frac{1}{|S_P|}
\sum_{y\in S_P}
\mathbf{1}
\left(
\min_{x\in S_{\hat{P}}}\|x-y\|_2 \le \epsilon_b
\right).
\label{eq:eval_boundary_recall}
\end{equation}
The corresponding boundary F-score is
\begin{equation}
\mathrm{B\text{-}F1}
=
\frac{2\,\mathrm{B\text{-}Prec}\,\mathrm{B\text{-}Rec}}
{\mathrm{B\text{-}Prec}+\mathrm{B\text{-}Rec}}.
\label{eq:eval_boundary_f1}
\end{equation}
When both $\mathrm{B\text{-}Prec}$ and $\mathrm{B\text{-}Rec}$ are zero, $\mathrm{B\text{-}F1}$ is set to zero.

Chamfer Distance is defined as
\begin{equation}
\mathrm{CD}
=
\frac{1}{2}
\left(
\frac{1}{|S_{\hat{P}}|}
\sum_{x\in S_{\hat{P}}}
\min_{y\in S_P}\|x-y\|_2
+
\frac{1}{|S_P|}
\sum_{y\in S_P}
\min_{x\in S_{\hat{P}}}\|x-y\|_2
\right).
\label{eq:eval_chamfer_distance}
\end{equation}

In this way, B-F1 and CD are computed from the boundary resampling sets, whereas P-Err and A-Err are computed from the polygon geometry itself. This separation of roles ensures stable local-boundary assessment and fair global-geometry assessment across different output paradigms.

All polygon-based geometric computations in this study are performed in the normalized image coordinate system $[0,1]^2$. Accordingly, $\epsilon_b$ is also expressed in normalized units. Under the default resized input resolution of $896\times896$, the reported setting $\epsilon_b = 0.002222$ corresponds to a boundary tolerance of approximately 2 pixels, whereas $\Delta s = 0.002$ corresponds to an arc-length sampling interval of approximately 1.8 pixels. This setting makes the local boundary-quality metrics consistent with practically meaningful visual deviations while preserving a unified basis for cross-model comparison.

\paragraph{Interpretation in the context of this study.}
The three-layer evaluation protocol plays distinct but complementary roles in the present work. Native task metrics preserve standard benchmark compatibility for detection and segmentation models. Unified polygon-space metrics provide a representation-aware geometric view that places boxes, masks, and contour-native outputs under common boundary semantics. Boundary and geometry metrics further quantify whether the recovered boundary has sufficient image-space geometric fidelity for structured record generation, archival, recovery, visualization, and comparison.

In this sense, the protocol is aligned with the central objective of FS-FSD. The proposed method is not intended merely to improve image-level recognition, but to produce a more reusable geometric representation for bridge inspection information workflows. A method may achieve strong native box or mask mAP while still exhibiting inferior polygon-space fidelity or larger geometric inconsistency. The evaluation protocol adopted here is therefore designed to expose not only recognition performance, but also the downstream geometric suitability of the predicted defect representation. Physical-scale measurement, component-level localization, and maintenance decision support require additional calibration, registration, and asset-information integration, and are not assumed as completed capabilities in this evaluation.

For compact display in the result tables, B-F1, P-Err, and A-Err are multiplied by $10^2$, whereas CD is multiplied by $10^3$; all actual computations are performed on the unscaled values.

\subsection{Training configuration and hyperparameter settings}
\label{sec:hyperparameters}

For reproducibility, the principal training settings used for all compared methods are summarized in Table~\ref{tab:training_configs}. The table reports the resized input resolution, batch size, epoch budget, optimizer, initial learning rate, momentum or $\beta_1$ for Adam-style optimizers, weight decay, and automatic mixed precision (AMP) setting. To reduce schedule-induced variation across heterogeneous frameworks, all methods were trained on the same train/validation split under a unified batch size of 16 and an equivalent training budget of 500 epochs. For iteration-based frameworks, such as the Detectron2-based Mask2Former implementation, the 500-epoch budget was converted into the corresponding maximum number of optimizer updates according to the training-set size and batch setting.

Stable framework-native optimizer choices were retained when appropriate, so that the benchmark did not depend on extensive method-specific hyperparameter search. Therefore, the purpose of this setting is to compare different representation paradigms under a controlled and reproducible training budget, rather than to maximize each baseline through exhaustive schedule tuning.

\begin{table*}[pos=h]
\centering
\caption{Principal training settings of the compared methods on the proposed dataset.}
\label{tab:training_configs}

\fontsize{5.78}{6.52}\selectfont
\setlength{\tabcolsep}{2.1pt}
\renewcommand{\arraystretch}{1.08}
\setlength{\aboverulesep}{0.28ex}
\setlength{\belowrulesep}{0.28ex}
\setlength{\cmidrulesep}{0.10ex}

\begin{threeparttable}
\begin{tabularx}{0.92\textwidth}{@{}>{\RaggedRight\arraybackslash}X
                                >{\centering\arraybackslash}p{0.095\textwidth}
                                >{\centering\arraybackslash}p{0.045\textwidth}
                                >{\centering\arraybackslash}p{0.055\textwidth}
                                >{\centering\arraybackslash}p{0.070\textwidth}
                                >{\centering\arraybackslash}p{0.060\textwidth}
                                >{\centering\arraybackslash}p{0.072\textwidth}
                                >{\centering\arraybackslash}p{0.068\textwidth}
                                >{\centering\arraybackslash}p{0.045\textwidth}@{}}
\toprule
\multirow{2}{*}{Method} &
\multicolumn{3}{c}{Training schedule} &
\multicolumn{5}{c}{Optimization} \\
\cmidrule(lr){2-4}\cmidrule(lr){5-9}
& Input & Batch & Epochs & Optim. & $LR_0$ & Mom./$\beta_1$ & WD & AMP \\
\midrule

\routebox\hspace{0.25em}Faster R-CNN~\cite{ren2015faster}
& $896\times896$ & 16 & 500 & SGD & 0.001 & 0.9 & $1\times10^{-4}$ & \na \\

\routebox\hspace{0.25em}RT-DETR-R50~\cite{zhao2024detrs}
& $896\times896$ & 16 & 500 & AdamW & $1\times10^{-4}$ & 0.9 & $1\times10^{-4}$ & \na \\

\routemask\hspace{0.25em}YOLOv5m-seg~\cite{jocher2022ultralytics}
& $896\times896$ & 16 & 500 & SGD & 0.01 & 0.937 & 0.0005 & \na \\

\routemask\hspace{0.25em}YOLO11m-seg~\cite{khanam2024yolov11}
& $896\times896$ & 16 & 500 & auto & 0.01 & 0.937 & 0.0005 & True \\

\routemask\hspace{0.25em}YOLO26m-seg~\cite{sapkota2025yolo26}
& $896\times896$ & 16 & 500 & auto & 0.01 & 0.937 & 0.0005 & True \\

\routemask\hspace{0.25em}Mask2Former~\cite{cheng2022masked}
& $896\times896$ & 16 & 500 & AdamW & $1\times10^{-4}$ & 0.9 & 0.05 & False \\

\routecont\hspace{0.25em}Deep Snake~\cite{peng2020deep}
& $896\times896$ & 16 & 500 & Adam & $1\times10^{-4}$ & 0.9 & 0 & \na \\

\routecont\hspace{0.25em}FSD~\cite{liu2025fourierseries}
& $896\times896$ & 16 & 500 & auto & 0.01 & 0.9 & 0.0005 & auto \\

\midrule
\rowcolor{oursbg}
\routecont\hspace{0.25em}FS-FSD
& $896\times896$ & 16 & 500 & auto & 0.01 & 0.9 & 0.0005 & auto \\
\bottomrule
\end{tabularx}

\vspace{1.2pt}

\begin{tablenotes}[flushleft]
\fontsize{5.12}{5.86}\selectfont
\item Optim. = optimizer; $LR_0$ = initial learning rate; Mom./$\beta_1$ = SGD momentum or Adam-style $\beta_1$; WD = weight decay; AMP = automatic mixed precision. For Detectron2-based Mask2Former, the epoch value denotes the equivalent epoch budget after converting the iteration-based training schedule. The AMP field is marked as False because mixed-precision training was not enabled in the adopted Detectron2 configuration.
\end{tablenotes}
\end{threeparttable}
\end{table*}

As shown in Table~\ref{tab:training_configs}, Faster R-CNN adopts SGD-based optimization, RT-DETR-R50 uses AdamW, Deep Snake follows an Adam-based setting, and the YOLO-style baselines together with FSD and FS-FSD use framework-provided automatic optimizer selection. Mask2Former~\cite{cheng2022masked} is trained using the Detectron2-based Mask2Former framework with AdamW optimization. All models use the same resized input resolution of $896\times896$, the same batch size, and the same equivalent epoch budget, which reduces schedule-related variation across different implementations.

FSD and FS-FSD are trained under identical input resolution, batch size, epoch budget, optimizer type, initial learning rate, momentum setting, weight decay, and AMP setting. This controlled configuration reduces training-schedule confounding when comparing the two contour-native methods. Therefore, the performance differences reported in the following sections should be interpreted mainly in relation to the proposed contour-native formulation and its frequency-domain learning strategy, rather than as a consequence of a more favorable optimization schedule.

\subsection{Quantitative comparison with baseline methods}
\label{subsec:comparison_baselines}

\begin{table*}[pos=h]
\centering
\caption{Overall quantitative comparison across native-space metrics, unified polygon-space mean Average Precision (mAP), and shape-output matched true-positive (matched-TP) boundary and geometry metrics.}
\label{tab:overall_main}

\fontsize{5.35}{6.18}\selectfont
\setlength{\tabcolsep}{0.92pt}
\renewcommand{\arraystretch}{1.12}
\setlength{\aboverulesep}{0.34ex}
\setlength{\belowrulesep}{0.34ex}
\setlength{\cmidrulesep}{0.14ex}

\begin{threeparttable}
\begin{tabular*}{\textwidth}{@{\extracolsep{\fill}}lccccccccccc@{}}
\toprule
\multirow{2}{*}{Method} &
\multirow{2}{*}{\makecell[c]{Params\\[-1pt](M)}} &
\multirow{2}{*}{\makecell[c]{Model Size\\[-1pt](MB)}} &
\multirow{2}{*}{\makecell[c]{GFLOPs\\[-1pt]896$\times$896}} &
\multicolumn{2}{c}{\makecell[c]{Native-space\\[-0.4pt]Metrics}} &
\multicolumn{2}{c}{\makecell[c]{Unified Polygon-space\\[-0.4pt]mAP}} &
\multicolumn{2}{c}{\makecell[c]{S2P Matched-TP\\[-0.4pt]Boundary Quality}} &
\multicolumn{2}{c}{\makecell[c]{S2P Matched-TP\\[-0.4pt]Geometry Error}} \\
\cmidrule(lr){5-6}\cmidrule(lr){7-8}\cmidrule(lr){9-10}\cmidrule(lr){11-12}
& & & &
\makecell[c]{Native Box mAP\\[-0.4pt]$\mathrm{mAP}_{50}/\mathrm{mAP}_{50{:}95}\,\uparrow$} &
\makecell[c]{Native Mask mAP\\[-0.4pt]$\mathrm{mAP}_{50}/\mathrm{mAP}_{50{:}95}\,\uparrow$} &
\makecell[c]{R2P Polygon mAP\\[-0.4pt]$\mathrm{mAP}_{50}/\mathrm{mAP}_{50{:}95}\,\uparrow$} &
\makecell[c]{S2P Polygon mAP\\[-0.4pt]$\mathrm{mAP}_{50}/\mathrm{mAP}_{50{:}95}\,\uparrow$} &
\makecell[c]{B-F1\\[-0.4pt]$(\times10^2)\,\uparrow$} &
\makecell[c]{CD\\[-0.4pt]$(\times10^3)\,\downarrow$} &
\makecell[c]{P-Err\\[-0.4pt]$(\times10^2)\,\downarrow$} &
\makecell[c]{A-Err\\[-0.4pt]$(\times10^2)\,\downarrow$} \\
\midrule

\makecell[l]{\routebox\hspace{0.35em}Faster R-CNN~\cite{ren2015faster}}
& 91.76
& 175.01
& 408.60
& 65.12 / 40.15
& \na
& 59.87 / 33.52
& \na
& \na
& \na
& \na
& \na \\
\addlinespace[1.9pt]

\makecell[l]{\routemask\hspace{0.35em}YOLOv5m-seg~\cite{jocher2022ultralytics}}
& 21.68
& 41.98
& 137.04
& 67.64 / 47.68
& 38.41 / 15.28
& 58.64 / 31.09
& 43.99 / 18.95
& 30.31
& 9.35
& 20.51
& 22.71 \\
\addlinespace[1.9pt]

\makecell[l]{\routemask\hspace{0.35em}YOLO11m-seg~\cite{khanam2024yolov11}}
& 22.37
& 43.16
& 222.55
& 70.89 / 50.56
& 46.86 / 19.09
& 61.64 / 34.92
& 44.92 / 23.75
& 37.85
& 6.11
& 16.31
& 18.87 \\
\addlinespace[1.9pt]

\makecell[l]{\routemask\hspace{0.35em}YOLO26m-seg~\cite{sapkota2025yolo26}}
& 26.98
& 52.03
& 258.57
& 67.15 / 43.92
& 40.08 / 15.01
& 63.19 / 31.33
& 49.68 / 24.16
& 31.20
& 9.58
& 21.13
& 22.97 \\
\addlinespace[1.9pt]

\makecell[l]{\routemask\hspace{0.35em}Mask2Former-R50~\cite{cheng2022masked}}
& 44.00
& 83.92
& 217.40
& 27.83 / 13.03
& 27.12 / 11.02
& 22.30 / 8.76
& 26.98 / 10.77
& 25.55
& 15.61
& 17.76
& 27.02 \\
\addlinespace[1.9pt]

\makecell[l]{\routebox\hspace{0.35em}RT-DETR-R50~\cite{zhao2024detrs}}
& 112.57
& 214.71
& 1156.80
& 74.23 / 50.87
& \na
& 63.46 / 35.78
& \na
& \na
& \na
& \na
& \na \\
\addlinespace[1.9pt]

\makecell[l]{\routecont\hspace{0.35em}Deep Snake~\cite{peng2020deep}}
& 27.05
& 51.59
& 304.08
& 24.93 / 10.97
& 23.77 / 9.13
& 22.88 / 8.85
& 23.55 / 8.92
& 27.05
& 12.42
& 13.44
& 25.77 \\
\addlinespace[1.9pt]

\makecell[l]{\routecont\hspace{0.35em}FSD~\cite{liu2025fourierseries}}
& 87.29
& 166.50
& 395.60
& \na
& \na
& 67.64 / 30.65
& 69.45 / 40.16
& 41.80
& 5.52
& 8.87
& 16.23 \\
\midrule

\rowcolor{oursbg}
\makecell[l]{\routecont\hspace{0.35em}\textbf{FS-FSD}}
& 87.32
& 166.55
& 396.30
& \na
& \na
& \bestnum{77.91} / \bestnum{36.94}
& \shapebestnum{86.93} / \shapebestnum{55.08}
& \geombestnum{48.85}
& \geombestnum{4.23}
& \geombestnum{7.21}
& \geombestnum{12.32} \\
\bottomrule
\end{tabular*}

\vspace{1.8pt}

\begin{tablenotes}[flushleft]
\fontsize{5.15}{5.95}\selectfont
\item Params denotes trainable parameters, and GFLOPs denotes billion floating-point operations. R2P converts boxes to rectangular polygons, whereas S2P converts shape outputs to polygons. Matched-TP columns report S2P shape-output quality only. \na denotes unavailable entries; B-F1, P-Err, and A-Err are in $\times10^2$, and CD is in $\times10^3$.
\end{tablenotes}
\end{threeparttable}
\end{table*}

Tables~\ref{tab:overall_main}--\ref{tab:perclass_aerr} summarize the quantitative comparison between FS-FSD, the baseline FSD, and representative box-based detection, mask-based segmentation, and contour-based methods. The compared baselines include Faster R-CNN~\cite{ren2015faster}, RT-DETR-R50~\cite{zhao2024detrs}, YOLOv5m-seg~\cite{jocher2022ultralytics}, YOLO11m-seg~\cite{khanam2024yolov11}, YOLO26m-seg~\cite{sapkota2025yolo26}, Mask2Former-R50~\cite{cheng2022masked}, and Deep Snake~\cite{peng2020deep}. Native-space metrics are retained for reference, whereas the main comparison is conducted under the unified polygon-space evaluation protocol defined in Section~\ref{sec:evaluation_metrics}. Under this protocol, box-producing methods are evaluated through the Rect-to-Polygon (R2P) route, whereas boundary-bearing outputs are evaluated through the Shape-to-Polygon (S2P) route. Specifically, YOLO segmentation variants and Mask2Former-R50 use S2P-Mask, Deep Snake uses S2P-Contour, and FSD and FS-FSD use S2P-Fourier. The detailed mask-to-polygon conversion rules and route assignments are given in Appendix~\ref{sec:mask_to_polygon_yolo_mask2former}. Therefore, the mask-based baselines are evaluated from their own boundary-bearing outputs rather than being reduced to rectangular boxes.

For matched-TP boundary and geometry metrics, we report only S2P shape-output quality. This design avoids directly comparing rectangle-derived polygons from bounding boxes with true boundary-bearing shape outputs. Accordingly, box-only detectors such as Faster R-CNN and RT-DETR-R50 are marked as \na in the matched-TP columns of Table~\ref{tab:overall_main} and are omitted from Tables~\ref{tab:perclass_bf1}--\ref{tab:perclass_aerr}. This ensures that B-F1, CD, P-Err, and A-Err reflect the quality of recovered shapes rather than the regular geometry of rectangular detections.

Table~\ref{tab:overall_main} reports native-space metrics, unified polygon-space mean Average Precision (mAP), and S2P matched true-positive (matched-TP) boundary and geometry metrics. Table~\ref{tab:perclass_ap} further decomposes AP by defect category and evaluation route, whereas Tables~\ref{tab:perclass_bf1}--\ref{tab:perclass_aerr} report class-wise S2P matched-TP shape quality. In the tables, blue highlights denote the best R2P values, orange highlights denote the best S2P values, lavender highlights denote the best matched-TP boundary or geometry values, and green row shading marks FS-FSD.

At the overall level, FS-FSD achieves the strongest polygon-space mAP in both principal evaluation routes. In the R2P route, the box branch of FS-FSD reaches \textbf{77.91/36.94}, outperforming all box-derived baselines. Among the box-only baselines, RT-DETR-R50 provides the strongest R2P result at \textbf{63.46/35.78}. FS-FSD improves over RT-DETR-R50 by \textbf{14.45} points in $\mathrm{mAP}_{50}$ and by \textbf{1.16} points in $\mathrm{mAP}_{50:95}$. Compared with the baseline FSD (\textbf{67.64/30.65}), FS-FSD further improves the R2P result by \textbf{10.27} points in $\mathrm{mAP}_{50}$ and by \textbf{6.29} points in $\mathrm{mAP}_{50:95}$. This result indicates that the benefit of FS-FSD is not restricted to the final shape output. Its contour-oriented formulation is also associated with improved geometric localization in the box branch, which leads to better consistency under polygon-space evaluation.

\begin{table*}[pos=h]
\centering
\caption{Mean and class-wise Average Precision (AP) under native-space and unified polygon-space evaluations.}
\label{tab:perclass_ap}

\fontsize{5.65}{6.55}\selectfont
\setlength{\tabcolsep}{0.95pt}
\renewcommand{\arraystretch}{1.18}
\setlength{\aboverulesep}{0.40ex}
\setlength{\belowrulesep}{0.40ex}
\setlength{\cmidrulesep}{0.16ex}

\begin{threeparttable}
\begin{tabularx}{\textwidth}{@{}>{\RaggedRight\arraybackslash}p{0.158\textwidth}
                                >{\RaggedRight\arraybackslash}p{0.088\textwidth}
                                *{9}{>{\centering\arraybackslash}X}@{}}
\toprule
\multirow{2}{*}{Method} &
\multirow{2}{*}{\makecell[c]{Eval.\\[-1pt]Space / Route}} &
\multicolumn{9}{c}{\makecell[c]{Mean / class-wise AP\\[-0.4pt]$AP_{50}/AP_{50{:}95}$ (\%) $\uparrow$}} \\
\cmidrule(lr){3-11}
& &
mAP &
Spall. &
Crack &
Hole &
\makecell[c]{Honey-\\comb.} &
Seepage &
\makecell[c]{Rebar\\Corr.} &
Rust &
Efflo. \\
\midrule

\multirow{2}{*}{\makecell[l]{\routebox\hspace{0.35em}Faster R-CNN~\cite{ren2015faster}}}
& Native Box
& 65.12 / 40.15
& 73.16 / 43.50
& 64.05 / 30.85
& 57.36 / 37.58
& 74.39 / 55.83
& 71.72 / 49.34
& 55.32 / 28.59
& 66.01 / 40.54
& 58.95 / 34.97 \\
& R2P
& 59.87 / 33.52
& 64.63 / 37.13
& 57.69 / 26.60
& 53.27 / 31.55
& 68.75 / 44.70
& 66.69 / 40.54
& 49.89 / 23.14
& 61.00 / 35.44
& 57.03 / 29.07 \\
\midrule

\multirow{4}{*}{\makecell[l]{\routemask\hspace{0.35em}YOLOv5m-seg~\cite{jocher2022ultralytics}}}
& Native Box
& 67.64 / 47.68
& 74.80 / 51.50
& 67.40 / 46.00
& 81.30 / 52.70
& 60.30 / 45.70
& 57.10 / 44.10
& 78.60 / 49.80
& 64.80 / 46.90
& 56.80 / 44.90 \\
& Native Mask
& 38.41 / 15.28
& 56.90 / 21.10
& 38.80 / 12.50
& 51.40 / 23.10
& 31.50 / 17.80
& 21.70 / 9.83
& 49.60 / 16.00
& 30.50 / 11.30
& 26.90 / 10.60 \\
& R2P
& 58.64 / 31.09
& 64.21 / 27.47
& 48.58 / 25.06
& 71.26 / 32.11
& 55.35 / 37.43
& 52.91 / 34.29
& 67.93 / 29.92
& 58.00 / 31.09
& 50.90 / 31.34 \\
& S2P-Mask
& 43.99 / 18.95
& 56.72 / 22.48
& 42.21 / 14.74
& 68.64 / 31.24
& 42.89 / 24.66
& 24.68 / 12.32
& 59.14 / 22.47
& 33.46 / 13.96
& 24.16 / 9.70 \\
\midrule

\multirow{4}{*}{\makecell[l]{\routemask\hspace{0.35em}YOLO11m-seg~\cite{khanam2024yolov11}}}
& Native Box
& 70.89 / 50.56
& 82.40 / 57.90
& 75.00 / 53.90
& 85.10 / 53.50
& 56.00 / 45.20
& 58.30 / 46.10
& 87.10 / 55.10
& 64.60 / 47.20
& 58.70 / 45.60 \\
& Native Mask
& 46.86 / 19.09
& 67.10 / 28.80
& 50.30 / 18.80
& 60.90 / 26.50
& 26.40 / 10.70
& 28.10 / 10.10
& 67.90 / 26.70
& 37.70 / 14.00
& 36.60 / 17.10 \\
& R2P
& 61.64 / 34.92
& 73.57 / 37.55
& 54.60 / \bestnum{29.78}
& 80.07 / 40.74
& 51.91 / 39.38
& 51.46 / 35.01
& \bestnum{76.66} / \bestnum{34.92}
& 56.95 / 32.03
& 47.90 / 29.97 \\
& S2P-Mask
& 44.92 / 23.75
& 64.33 / 36.95
& 45.83 / 18.32
& 75.77 / 43.31
& 16.86 / 10.35
& 23.30 / 11.15
& 73.27 / \shapebestnum{40.07}
& 29.92 / 14.97
& 30.10 / 14.86 \\
\midrule

\multirow{4}{*}{\makecell[l]{\routemask\hspace{0.35em}YOLO26m-seg~\cite{sapkota2025yolo26}}}
& Native Box
& 67.15 / 43.92
& 75.70 / 46.60
& 65.60 / 42.30
& 81.80 / 46.30
& 52.40 / 38.40
& 60.40 / 44.60
& 79.10 / 46.20
& 64.20 / 44.50
& 58.10 / 42.50 \\
& Native Mask
& 40.08 / 15.01
& 51.40 / 20.20
& 39.30 / 13.80
& 52.00 / 20.30
& 29.00 / 8.59
& 28.30 / 10.40
& 56.20 / 20.00
& 27.50 / 10.80
& 36.90 / 16.00 \\
& R2P
& 63.19 / 31.33
& 70.98 / 30.53
& 46.90 / 20.37
& 80.38 / 38.10
& 65.64 / 35.40
& 57.49 / 33.24
& 71.14 / 30.50
& 59.76 / 33.16
& 53.24 / 29.36 \\
& S2P-Mask
& 49.68 / 24.16
& 63.47 / 31.21
& 45.42 / 18.61
& \shapebestnum{76.21} / 40.14
& 33.01 / 15.14
& 39.77 / 19.04
& 67.47 / 33.38
& 33.85 / 16.90
& 38.22 / 18.83 \\
\midrule

\multirow{2}{*}{\makecell[l]{\routemask\hspace{0.35em}Mask2Former-R50~\cite{cheng2022masked}}}
& R2P
& 22.30 / 8.76
& 11.46 / 3.71
& 13.83 / 5.28
& 50.27 / 21.37
& 19.20 / 9.43
& 19.80 / 7.39
& 24.51 / 6.53
& 11.99 / 4.85
& 27.37 / 11.48 \\
& S2P-Mask
& 26.98 / 10.77
& 13.07 / 4.20
& 23.02 / 7.70
& 50.88 / 23.20
& 20.60 / 10.09
& 27.85 / 10.92
& 25.17 / 7.41
& 17.73 / 5.60
& 37.53 / 17.01 \\
\midrule

\multirow{2}{*}{\makecell[l]{\routebox\hspace{0.35em}RT-DETR-R50~\cite{zhao2024detrs}}}
& Native Box
& 74.23 / 50.87
& 80.96 / 53.76
& 71.78 / 37.92
& 65.05 / 45.98
& 84.56 / 68.29
& 82.43 / 62.38
& 62.31 / 38.21
& 75.87 / 55.36
& 70.89 / 45.07 \\
& R2P
& 63.46 / 35.78
& 71.03 / 37.61
& 63.39 / 25.69
& 57.08 / 31.56
& 72.87 / 47.54
& 71.04 / \bestnum{44.78}
& 52.68 / 26.41
& 62.12 / \bestnum{39.88}
& \bestnum{57.47} / \bestnum{32.76} \\
\midrule

\multirow{2}{*}{\makecell[l]{\routecont\hspace{0.35em}Deep Snake~\cite{peng2020deep}}}
& R2P
& 22.88 / 8.85
& 21.25 / 6.56
& 11.21 / 3.68
& 37.18 / 13.34
& 17.41 / 8.30
& 23.79 / 11.80
& 21.14 / 5.91
& 16.77 / 5.82
& 34.32 / 15.35 \\
& S2P-Contour
& 23.55 / 8.92
& 21.92 / 7.38
& 10.13 / 3.37
& 38.21 / 15.54
& 17.44 / 8.25
& 25.99 / 10.65
& 22.89 / 5.91
& 18.19 / 5.85
& 33.64 / 14.40 \\
\midrule

\multirow{2}{*}{\makecell[l]{\routecont\hspace{0.35em}FSD~\cite{liu2025fourierseries}}}
& R2P
& 67.64 / 30.65
& 85.88 / 40.55
& 47.75 / 20.81
& 67.25 / 34.64
& 95.51 / 52.26
& 72.01 / 29.37
& 65.36 / 22.80
& 60.13 / 27.70
& 47.25 / 17.07 \\
& S2P-Fourier
& 69.45 / 40.16
& 82.54 / 46.31
& 47.74 / 22.80
& 62.09 / 38.23
& 99.13 / 67.43
& 87.71 / 52.90
& 52.58 / 22.59
& 59.00 / 39.60
& 64.80 / 31.42 \\
\midrule

\rowcolor{oursbg}
& R2P
& \bestnum{77.91} / \bestnum{36.94}
& \bestnum{95.74} / \bestnum{47.31}
& \bestnum{65.36} / 29.08
& \bestnum{82.57} / \bestnum{43.90}
& \bestnum{96.78} / \bestnum{55.51}
& \bestnum{82.67} / 33.85
& 75.47 / 27.53
& \bestnum{73.31} / 37.45
& 51.41 / 20.93 \\
\rowcolor{oursbg}
\multirow{-2}{*}{%
  \cellcolor{oursbg}%
  \makecell[l]{%
    \routecont\hspace{0.35em}\textbf{FS-FSD}%
  }%
}
& S2P-Fourier
& \shapebestnum{86.93} / \shapebestnum{55.08}
& \shapebestnum{96.31} / \shapebestnum{61.66}
& \shapebestnum{82.82} / \shapebestnum{42.21}
& 76.13 / \shapebestnum{49.77}
& \shapebestnum{99.50} / \shapebestnum{72.25}
& \shapebestnum{96.99} / \shapebestnum{68.34}
& \shapebestnum{73.28} / 39.96
& \shapebestnum{89.04} / \shapebestnum{58.08}
& \shapebestnum{81.40} / \shapebestnum{48.37} \\
\bottomrule
\end{tabularx}

\vspace{1.8pt}

\begin{tablenotes}[flushleft]
\fontsize{5.28}{6.05}\selectfont
\item Native rows are reported for reference only. The mAP column is the mean over valid categories. Native rows are excluded from best-value highlighting, while R2P and S2P best values are highlighted separately.
\end{tablenotes}
\end{threeparttable}
\end{table*}

In the S2P route, the Fourier shape branch of FS-FSD obtains \textbf{86.93/55.08}, whereas the strongest mask-based segmentation baseline, YOLO26m-seg, achieves \textbf{49.68/24.16}. Mask2Former-R50 obtains \textbf{26.98/10.77} under the same S2P-Mask route. Relative to YOLO26m-seg, FS-FSD improves $\mathrm{mAP}_{50}$ by \textbf{37.25} points and $\mathrm{mAP}_{50:95}$ by \textbf{30.92} points. Relative to the baseline FSD under S2P-Fourier (\textbf{69.45/40.16}), FS-FSD further improves the shape-output result by \textbf{17.48} points in $\mathrm{mAP}_{50}$ and by \textbf{14.92} points in $\mathrm{mAP}_{50:95}$. The large margin under $\mathrm{mAP}_{50:95}$ is particularly important, because this metric evaluates performance across stricter overlap thresholds and is more sensitive to geometric consistency. These results suggest that Fourier-based contour-native prediction is more suitable for bridge-defect boundary recovery than polygon generation from raster masks.

\begin{table*}[pos=h]
\centering
\caption{Shape-output matched true-positive (matched-TP) micro-averaged and class-wise Boundary F-score (B-F1) under unified polygon-space evaluation.}
\label{tab:perclass_bf1}

\fontsize{5.10}{5.90}\selectfont
\setlength{\tabcolsep}{0.82pt}
\renewcommand{\arraystretch}{1.12}
\setlength{\aboverulesep}{0.34ex}
\setlength{\belowrulesep}{0.34ex}

\begin{threeparttable}
\begin{tabularx}{\textwidth}{@{}>{\RaggedRight\arraybackslash}m{0.148\textwidth}
                             >{\centering\arraybackslash}m{0.066\textwidth}
                             *{9}{>{\centering\arraybackslash}X}@{}}
\toprule
Method &
\makecell[c]{Eval.\\[-1pt]Route} &
\makecell[c]{Matched-TP\\[-1pt]Micro Avg.} &
Spall. &
Crack &
Hole &
\makecell[c]{Honey-\\comb.} &
Seepage &
\makecell[c]{Rebar\\Corr.} &
Rust &
Efflo. \\
\midrule

\makecell[l]{\routemask\hspace{0.35em}YOLOv5m-seg~\cite{jocher2022ultralytics}}
& S2P-Mask
& 30.31
& 26.64
& 19.30
& 43.48
& 14.32
& 15.63
& 36.99
& 22.84
& 24.92 \\

\makecell[l]{\routemask\hspace{0.35em}YOLO11m-seg~\cite{khanam2024yolov11}}
& S2P-Mask
& 37.85
& 37.24
& \geombestnum{29.00}
& 46.27
& \geombestnum{22.15}
& 21.71
& 44.64
& 31.39
& 29.50 \\

\makecell[l]{\routemask\hspace{0.35em}YOLO26m-seg~\cite{sapkota2025yolo26}}
& S2P-Mask
& 31.20
& 28.29
& 21.42
& 45.03
& 14.08
& 16.43
& 41.61
& 25.08
& 23.37 \\

\makecell[l]{\routemask\hspace{0.35em}Mask2Former-R50~\cite{cheng2022masked}}
& S2P-Mask
& 25.55
& 18.28
& 11.90
& 41.51
& 8.13
& 16.25
& 28.54
& 17.74
& 19.16 \\
\midrule

\makecell[l]{\routecont\hspace{0.35em}Deep Snake~\cite{peng2020deep}}
& S2P-Contour
& 27.05
& 21.19
& 14.41
& 44.99
& 10.12
& 12.83
& 34.34
& 21.63
& 20.16 \\
\midrule

\makecell[l]{\routecont\hspace{0.35em}FSD~\cite{liu2025fourierseries}}
& S2P-Fourier
& 41.80
& 36.75
& 23.93
& 67.79
& 16.91
& 24.64
& 52.20
& 33.01
& 26.05 \\

\rowcolor{oursbg}
\makecell[l]{\routecont\hspace{0.35em}\textbf{FS-FSD}}
& S2P-Fourier
& \geombestnum{48.85}
& \geombestnum{42.92}
& 26.70
& \geombestnum{75.01}
& 21.74
& \geombestnum{29.84}
& \geombestnum{67.08}
& \geombestnum{41.39}
& \geombestnum{34.02} \\
\bottomrule
\end{tabularx}

\vspace{1.4pt}

\begin{tablenotes}[flushleft]
\fontsize{5.02}{5.72}\selectfont
\item S2P matched-TP B-F1 is computed on polygon-space true-positive pairs. Values are in $\times10^2$; higher is better. Best values are highlighted among available S2P shape-output entries.
\end{tablenotes}

\end{threeparttable}
\end{table*}

The comparison also shows that native-space metrics and unified polygon-space metrics do not necessarily produce the same ranking. RT-DETR-R50 reports the highest native box metric among the box-oriented baselines (\textbf{74.23/50.87}), but its R2P polygon-space result decreases to \textbf{63.46/35.78}. A similar trend is observed for Faster R-CNN, whose native box metric decreases from \textbf{65.12/40.15} to \textbf{59.87/33.52} under R2P evaluation. This discrepancy indicates that native box mAP can overestimate boundary quality in bridge-defect analysis, because a rectangular detection region does not represent the irregular defect contour required by geometry-sensitive evaluation.

The segmentation baselines further support the need for polygon-space evaluation. Their S2P-Mask results are computed from native mask outputs rather than from bounding boxes, following Appendix~\ref{sec:mask_to_polygon_yolo_mask2former}. Therefore, the performance gap between the mask-based baselines and FS-FSD should not be attributed to a weakened box-only evaluation. Instead, it reflects the difference between raster-derived boundary approximation and contour-native geometric prediction. Although mask outputs contain boundary information, their polygonal recovery is still affected by raster discretization, mask resolution, and post-processing, which can reduce strict polygon-space accuracy.

\begin{table*}[pos=h]
\centering
\caption{Shape-output matched true-positive (matched-TP) micro-averaged and class-wise Chamfer Distance (CD) under unified polygon-space evaluation.}
\label{tab:perclass_cd}

\fontsize{5.10}{5.90}\selectfont
\setlength{\tabcolsep}{0.82pt}
\renewcommand{\arraystretch}{1.12}
\setlength{\aboverulesep}{0.34ex}
\setlength{\belowrulesep}{0.34ex}

\begin{threeparttable}
\begin{tabularx}{\textwidth}{@{}>{\RaggedRight\arraybackslash}m{0.148\textwidth}
                             >{\centering\arraybackslash}m{0.066\textwidth}
                             *{9}{>{\centering\arraybackslash}X}@{}}
\toprule
Method &
\makecell[c]{Eval.\\[-1pt]Route} &
\makecell[c]{Matched-TP\\[-1pt]Micro Avg.} &
Spall. &
Crack &
Hole &
\makecell[c]{Honey-\\comb.} &
Seepage &
\makecell[c]{Rebar\\Corr.} &
Rust &
Efflo. \\
\midrule

\makecell[l]{\routemask\hspace{0.35em}YOLOv5m-seg~\cite{jocher2022ultralytics}}
& S2P-Mask
& 9.35
& 9.02
& 16.99
& 3.56
& 20.90
& 14.99
& 4.86
& 14.36
& 13.27 \\

\makecell[l]{\routemask\hspace{0.35em}YOLO11m-seg~\cite{khanam2024yolov11}}
& S2P-Mask
& 6.11
& 5.35
& 10.72
& 3.21
& 14.14
& 11.43
& 3.96
& 7.78
& 9.61 \\

\makecell[l]{\routemask\hspace{0.35em}YOLO26m-seg~\cite{sapkota2025yolo26}}
& S2P-Mask
& 9.58
& 9.38
& 16.40
& 3.48
& 23.03
& 15.28
& 4.36
& 12.83
& 13.13 \\

\makecell[l]{\routemask\hspace{0.35em}Mask2Former-R50~\cite{cheng2022masked}}
& S2P-Mask
& 15.61
& 18.24
& 27.66
& 7.68
& 34.99
& 16.96
& 8.10
& 19.15
& 19.04 \\
\midrule

\makecell[l]{\routecont\hspace{0.35em}Deep Snake~\cite{peng2020deep}}
& S2P-Contour
& 12.42
& 12.45
& 23.83
& 3.35
& 30.27
& 17.54
& 4.89
& 20.76
& 15.90 \\
\midrule

\makecell[l]{\routecont\hspace{0.35em}FSD~\cite{liu2025fourierseries}}
& S2P-Fourier
& 5.52
& 5.86
& 7.36
& 2.06
& 10.39
& 7.61
& 2.86
& 7.06
& 8.72 \\

\rowcolor{oursbg}
\makecell[l]{\routecont\hspace{0.35em}\textbf{FS-FSD}}
& S2P-Fourier
& \geombestnum{4.23}
& \geombestnum{4.41}
& \geombestnum{6.85}
& \geombestnum{1.76}
& \geombestnum{7.97}
& \geombestnum{5.40}
& \geombestnum{2.15}
& \geombestnum{4.69}
& \geombestnum{6.15} \\
\bottomrule
\end{tabularx}

\vspace{1.4pt}

\begin{tablenotes}[flushleft]
\fontsize{5.02}{5.72}\selectfont
\item S2P matched-TP CD is computed on polygon-space true-positive pairs. Values are in $\times10^3$; lower is better. Best values are highlighted among available S2P shape-output entries.
\end{tablenotes}
\end{threeparttable}
\end{table*}

The matched-TP boundary and geometry metrics provide a more detailed view of the recovered shapes after category-consistent matching has already been established. Under the S2P shape-output comparison, FS-FSD achieves the highest micro-averaged B-F1, with \textbf{48.85}, and also obtains the lowest overall Chamfer Distance (CD), perimeter error (P-Err), and area error (A-Err), with \textbf{4.23}, \textbf{7.21}, and \textbf{12.32}, respectively. The strongest mask-based baseline in micro-averaged B-F1 is YOLO11m-seg, with \textbf{37.85}, while Mask2Former-R50 obtains \textbf{25.55}. Compared with the baseline FSD, FS-FSD improves B-F1 from \textbf{41.80} to \textbf{48.85} and reduces CD from \textbf{5.52} to \textbf{4.23}. These results show that the improvement of FS-FSD is not limited to detection confidence or overlap-based AP, but also appears in local boundary adherence and global polygon geometry.

Because the matched-TP boundary and geometry columns are restricted to S2P shape outputs, box-only detectors are not ranked in these metrics. This avoids treating the regular perimeter or area statistics of rectangle-derived polygons as evidence of shape recovery. The matched-TP comparison therefore focuses on the quality of actual boundary-bearing predictions, including mask-derived polygons, explicit contours, and Fourier-derived contours.

From the perspective of model complexity, FS-FSD is not the smallest model, but its computational cost is reasonable relative to its geometric gains. FS-FSD has \textbf{87.32M} parameters and requires \textbf{396.30} billion floating-point operations (GFLOPs) at the 896-pixel input resolution. These values are nearly identical to those of the baseline FSD (\textbf{87.29M} parameters and \textbf{395.60} GFLOPs), close to Faster R-CNN (\textbf{91.76M} parameters and \textbf{408.60} GFLOPs), and much lower than RT-DETR-R50 (\textbf{1156.80} GFLOPs). Mask2Former-R50 has fewer parameters and lower computation (\textbf{44.00M} parameters and \textbf{217.40} GFLOPs), but its S2P polygon-space result remains substantially lower than that of FS-FSD. Thus, the advantage of FS-FSD should not be interpreted as a simple consequence of larger model scale. The most controlled comparison is between FSD and FS-FSD, where model size and computation are almost unchanged, but FS-FSD produces much stronger polygon-space and geometry results.

\begin{table*}[pos=h]
\centering
\caption{Shape-output matched true-positive (matched-TP) micro-averaged and class-wise relative absolute perimeter error (P-Err) under unified polygon-space evaluation.}
\label{tab:perclass_perr}

\fontsize{5.10}{5.90}\selectfont
\setlength{\tabcolsep}{0.82pt}
\renewcommand{\arraystretch}{1.12}
\setlength{\aboverulesep}{0.34ex}
\setlength{\belowrulesep}{0.34ex}

\begin{threeparttable}
\begin{tabularx}{\textwidth}{@{}>{\RaggedRight\arraybackslash}m{0.148\textwidth}
                             >{\centering\arraybackslash}m{0.066\textwidth}
                             *{9}{>{\centering\arraybackslash}X}@{}}
\toprule
Method &
\makecell[c]{Eval.\\[-1pt]Route} &
\makecell[c]{Matched-TP\\[-1pt]Micro Avg.} &
Spall. &
Crack &
Hole &
\makecell[c]{Honey-\\comb.} &
Seepage &
\makecell[c]{Rebar\\Corr.} &
Rust &
Efflo. \\
\midrule

\makecell[l]{\routemask\hspace{0.35em}YOLOv5m-seg~\cite{jocher2022ultralytics}}
& S2P-Mask
& 20.51
& 21.12
& 30.11
& 15.21
& 22.19
& 19.23
& 17.58
& 19.26
& 25.54 \\

\makecell[l]{\routemask\hspace{0.35em}YOLO11m-seg~\cite{khanam2024yolov11}}
& S2P-Mask
& 16.31
& 13.35
& 25.57
& 14.80
& 25.04
& 21.82
& 13.77
& 17.35
& 18.22 \\

\makecell[l]{\routemask\hspace{0.35em}YOLO26m-seg~\cite{sapkota2025yolo26}}
& S2P-Mask
& 21.13
& 19.39
& 30.58
& 15.95
& 28.43
& 28.23
& 15.30
& 29.40
& 23.52 \\

\makecell[l]{\routemask\hspace{0.35em}Mask2Former-R50~\cite{cheng2022masked}}
& S2P-Mask
& 17.76
& 21.24
& 20.02
& 16.40
& 27.33
& 14.18
& 15.90
& 17.87
& 15.10 \\
\midrule

\makecell[l]{\routecont\hspace{0.35em}Deep Snake~\cite{peng2020deep}}
& S2P-Contour
& 13.44
& 13.60
& 15.21
& 14.10
& 14.12
& 11.72
& 13.24
& 16.60
& 10.65 \\
\midrule

\makecell[l]{\routecont\hspace{0.35em}FSD~\cite{liu2025fourierseries}}
& S2P-Fourier
& 8.87
& 8.12
& 7.33
& 10.00
& 6.84
& 10.21
& 9.18
& 7.86
& 9.31 \\

\rowcolor{oursbg}
\makecell[l]{\routecont\hspace{0.35em}\textbf{FS-FSD}}
& S2P-Fourier
& \geombestnum{7.21}
& \geombestnum{7.32}
& \geombestnum{5.73}
& \geombestnum{8.71}
& \geombestnum{6.01}
& \geombestnum{5.86}
& \geombestnum{6.01}
& \geombestnum{6.83}
& \geombestnum{7.00} \\
\bottomrule
\end{tabularx}

\vspace{1.4pt}

\begin{tablenotes}[flushleft]
\fontsize{5.02}{5.72}\selectfont
\item S2P matched-TP P-Err is computed on polygon-space true-positive pairs. Values are in $\times10^2$; lower is better. Best values are highlighted among available S2P shape-output entries.
\end{tablenotes}
\end{threeparttable}
\end{table*}

Table~\ref{tab:perclass_ap} further shows that the advantage of FS-FSD is not concentrated in only a few defect categories. In the R2P route, FS-FSD obtains the highest overall AP$_{50}$/AP$_{50:95}$ (\textbf{77.91/36.94}) and achieves the best AP$_{50}$ on spalling, crack, hole, honeycombing, seepage, and rust. At stricter thresholds, several box-derived baselines remain competitive in specific categories. YOLO11m-seg obtains the highest AP$_{50:95}$ on crack (\textbf{29.78}) and rebar corrosion (\textbf{34.92}), whereas RT-DETR-R50 obtains the highest AP$_{50:95}$ on seepage (\textbf{44.78}), rust (\textbf{39.88}), and efflorescence (\textbf{32.76}). These category-level results show that R2P performance is affected by both detector localization and class-specific shape characteristics. Nevertheless, the overall improvement from FSD to FS-FSD confirms that the proposed formulation improves geometric localization quality beyond the original framework.

In the S2P route, FS-FSD achieves the best overall AP$_{50}$/AP$_{50:95}$ (\textbf{86.93/55.08}) and provides the strongest strict-threshold performance among all compared methods. Hole is the only category in which a mask-derived baseline obtains a slightly higher AP$_{50}$, with YOLO26m-seg reaching \textbf{76.21} compared with \textbf{76.13} from FS-FSD. However, FS-FSD achieves a much higher AP$_{50:95}$ on this category (\textbf{49.77} versus \textbf{40.14}). This indicates that mask-derived polygons may occasionally approach Fourier-based contour outputs under loose overlap criteria, but the contour-native representation remains more stable when stricter geometric agreement is required. For rebar corrosion, FS-FSD obtains the highest AP$_{50}$ (\textbf{73.28}), while YOLO11m-seg remains marginally higher on AP$_{50:95}$ (\textbf{40.07} versus \textbf{39.96}). Mask2Former-R50 does not change the overall pattern, with \textbf{26.98/10.77} in the S2P-Mask route. Compared with the baseline FSD, FS-FSD improves the overall S2P-Fourier result from \textbf{69.45/40.16} to \textbf{86.93/55.08}, indicating stronger performance in both coarse overlap and strict polygon consistency.

\begin{table*}[pos=h]
\centering
\caption{Shape-output matched true-positive (matched-TP) micro-averaged and class-wise relative absolute area error (A-Err) under unified polygon-space evaluation.}
\label{tab:perclass_aerr}

\fontsize{5.10}{5.90}\selectfont
\setlength{\tabcolsep}{0.82pt}
\renewcommand{\arraystretch}{1.12}
\setlength{\aboverulesep}{0.34ex}
\setlength{\belowrulesep}{0.34ex}

\begin{threeparttable}
\begin{tabularx}{\textwidth}{@{}>{\RaggedRight\arraybackslash}m{0.148\textwidth}
                             >{\centering\arraybackslash}m{0.066\textwidth}
                             *{9}{>{\centering\arraybackslash}X}@{}}
\toprule
Method &
\makecell[c]{Eval.\\[-1pt]Route} &
\makecell[c]{Matched-TP\\[-1pt]Micro Avg.} &
Spall. &
Crack &
Hole &
\makecell[c]{Honey-\\comb.} &
Seepage &
\makecell[c]{Rebar\\Corr.} &
Rust &
Efflo. \\
\midrule

\makecell[l]{\routemask\hspace{0.35em}YOLOv5m-seg~\cite{jocher2022ultralytics}}
& S2P-Mask
& 22.71
& 21.98
& 25.05
& 23.26
& 19.88
& 22.79
& 23.33
& 22.03
& 21.94 \\

\makecell[l]{\routemask\hspace{0.35em}YOLO11m-seg~\cite{khanam2024yolov11}}
& S2P-Mask
& 18.87
& 16.06
& 21.42
& 21.34
& 18.72
& 18.83
& 19.76
& 16.98
& 17.28 \\

\makecell[l]{\routemask\hspace{0.35em}YOLO26m-seg~\cite{sapkota2025yolo26}}
& S2P-Mask
& 22.97
& 20.94
& 22.78
& 23.78
& 24.96
& 27.18
& 22.83
& 21.81
& 23.90 \\

\makecell[l]{\routemask\hspace{0.35em}Mask2Former-R50~\cite{cheng2022masked}}
& S2P-Mask
& 27.02
& 28.72
& 25.77
& 27.06
& 31.52
& 21.59
& 29.37
& 31.90
& 23.43 \\
\midrule

\makecell[l]{\routecont\hspace{0.35em}Deep Snake~\cite{peng2020deep}}
& S2P-Contour
& 25.77
& 27.60
& 26.97
& 26.73
& 19.50
& 21.94
& 29.05
& 25.67
& 21.96 \\
\midrule

\makecell[l]{\routecont\hspace{0.35em}FSD~\cite{liu2025fourierseries}}
& S2P-Fourier
& 16.23
& 14.80
& 19.45
& 15.67
& 12.29
& 12.72
& 21.34
& 17.12
& 16.54 \\

\rowcolor{oursbg}
\makecell[l]{\routecont\hspace{0.35em}\textbf{FS-FSD}}
& S2P-Fourier
& \geombestnum{12.32}
& \geombestnum{10.86}
& \geombestnum{14.97}
& \geombestnum{13.34}
& \geombestnum{7.69}
& \geombestnum{11.12}
& \geombestnum{12.82}
& \geombestnum{10.71}
& \geombestnum{13.03} \\
\bottomrule
\end{tabularx}

\vspace{1.4pt}

\begin{tablenotes}[flushleft]
\fontsize{5.02}{5.72}\selectfont
\item S2P matched-TP A-Err is computed on polygon-space true-positive pairs. Values are in $\times10^2$; lower is better. Best values are highlighted among available S2P shape-output entries.
\end{tablenotes}
\end{threeparttable}
\end{table*}

Tables~\ref{tab:perclass_bf1}--\ref{tab:perclass_aerr} show that the contour-native advantage of FS-FSD remains visible when the comparison is restricted to matched true-positive pairs. This conditioning reduces the influence of whether a defect has been detected and focuses more directly on the quality of the recovered geometry after a correct match has been established. In this sense, the matched-TP metrics complement AP by isolating boundary and shape quality from confidence ranking and recall.

At the class level, FS-FSD achieves the lowest S2P CD for all eight defect categories. It also obtains the lowest S2P P-Err and A-Err across all categories. B-F1 is more category-dependent: YOLO11m-seg obtains the best B-F1 on crack and honeycombing, while FS-FSD achieves the best micro-averaged B-F1 and the best class-wise B-F1 on spalling, hole, seepage, rebar corrosion, rust, and efflorescence. These results indicate that the advantage of FS-FSD is not limited to overlap-based AP. It also extends to distance-based boundary adherence and more stable global polygon-geometry recovery.

Overall, the comparison leads to a consistent experimental finding. Native box or mask metrics provide useful reference values, but they do not fully reflect the geometric quality required for bridge-defect boundary recovery. Once all methods are evaluated in a unified polygon space, FS-FSD achieves the strongest overall polygon-space AP, the best strict S2P polygon consistency, the highest S2P matched-TP boundary F-score, the lowest boundary distance, and the most accurate perimeter and area recovery. The improvement over FSD is particularly important because both methods have nearly identical parameter counts and computational costs. This indicates that the performance gain mainly comes from the contour-native and frequency-aware formulation rather than from increased model capacity.

\subsection{Ablation analysis of training formulation and Fourier order}
\label{sec:ablation_studies}

Following the baseline comparison in Section~\ref{subsec:comparison_baselines}, this subsection analyzes which design choices contribute to the performance gains of FS-FSD. The ablation study is conducted from two complementary perspectives: the \emph{training formulation} and the \emph{Fourier order}. The tabulated ablation results in this subsection are reported as mean Average Precision (mAP) under the Fourier Shape-to-Polygon route, namely the S2P-Fourier polygon-space evaluation protocol used for FS-FSD in Tables~\ref{tab:overall_main} and~\ref{tab:perclass_ap}. Therefore, the conclusions are drawn in the same unified polygon space as the main comparison, rather than from native-space metrics. In particular, the full $n_f=16$ configuration in this subsection reports the same FS-FSD S2P-Fourier result as the main comparison, i.e., \textbf{86.93/55.08}.

Three key components of the training formulation are examined: \textbf{Unit-grid Direct Frequency-domain Supervision (U-DFS)}, \textbf{Order-aware Coefficient Normalization}, abbreviated as \textbf{FT norm}, and \textbf{Closed-form Harmonic Phase Alignment (CHPA)}. U-DFS supervises the predicted Fourier coefficients directly in the unit-grid coefficient space, instead of applying the contour loss after inverse Fourier reconstruction in the spatial domain. FT norm balances coefficient errors across different harmonic orders. CHPA resolves the phase ambiguity caused by the arbitrary starting point of a closed contour.

\begin{table}[pos=h]
\centering
\caption{Ablation of FS-FSD under S2P-Fourier polygon-space evaluation.}
\label{tab:ablation_ftfsd}

\fontsize{5.85}{6.60}\selectfont
\setlength{\tabcolsep}{1.00pt}
\renewcommand{\arraystretch}{1.07}
\setlength{\aboverulesep}{0.26ex}
\setlength{\belowrulesep}{0.26ex}
\setlength{\cmidrulesep}{0.10ex}

\begin{threeparttable}
\begin{tabular}{@{}>{\RaggedRight\arraybackslash}p{0.268\linewidth}
                >{\centering\arraybackslash}m{0.042\linewidth}
                >{\centering\arraybackslash}m{0.050\linewidth}
                >{\centering\arraybackslash}m{0.042\linewidth}
                >{\centering\arraybackslash}m{0.032\linewidth}
                >{\centering\arraybackslash}m{0.090\linewidth}
                >{\centering\arraybackslash}m{0.102\linewidth}
                >{\centering\arraybackslash}m{0.094\linewidth}
                >{\centering\arraybackslash}m{0.106\linewidth}@{}}
\toprule
\multirow{2}{*}{Variant} &
\multicolumn{3}{c}{Components} &
\multirow{2}{*}{$n_f$} &
\multicolumn{2}{c}{S2P-Fourier Polygon mAP} &
\multicolumn{2}{c}{Recovered gain (p.p.)} \\
\cmidrule(lr){2-4}\cmidrule(lr){6-7}\cmidrule(lr){8-9}
& \makecell[c]{U-DFS}
& \makecell[c]{FT\\norm}
& CHPA
&
& \makecell[c]{$mAP_{50}$ $\uparrow$\\[-0.5pt](\%)}
& \makecell[c]{$mAP_{50{:}95}$ $\uparrow$\\[-0.5pt](\%)}
& \makecell[c]{$mAP_{50}$}
& \makecell[c]{$mAP_{50{:}95}$} \\
\midrule

\multicolumn{9}{@{}l}{\textit{A. Training formulation with fixed $n_f=16$}} \\[-1pt]
FS-FSD without U-DFS
& \abno & \abno & \abyes & 16
& 66.54 & 39.80
& +18.96 & +11.99 \\
FS-FSD without FT norm
& \abyes & \abno & \abyes & 16
& 85.50 & 51.79
& +1.43 & +3.29 \\
FS-FSD without CHPA
& \abyes & \abyes & \abno & 16
& 78.35 & 46.20
& +8.58 & +8.88 \\
FS-FSD full
& \abyes & \abyes & \abyes & 16
& \abbest{86.93} & 55.08
& n.a. & n.a. \\
\midrule

\multicolumn{9}{@{}l}{\textit{B. Fourier order with full formulation}} \\[-1pt]
FS-FSD with $n_f=8$
& \abyes & \abyes & \abyes & 8
& 85.48 & \abbest{55.28}
& n.a. & n.a. \\
FS-FSD with $n_f=12$
& \abyes & \abyes & \abyes & 12
& 86.32 & \abbest{55.28}
& n.a. & n.a. \\
FS-FSD with $n_f=16$
& \abyes & \abyes & \abyes & 16
& \abbest{86.93} & 55.08
& n.a. & n.a. \\
\bottomrule
\end{tabular}

\vspace{1.2pt}

\begin{tablenotes}[flushleft]
\fontsize{5.12}{5.84}\selectfont
\item U-DFS = Unit-grid Direct Frequency-domain Supervision; FT norm = Order-aware Coefficient Normalization; CHPA = Closed-form Harmonic Phase Alignment; p.p. = percentage points. The full $n_f=16$ setting corresponds to the FS-FSD S2P-Fourier result reported in Tables~\ref{tab:overall_main} and~\ref{tab:perclass_ap}. Since FT norm acts on frequency-domain coefficient errors, it is disabled when U-DFS is removed. Recovered gain denotes the improvement obtained by restoring the removed component under the closest matched setting.
\end{tablenotes}
\end{threeparttable}
\end{table}

Table~\ref{tab:ablation_ftfsd} first examines the contribution of the training formulation. In Block~A, the Fourier order is fixed to $n_f=16$, and one component is removed at a time. The recovered-gain columns report the improvement obtained after the removed component is restored under the closest matched setting. Since FT norm operates on coefficient-domain errors, it is not activated when U-DFS is removed. Therefore, the contribution of U-DFS is evaluated by comparing the variant without U-DFS against the closest coefficient-domain counterpart, namely the variant without FT norm. Both settings keep CHPA enabled and keep FT norm disabled.

The results show that \textbf{U-DFS is the most influential component} in the training formulation. When U-DFS is removed and the contour loss is moved back to the spatial domain after inverse Fourier reconstruction, the result decreases to \textbf{66.54/39.80}. Under the matched setting where FT norm remains disabled and CHPA is enabled, restoring U-DFS improves the result to \textbf{85.50/51.79}. This corresponds to recovered gains of \textbf{18.96} percentage points in $mAP_{50}$ and \textbf{11.99} percentage points in $mAP_{50{:}95}$. This large difference indicates that direct coefficient-domain supervision is essential for effective Fourier contour learning in FS-FSD.

This result is also consistent with the observed training behavior. When contour supervision is applied only after inverse Fourier reconstruction, the optimization becomes less robust to hyperparameter choices such as learning rate, loss weighting, and initialization. In contrast, U-DFS applies supervision directly in the unit-grid Fourier coefficient space, which is aligned with the output parameterization predicted by the network. This reduces the instability introduced by repeatedly relying on inverse Fourier reconstruction inside the learning loop. Therefore, U-DFS improves not only contour accuracy but also training stability.

The contribution of \textbf{FT norm} is smaller than that of U-DFS, but it is systematic. Starting from the U-DFS formulation without order-aware normalization, introducing FT norm improves the result from \textbf{85.50/51.79} to the full FS-FSD result of \textbf{86.93/55.08}. The corresponding gains are \textbf{1.43} percentage points in $mAP_{50}$ and \textbf{3.29} percentage points in $mAP_{50{:}95}$. The larger improvement on the stricter metric suggests that coefficient balancing across harmonic orders is particularly useful for high-overlap contour quality. Without such normalization, low-order harmonics tend to dominate the loss because they carry most of the global shape energy, whereas higher-order harmonics, which describe local curvature and small boundary irregularities, may receive insufficient effective supervision. FT norm alleviates this imbalance and improves the learning of higher-order components.

The effect of \textbf{CHPA} is also substantial. Removing Closed-form Harmonic Phase Alignment reduces the result to \textbf{78.35/46.20}. Restoring CHPA recovers the full FS-FSD result of \textbf{86.93/55.08}, corresponding to gains of \textbf{8.58} percentage points in $mAP_{50}$ and \textbf{8.88} percentage points in $mAP_{50{:}95}$. This confirms that phase ambiguity is a practical issue in Fourier contour regression. For a closed contour, geometrically identical or nearly identical curves can correspond to different coefficient sequences if the sampling starting point is shifted. CHPA aligns the harmonic phase between prediction and target before computing the supervision signal, thereby reducing spurious coefficient errors caused by starting-point mismatch. The comparable gains on both loose and strict metrics indicate that phase-consistent supervision improves both global contour stability and local boundary fitting.

Block~B of Table~\ref{tab:ablation_ftfsd} evaluates the effect of the Fourier order under the full formulation. Increasing the order from $n_f=8$ to $n_f=16$ raises $mAP_{50}$ from \textbf{85.48} to \textbf{86.93}, showing that additional harmonics improve coarse and medium-scale contour coverage. In contrast, $mAP_{50{:}95}$ is already close to saturation, varying only from \textbf{55.08} to \textbf{55.28}. The slight peak at $n_f=8$ and $n_f=12$ is only \textbf{0.20} percentage points above the $n_f=16$ setting, which is a marginal difference at the reported precision. This suggests that moderate Fourier orders are already sufficient to capture most high-overlap geometric structures, whereas higher orders mainly provide additional capacity for category-dependent local curvature and small boundary details. In the remainder of this paper, $n_f=16$ is used as the default configuration because it provides the highest overall $mAP_{50}$, matches the configuration used in the main comparison, and maintains essentially the same $mAP_{50{:}95}$ level.

\begin{figure}[pos=h]
    \centering
    \includegraphics[width=\linewidth]{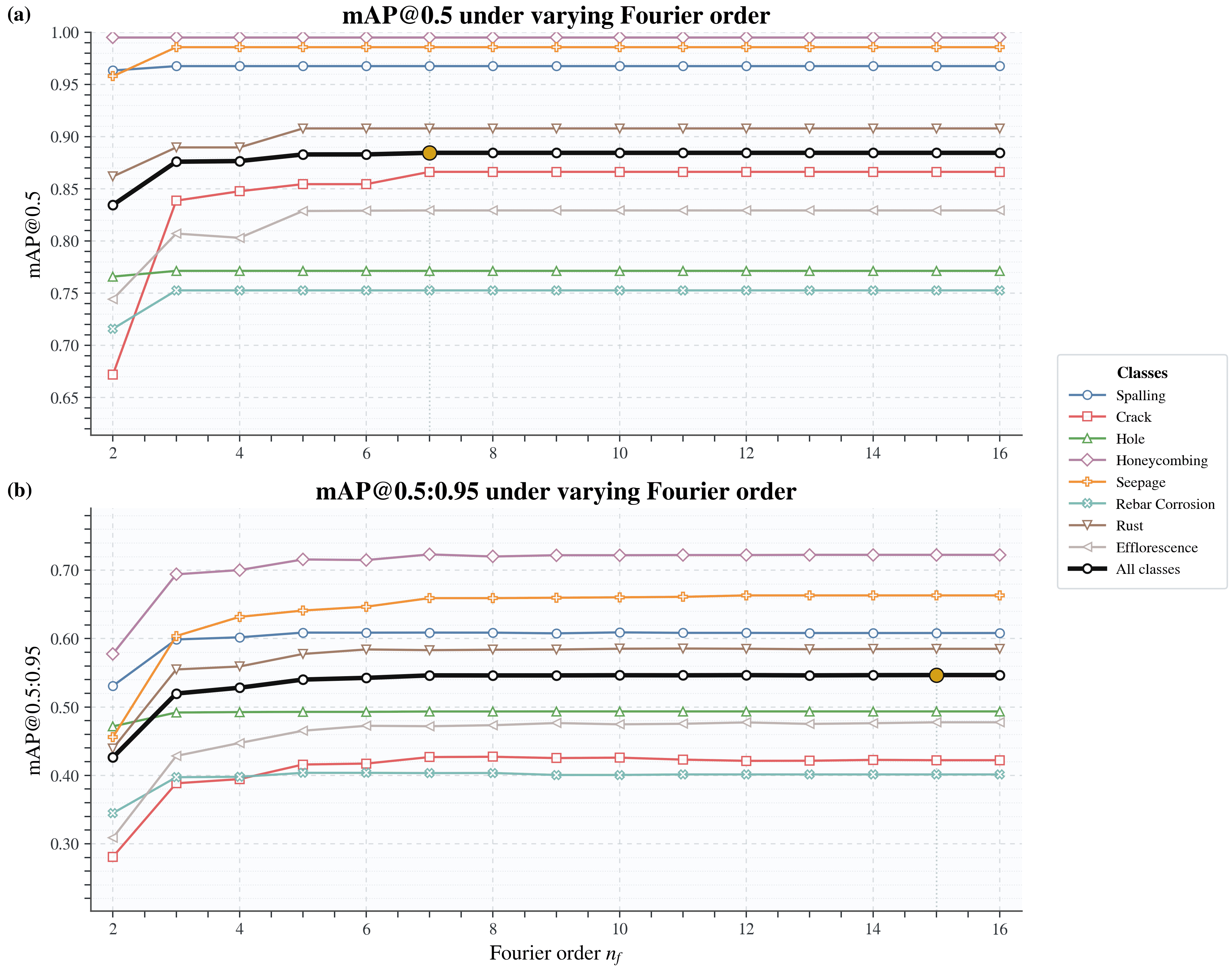}
    \caption{mAP across Fourier orders under coefficient truncation. A trained $n_f=16$ FS-FSD model is evaluated by retaining only the first $k$ Fourier orders, while the $n_f=16$ ground-truth contour is kept as the fixed reference.}
    \label{fig:ft_nf_map}
\end{figure}

To isolate the representational effect of Fourier order more explicitly, the full trained $n_f=16$ model corresponding to the main FS-FSD configuration is further evaluated by progressively truncating its predicted Fourier coefficients to orders 2 through 16, while keeping the \textbf{$n_f=16$ ground-truth contour} as the fixed reference. Unlike Block~B, this analysis does not retrain separate models for different orders. Instead, it reinterprets the prediction of the same high-order model as lower-order contour approximations at validation time, which more directly reveals the representational role of harmonic order itself. The resulting mAP curves across Fourier orders are shown in Fig.~\ref{fig:ft_nf_map}. As the retained order increases, both $mAP_{50}$ and $mAP_{50{:}95}$ rise rapidly in the low-order regime and then gradually saturate. This behavior indicates that the global support region and dominant shape of a defect contour are captured by the first few harmonics, whereas additional higher-order terms mainly contribute to local geometric refinement.

\begin{figure*}[pos=h]
    \centering
    \includegraphics[width=\textwidth]{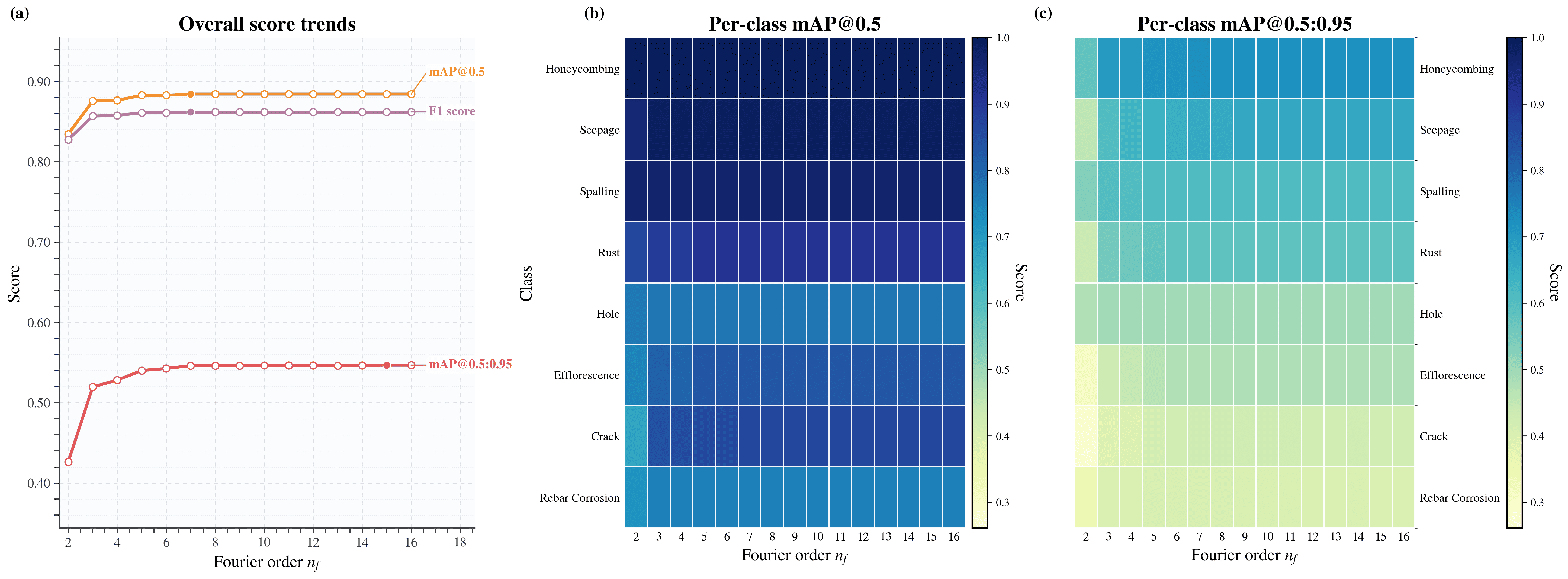}
    \caption{Metric and category-level AP summaries across Fourier orders under the fixed $n_f=16$ contour reference. The left panel reports overall $mAP_{50}$, $mAP_{50{:}95}$, and F1-score, while the middle and right panels show category-level $mAP_{50}$ and $mAP_{50{:}95}$ heatmaps across Fourier orders.}
    \label{fig:ft_nf_summary}
\end{figure*}

A more complete summary is provided in Fig.~\ref{fig:ft_nf_summary}. The left panel reports the overall $mAP_{50}$, $mAP_{50{:}95}$, and F1-score curves under the same fixed $n_f=16$ reference evaluation. The trends are consistent with Fig.~\ref{fig:ft_nf_map}, showing a fast increase at low orders followed by clear saturation. The middle panel presents the category-level $mAP_{50}$ heatmap over different orders, and the right panel presents the corresponding category-level $mAP_{50{:}95}$ heatmap. In both heatmaps, darker colors indicate higher accuracy. These plots further show that the sensitivity to order truncation is category dependent. Honeycombing, seepage, and spalling remain strong across most orders, indicating that these categories can be represented well with relatively low or medium orders. By contrast, crack and rebar corrosion saturate more slowly, especially under $mAP_{50{:}95}$, which implies that thin, elongated, or locally irregular boundaries require more harmonics to preserve fine geometry.

\begin{figure*}[pos=h]
    \centering
    \includegraphics[width=\textwidth]{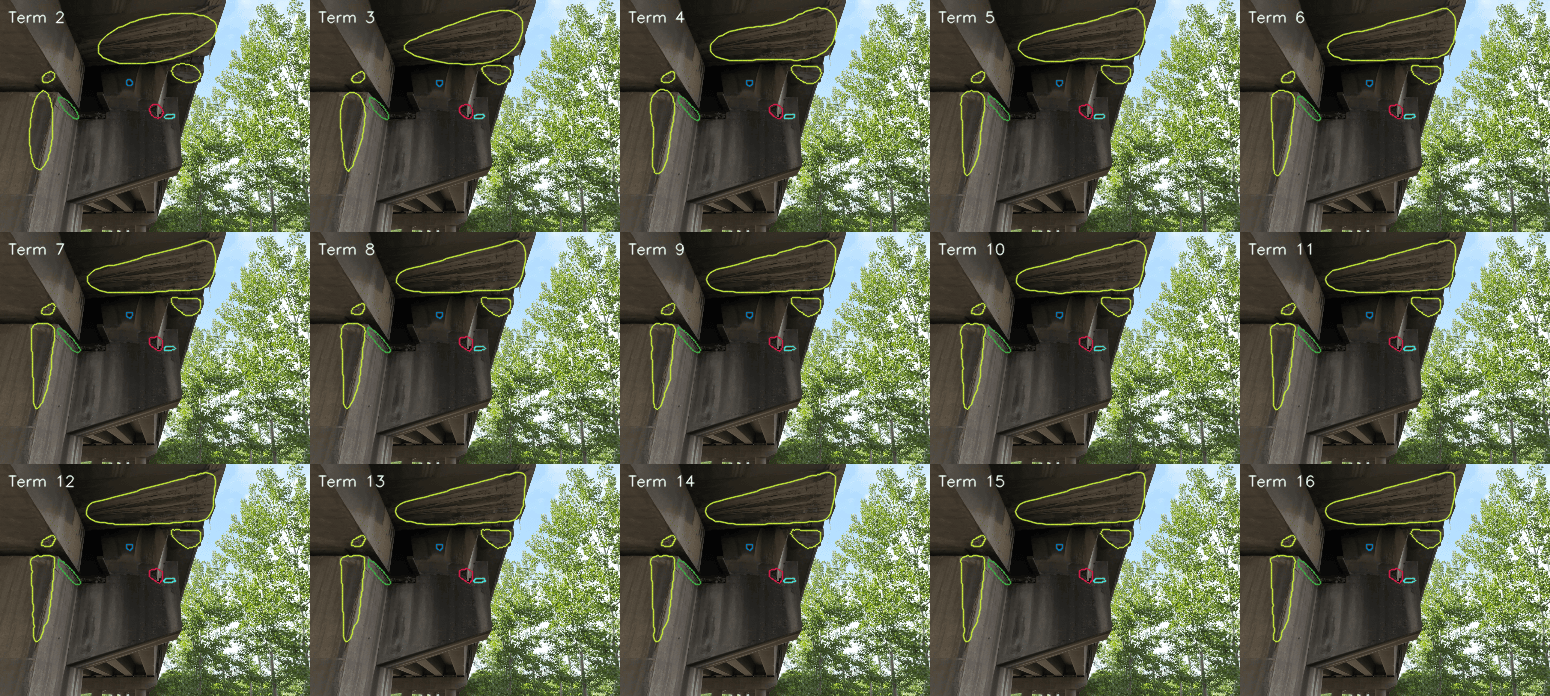}
    \caption{Qualitative contour reconstruction under different Fourier orders. Increasing the order progressively restores elongated structures, local protrusions, curvature changes, and small boundary irregularities.}
    \label{fig:ft_order_vis}
\end{figure*}

The geometric meaning of the Fourier order is illustrated qualitatively in Fig.~\ref{fig:ft_order_vis}, where the reconstructed contour of a representative bridge image is visualized from order 2 to order 16. At very low orders, the recovered shape captures only the coarse support region and dominant orientation of each defect instance. As the order increases, major elongated structures, local protrusions, curvature changes, and small boundary irregularities are progressively restored. After the medium-order range, the global geometry is largely stabilized, and the remaining higher-order terms mainly contribute to local curvature correction and fine boundary adjustment. This visual behavior agrees with the quantitative saturation observed in Figs.~\ref{fig:ft_nf_map} and~\ref{fig:ft_nf_summary}.

Overall, the ablation study shows that the final FS-FSD design is supported by both optimization and representation evidence. U-DFS provides the largest contribution and is essential for stable and accurate Fourier contour learning. FT norm and CHPA provide complementary gains: FT norm balances the learning of different harmonic orders, whereas CHPA removes phase inconsistency caused by starting-point mismatch in closed contours. The Fourier-order analysis further shows that moderate orders already capture the dominant contour structure, while higher orders mainly refine category-dependent boundary details. These findings support the use of the full formulation with $n_f=16$ as the default configuration throughout the paper.

\section{Proof-of-Concept Workflow Validation}
\label{sec:poc_validation}

The previous sections showed that FS-FSD improves image-space geometric prediction under the unified polygon-based evaluation protocol. Beyond benchmark accuracy, an important practical question is whether the contour-native output can also reduce the cost of storing, recovering, visualizing, and reusing prediction records after inference. This section therefore presents a proof-of-concept workflow validation that focuses on post-prediction data handling and component-level deployability, rather than treating the evaluation as a full inspection-management-system deployment.

Two complementary aspects are considered. The first aspect concerns post-prediction archival and reuse, including SQLite-based archival, coefficient recovery, and browser-side visualization of archived records. It examines whether the predicted contour representation can be stored, reopened, reconstructed, and reused as an image-space geometric record. The second aspect concerns edge-side executability and evaluates whether the same contour-native formulation remains practical on resource-constrained embedded hardware. Together, these analyses extend the evaluation from prediction accuracy to the practical chain of prediction generation, compact storage, recovery, visualization, and reuse.

All geometric quantities in this section are defined in image space. The recovered polygons, areas, perimeters, centroids, orientations, and elongation descriptors are computed in pixel coordinates or derived from pixel-coordinate polygons. Accordingly, they should be interpreted as image-space geometric records rather than physical-space measurements, metric-scale defect dimensions, or component-level bridge coordinates. Establishing physical-space geometry would require additional camera calibration, scale recovery, surface registration, or other mapping procedures, which are outside the scope of this proof-of-concept validation.


\subsection{Archival and Recovery of Image-Space Damage Records}
\label{subsec:archive_recovery}

The first proof-of-concept perspective addresses the post-prediction stage, namely whether the contour-native output of FS-FSD can be efficiently archived, reliably recovered, and subsequently reused as an image-space polygonal record. To keep this perspective structurally coherent, the validation is organized into database-oriented archival and recovery benchmarking, followed by browser-side visualization in the subsequent workflow validation. The present subsection focuses on the database-oriented part, where archive compactness and recovery efficiency are assessed under a consistent post-prediction reuse setting.

\subsubsection{SQLite-based archival and image-space recovery}
\label{subsubsec:database_archive_recovery}

The preceding experiments established that FS-FSD provides stronger polygon-space geometric fidelity than box- and mask-oriented baselines. For inspection data workflows, however, prediction quality is not the only relevant factor. Once defects are detected, their records may need to be stored, transferred, reopened, and converted into polygonal representations for later review, visualization, or descriptor computation. This benchmark therefore isolates a practical representation-layer question: whether a contour-native output can reduce the post-prediction archival and recovery burden of image-space defect records, rather than evaluating the implementation of a complete inspection management system.

A database-oriented proof-of-concept benchmark is constructed using SQLite as a lightweight and reproducible surrogate for local inspection-side storage. The benchmark is conducted on the same \textbf{377-image test split} as the main experiments, and all images are processed at the same \textbf{896$\times$896} inference resolution used in the preceding sections. FS-FSD is benchmarked in its native archive route, namely the image-coordinate Fourier coefficient vector produced directly by the detection head and stored without raster-mask materialization, mask thresholding, contour extraction, or contour-to-Fourier fitting.

YOLO26m-seg is selected as the raster baseline because it is the strongest segmentation model in the main polygon-space shape comparison. To make the representational difference explicit, YOLO26m-seg is evaluated under four archive routes derived from the same predicted masks: \textbf{RLE-full}, which stores a full-image uint32 run-length-encoded mask; \textbf{RLE-crop}, which stores a foreground-tight cropped RLE mask; \textbf{Poly-256}, which stores the largest external contour resampled to 256 points; and \textbf{Fourier-fit}, which stores an order-16 Fourier descriptor fitted from the mask contour. These routes correspond to raster-native storage, compressed raster storage, direct polygon storage, and post-hoc vector fitting, respectively.

For comparability, the benchmark is divided into an archive stage and a recovery stage. The archive stage measures the cost of transforming model predictions into persistent records and inserting them into the database. The recovery stage measures the cost of reopening those records and restoring them into usable image-space polygonal records. To ensure that all routes are compared under the same downstream target, every representation is required to produce the same recovered object at the end of recovery, namely a \textbf{256-point polygon in image coordinates}. On top of this polygon, a common set of image-space geometric descriptors is computed, including area, perimeter, centroid, orientation, and elongation. These descriptors are introduced only to impose a consistent downstream workload for the archive-and-recovery benchmark. They are not additional direct prediction outputs of FS-FSD and should be interpreted as derived image-space descriptors rather than physical-space engineering measurements. Under this design, the comparison is between different representation routes that must ultimately support the same image-space recovery target.

A key methodological issue is how to establish a comparable cost basis across archive routes. If recovery-stage latency alone is used as the headline metric, routes that have already materialized geometry during archival may appear artificially favorable, even though part of their cost has merely been shifted upstream rather than eliminated. This issue is especially relevant to \textbf{Poly-256} and \textbf{Fourier-fit}, because both routes already incur mask-side conversion cost before recovery begins. To avoid this distortion, a route-consistent cost-attribution strategy is adopted: whenever a representation-conversion cost is incurred during archival, it is attributed to the route that incurred it. Raw recovery-only latency is therefore retained only as a supplementary diagnostic quantity rather than as the principal basis for route-level conclusions.

A second comparability issue arises because FS-FSD and YOLO26m-seg do not generate the same number of archived predictions on the test split, namely \textbf{1,660} and \textbf{1,849}, respectively. Archive-stage time cannot be compared fairly on a per-image basis alone, because a method that archives more defect instances would otherwise carry a larger burden simply due to higher prediction density. To reduce this effect, archive-side serialization and database-insertion time are normalized by the mean number of stored predictions per image, so that route cost is expressed on a per-defect basis. In the present benchmark, the mean prediction density is \textbf{4.40} defects/image for FS-FSD and \textbf{4.90} defects/image for YOLO26m-seg.

Let $N_{\mathrm{img}}$ denote the number of test images and $N_{\mathrm{pred}}$ denote the number of stored predictions. The prediction density is defined as
\begin{equation}
D_{\mathrm{pred}}=\frac{N_{\mathrm{pred}}}{N_{\mathrm{img}}}.
\label{eq:fair_pred_density}
\end{equation}
Based on $D_{\mathrm{pred}}$, the archive-route overhead per defect excluding inference is defined as
\begin{equation}
T_{\mathrm{arch}}^{\mathrm{route}}
=
\frac{T_{\mathrm{ser}}^{\mathrm{img}}+T_{\mathrm{db}}^{\mathrm{img}}}{D_{\mathrm{pred}}},
\label{eq:fair_archive_route_overhead}
\end{equation}
where $T_{\mathrm{ser}}^{\mathrm{img}}$ and $T_{\mathrm{db}}^{\mathrm{img}}$ denote image-level serialization time and database insertion time, respectively. Importantly, $T_{\mathrm{ser}}^{\mathrm{img}}$ includes any route-specific representation materialization required to produce the archived payload, such as contour extraction, polygon resampling, or Fourier fitting. Let $T_{\mathrm{usable}}^{\mathrm{raw}}$ denote the raw recovery-to-usable latency per defect, i.e., database read, payload decoding, polygon reconstruction, and image-space descriptor extraction. The fair route-to-usable latency excluding inference is then written as
\begin{equation}
T_{\mathrm{usable}}^{\mathrm{route,ex}}
=
T_{\mathrm{arch}}^{\mathrm{route}}
+
T_{\mathrm{usable}}^{\mathrm{raw}},
\label{eq:fair_route_usable_ex}
\end{equation}
and the full route-to-usable latency including inference is defined as
\begin{equation}
T_{\mathrm{usable}}^{\mathrm{route,in}}
=
\frac{T_{\mathrm{arch,e2e}}^{\mathrm{img}}}{D_{\mathrm{pred}}}
+
T_{\mathrm{usable}}^{\mathrm{raw}},
\label{eq:fair_route_usable_in}
\end{equation}
where $T_{\mathrm{arch,e2e}}^{\mathrm{img}}$ denotes archive end-to-end latency per image. Finally, the fair route throughput excluding inference is
\begin{equation}
Q_{\mathrm{route}}^{\mathrm{ex}}
=
\frac{1000}{T_{\mathrm{usable}}^{\mathrm{route,ex}}},
\label{eq:fair_route_throughput}
\end{equation}
which measures how many archived defects can be converted into recovered image-space polygonal records per second under a consistent route-level cost basis.

\begin{table}[pos=h]
\centering
\caption{Headline metrics used in the route-comparable archive-and-recovery benchmark.}
\label{tab:archive_fair_metric_defs}

\fontsize{5.85}{6.65}\selectfont
\setlength{\tabcolsep}{4.0pt}
\renewcommand{\arraystretch}{1.10}
\setlength{\aboverulesep}{0.30ex}
\setlength{\belowrulesep}{0.30ex}

\begin{tabularx}{\linewidth}{@{}>{\RaggedRight\arraybackslash}p{0.22\linewidth}>{\RaggedRight\arraybackslash}X@{}}
\toprule
\textbf{Metric} & \textbf{Definition and interpretation} \\
\midrule
Payload / defect & Total geometry payload bytes divided by the number of stored defects. This measures the compactness of the archived geometric carrier itself. \\

Record / defect & Total bytes of the complete stored record, i.e., geometry payload plus the compact logical header used for indexing, query, and reuse. This better reflects the practical archive burden of a reusable image-space record than payload size alone. \\

Archive route overhead / defect (excl.\ infer) & Per-defect archive-side representation overhead, computed as image-level serialization, including route-specific representation conversion, plus database insertion, normalized by the mean number of stored predictions per image. This isolates the cost introduced by the archive route itself rather than detector forward inference. \\

Route-to-usable / defect (excl.\ infer) & Route-level cost of turning one prediction into a recovered image-space polygonal record when inference is excluded, i.e., archive route overhead plus raw recovery-to-usable latency. This is the principal route-comparable latency metric. \\

Route-to-usable / defect (incl.\ infer) & Full route-level cost of turning one prediction into a recovered image-space polygonal record when inference is also included. This reports the complete benchmarked cost from prediction generation to record recovery. \\

Route throughput / defects/s (excl.\ infer) & The number of stored defects that can be transformed into recovered image-space polygonal records per second under the route-level latency excluding inference. Higher values indicate a more efficient archive-and-reuse route under this benchmark setting. \\
\bottomrule
\end{tabularx}
\end{table}

Table~\ref{tab:archive_fair_routes} reports the route-level comparison under these metrics. The first block summarizes direct footprint indicators, whereas the second block reports route-level costs obtained by combining archive-side route overhead with downstream recovery burden. Because the four YOLO-side representations are stored in a shared physical SQLite database in the implementation, route-wise SQLite file size is not used as a headline discriminative metric. Instead, \emph{payload / defect} and \emph{record / defect} are used to characterize the practical archive footprint of each route.

\begin{table*}[pos=h]
\centering
\caption{Route-comparable archive-and-recovery comparison between FS-FSD and YOLO26m-seg under different YOLO-side storage routes.}
\label{tab:archive_fair_routes}

\fontsize{5.70}{6.42}\selectfont
\setlength{\tabcolsep}{2.6pt}
\renewcommand{\arraystretch}{1.14}
\setlength{\aboverulesep}{0.30ex}
\setlength{\belowrulesep}{0.30ex}

\begin{threeparttable}
\begin{tabularx}{\textwidth}{@{}>{\RaggedRight\arraybackslash}X
                                >{\centering\arraybackslash}m{0.11\textwidth}
                                >{\centering\arraybackslash}m{0.14\textwidth}
                                >{\centering\arraybackslash}m{0.14\textwidth}
                                >{\centering\arraybackslash}m{0.14\textwidth}
                                >{\centering\arraybackslash}m{0.14\textwidth}@{}}
\toprule
Metric &
\makecell[c]{\routecont\hspace{0.25em}FS-FSD\\[-1pt]\scriptsize native} &
\makecell[c]{\routemask\hspace{0.25em}YOLO26m-seg\\[-1pt]\scriptsize RLE-full} &
\makecell[c]{\routemask\hspace{0.25em}YOLO26m-seg\\[-1pt]\scriptsize RLE-crop} &
\makecell[c]{\routemask\hspace{0.25em}YOLO26m-seg\\[-1pt]\scriptsize Poly-256} &
\makecell[c]{\routemask\hspace{0.25em}YOLO26m-seg\\[-1pt]\scriptsize Fourier-fit} \\
\midrule

\rowcolor{summarygray}
\rule{0pt}{1.05em}\textit{Stage 1: Direct footprint} & & & & & \\
\#Pred
& 1,660
& 1,849
& 1,849
& 1,849
& 1,849 \\
Payload / defect (B) $\downarrow$
& \bestnum{132.0}
& 8,102.8
& 5,998.3
& 1,024.0
& \bestnum{132.0} \\
Record / defect (B) $\downarrow$
& \bestnum{479.5}
& 8,480.4
& 6,437.7
& 1,353.7
& 488.7 \\

\midrule

\rowcolor{summarygray}
\rule{0pt}{1.05em}\textit{Stage 2: Route cost} & & & & & \\
Archive route overhead / defect (ms, excl.\ infer) $\downarrow$
& \bestnum{1.012}
& 31.49
& 33.93
& 21.45
& 21.66 \\
Route-to-usable / defect (ms, excl.\ infer) $\downarrow$
& \bestnum{1.165}
& 75.15
& 36.17
& 21.54
& 21.84 \\
Route-to-usable / defect (ms, incl.\ infer) $\downarrow$
& \bestnum{5.548}
& 80.50
& 41.51
& 26.89
& 27.19 \\
\makecell[l]{Route throughput\\[-1pt](defects/s, excl.\ infer) $\uparrow$}
& \bestnum{858.3}
& 13.31
& 27.65
& 46.43
& 45.78 \\
\bottomrule
\end{tabularx}

\vspace{1.4pt}

\begin{tablenotes}[flushleft]
\fontsize{5.12}{5.84}\selectfont
\item RLE-full stores a full-image uint32 run-length-encoded mask; RLE-crop stores a foreground-tight cropped uint32 RLE mask; Poly-256 stores the largest external contour resampled to 256 points; and Fourier-fit stores an order-16 Fourier descriptor fitted from the mask contour. Because FS-FSD and YOLO26m-seg archive different numbers of predictions, all route-level latency metrics normalize archive-stage time by the mean number of stored predictions per image for the corresponding method. Raw recovery-only latency is reported for diagnosis only and is not used as the principal route-level conclusion, because routes such as Poly-256 and Fourier-fit may shift part of the representation-conversion cost to the archive stage. The recovered polygons and descriptors are image-space quantities. Best values in each row are highlighted. Values are rounded for compact display.
\end{tablenotes}
\end{threeparttable}
\end{table*}

The footprint results in Table~\ref{tab:archive_fair_routes} show that FS-FSD is the most compact route among the compared storage representations. Its native payload is fixed at \textbf{132.0\,B/defect}, whereas the YOLO-side routes require \textbf{8,102.8\,B/defect} for RLE-full, \textbf{5,998.3\,B/defect} for RLE-crop, and \textbf{1,024.0\,B/defect} for Poly-256. Fourier-fit reaches payload parity because it ultimately stores an order-16 Fourier representation with the same coefficient dimensionality as FS-FSD. The same pattern remains at the full-record level: FS-FSD requires \textbf{479.5\,B/defect}, compared with \textbf{8,480.4\,B/defect}, \textbf{6,437.7\,B/defect}, and \textbf{1,353.7\,B/defect} for RLE-full, RLE-crop, and Poly-256, respectively. Even against Fourier-fit, FS-FSD retains a small full-record advantage (\textbf{479.5} vs.\ \textbf{488.7\,B/defect}). This indicates that compact coefficient dimensionality is helpful, while the native representation route also affects the final archive footprint.

\begin{figure*}[pos=h]
    \centering
    \includegraphics[width=\textwidth]{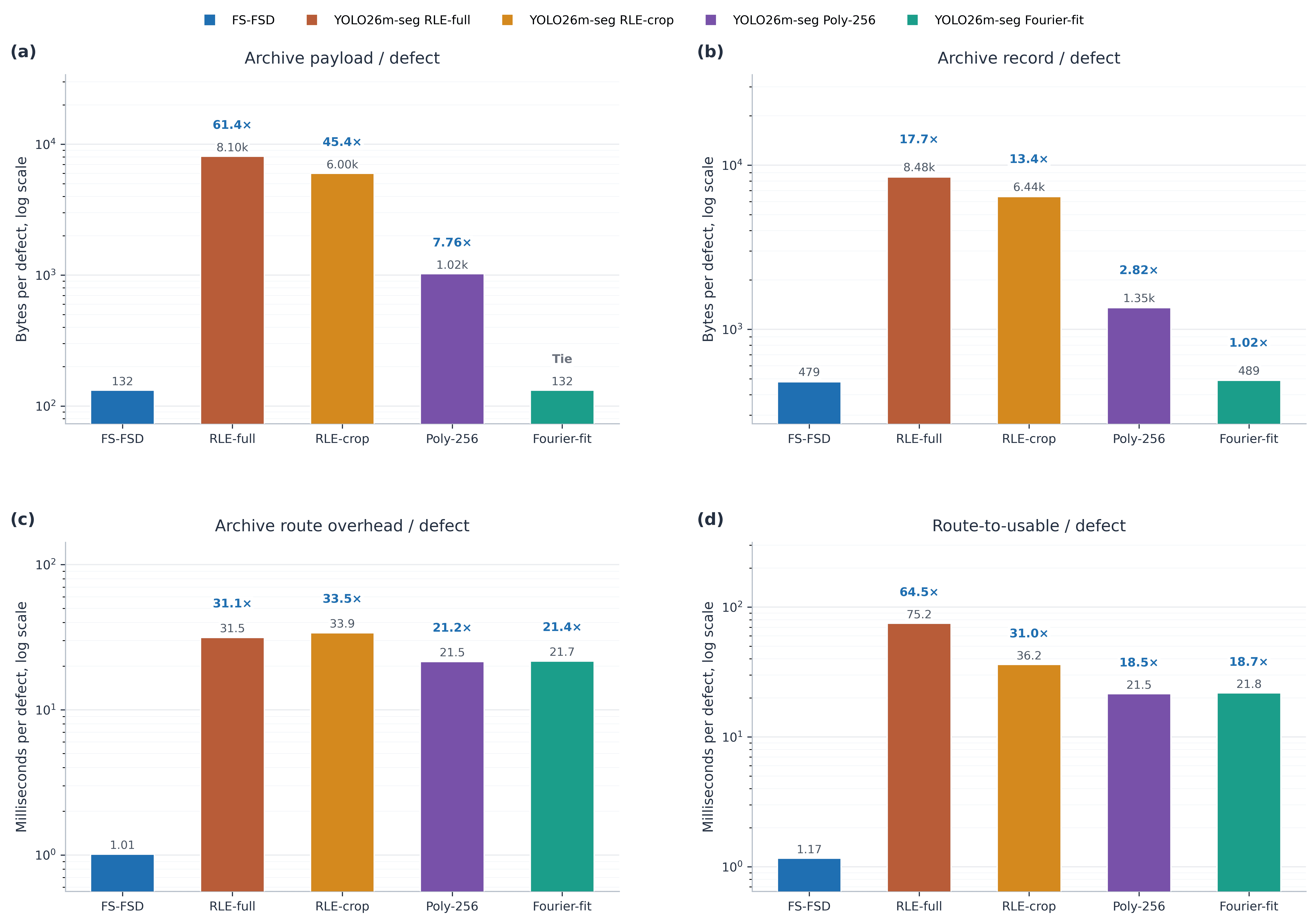}
    \caption{Method-centric grouped-bar comparison of FS-FSD and YOLO-side storage routes under route-comparable metrics. The x-axis lists FS-FSD, RLE-full, RLE-crop, Poly-256, and Fourier-fit. Panels (a)--(d) show payload per defect, record size per defect, archive route overhead per defect excluding inference, and route-to-usable latency per defect excluding inference, respectively.}
    \label{fig:archive_fair_core}
\end{figure*}

The route-cost results further show that the compactness of FS-FSD is accompanied by lower archive-side overhead. FS-FSD requires only \textbf{1.012\,ms/defect} of archive route overhead excluding inference. In contrast, the corresponding YOLO-side overheads are \textbf{31.49\,ms/defect} for RLE-full, \textbf{33.93\,ms/defect} for RLE-crop, \textbf{21.45\,ms/defect} for Poly-256, and \textbf{21.66\,ms/defect} for Fourier-fit. Thus, before recovery begins, the route-specific cost of turning a YOLO mask prediction into an archived reusable representation is already more than twenty times higher than that of FS-FSD in this benchmark. RLE-crop is also informative: although it reduces payload size relative to RLE-full, its archive route overhead is slightly higher (\textbf{33.93} vs.\ \textbf{31.49\,ms/defect}). This suggests that foreground-tight cropping reduces raster footprint but does not remove the conversion and serialization burden of raster-derived storage.

The archive-side burden propagates into the complete route from raw prediction to recovered image-space polygonal record. The route-to-usable latency excluding inference is \textbf{1.165\,ms/defect} for FS-FSD, compared with \textbf{75.15\,ms/defect} for RLE-full, \textbf{36.17\,ms/defect} for RLE-crop, \textbf{21.54\,ms/defect} for Poly-256, and \textbf{21.84\,ms/defect} for Fourier-fit. These correspond to latency ratios of \textbf{64.5}$\times$, \textbf{31.0}$\times$, \textbf{18.5}$\times$, and \textbf{18.7}$\times$ relative to FS-FSD, respectively. The same ranking is preserved when inference is included: FS-FSD requires \textbf{5.548\,ms/defect}, whereas the four YOLO-side routes require \textbf{80.50}, \textbf{41.51}, \textbf{26.89}, and \textbf{27.19\,ms/defect}. Under this full route setting, the corresponding latency ratios remain \textbf{14.5}$\times$, \textbf{7.48}$\times$, \textbf{4.85}$\times$, and \textbf{4.90}$\times$. In throughput terms, FS-FSD reaches \textbf{858.3 defects/s} under the route-comparable metric excluding inference, whereas the corresponding YOLO-side routes reach \textbf{13.31}, \textbf{27.65}, \textbf{46.43}, and \textbf{45.78 defects/s}.

Fig.~\ref{fig:archive_fair_core} presents the same comparison in a method-centric grouped-bar layout, placing FS-FSD and the four YOLO-side routes on a shared x-axis. This organization makes the route ranking easier to interpret under the same cost basis. Panels (a) and (b) confirm that FS-FSD has the smallest storage footprint among the compared routes, except for payload parity with Fourier-fit. Panels (c) and (d) show that the main burden of YOLO-side routes lies not only in recovery, but also in the conversion chain jointly formed by archive-stage and recovery-stage operations.

This distinction is especially important for \textbf{Poly-256}. If only recovery-stage timing is considered, Poly-256 may appear favorable because the polygon has already been extracted and stored during archival. The diagnostic raw recovery-only latency is \textbf{0.0867\,ms/defect} for Poly-256, compared with \textbf{0.1533\,ms/defect} for FS-FSD. However, this local advantage disappears once archive-stage polygon materialization is attributed back to the route. Under the route-comparable metric, Poly-256 requires \textbf{21.54\,ms/defect} for route-to-usable latency excluding inference, which is \textbf{18.5}$\times$ higher than FS-FSD. This explains why route-comparable metrics are used as the basis for workflow-level interpretation, while raw recovery-only latency is treated only as a diagnostic quantity.

\begin{figure*}[pos=h]
    \centering
    \includegraphics[width=\textwidth]{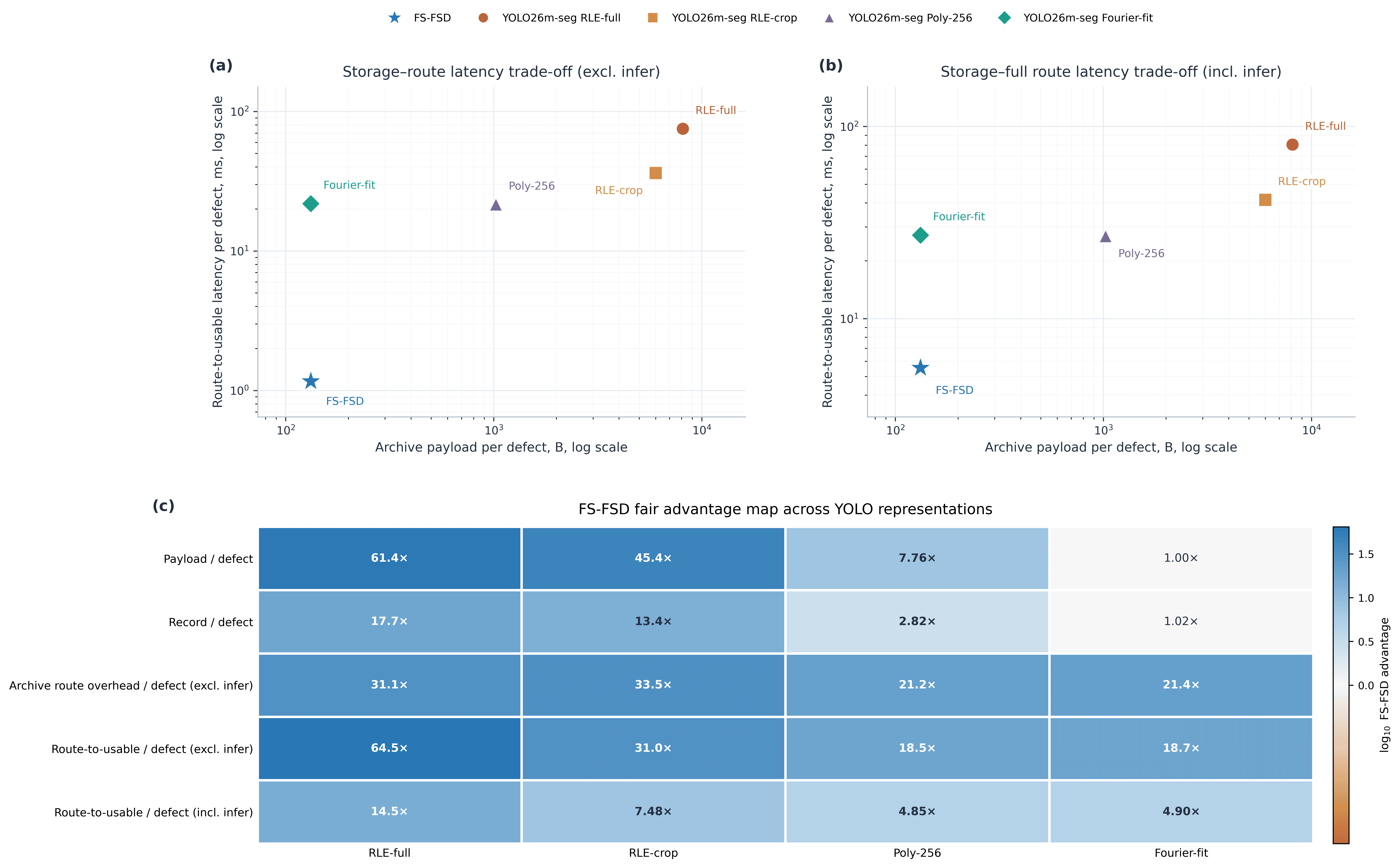}
    \caption{Storage--latency trade-offs and FS-FSD advantage map across YOLO-side storage routes under route-comparable metrics. In panels (a) and (b), the lower-left region is preferable because it indicates both lower archive footprint and lower route cost. In panel (c), the ratio direction is chosen so that values greater than 1 favor FS-FSD.}
    \label{fig:archive_fair_tradeoff}
\end{figure*}

The storage-latency trade-off is summarized in Fig.~\ref{fig:archive_fair_tradeoff}. Fig.~\ref{fig:archive_fair_tradeoff}(a) plots archive payload per defect against route-to-usable latency excluding inference, and Fig.~\ref{fig:archive_fair_tradeoff}(b) plots archive payload per defect against the corresponding full route latency including inference. In both panels, the favorable operating region is the lower-left corner, corresponding to lower storage burden and lower route cost. FS-FSD lies closest to this region among the compared routes. RLE-full and RLE-crop remain far from this region, indicating both large footprint and high route cost. Poly-256 moves leftward relative to the raster routes in storage terms and reduces part of the recovery-side burden, but it remains substantially above FS-FSD once archive-stage polygon materialization is counted. Fourier-fit achieves payload parity with FS-FSD, yet it still occupies a higher latency position because the compact vector representation is obtained only after mask-side contour extraction and Fourier fitting.

Fig.~\ref{fig:archive_fair_tradeoff}(c) summarizes the same comparison as an FS-FSD advantage map. For lower-is-better metrics, the plotted factor is defined as the YOLO-side value divided by the FS-FSD value. For higher-is-better throughput, the ratio direction is reversed so that values greater than 1 consistently favor FS-FSD. Across the principal route-comparable metrics and the four YOLO-side routes, FS-FSD attains the most favorable value in \textbf{23} of the \textbf{24} comparable cases, with the remaining case corresponding to payload parity against \textbf{Fourier-fit}. No YOLO-side route provides a better value on the principal route-comparable metrics in this benchmark. This result reflects the total route burden that an archive-and-recovery workflow would need to handle, rather than only the local cost of reopening an already materialized representation.

From the standpoint of inspection data workflows, these findings indicate a practical advantage of contour-native archival in the tested setting. In such workflows, model outputs may need to be stored after inference and later reopened for review, visualization, descriptor computation, or data transfer. Therefore, in addition to prediction accuracy, the archive and recovery cost of a representation is also relevant to post-prediction data handling. The present proof-of-concept benchmark shows that FS-FSD reduces this post-prediction overhead by storing a compact native vector representation and recovering the required image-space polygon with low route-level latency.

Overall, the database-oriented validation supports the practical value of FS-FSD from an archive-and-recovery perspective. Within an image-space benchmark using SQLite storage and a fixed polygon recovery target, the contour-native representation of FS-FSD provides the smallest archive footprint, the lowest route-level overhead, and the fastest conversion from stored prediction to recovered polygonal record among the compared routes. These results support the use of contour-native prediction not only as an accuracy-oriented modeling choice, but also as a compact and reusable image-space data carrier for post-prediction inspection data handling. They do not, however, constitute a claim of physical-space measurement or full inspection-management integration.

\subsubsection{WebGL-based visualization of archived image-space damage records}
\label{sec:webgl_visualization}

To complement the preceding quantitative archive-and-recovery benchmark, this subsection introduces a lightweight Web Graphics Library (WebGL)-based visualization prototype for reviewing archived FS-FSD damage records in a browser. Rather than proposing a new visualization algorithm or a complete inspection management platform, the prototype provides a proof-of-concept check of whether the archived contour-native representation can be reopened, reconstructed, displayed, and inspected as an image-space polygonal record after inference.

In inspection data workflows, compact records are more useful if they can also be reviewed together with the original image context. A storage-efficient representation would have limited practical value if it required heavy post-processing or pre-rendered overlays before visual inspection. The WebGL prototype is therefore used to verify the operability of the FS-FSD archive format at the browser side: the archived Fourier descriptor is decoded on demand, reconstructed into an image-coordinate polygon, rendered over the corresponding image, and displayed together with semantic and descriptor information. All geometry shown in this subsection remains in image space. The displayed polygons and derived descriptors should therefore be interpreted as image-space quantities rather than physical-space measurements or metric-scale defect dimensions.

\begin{figure*}[pos=h]
    \centering
    \includegraphics[width=\textwidth]{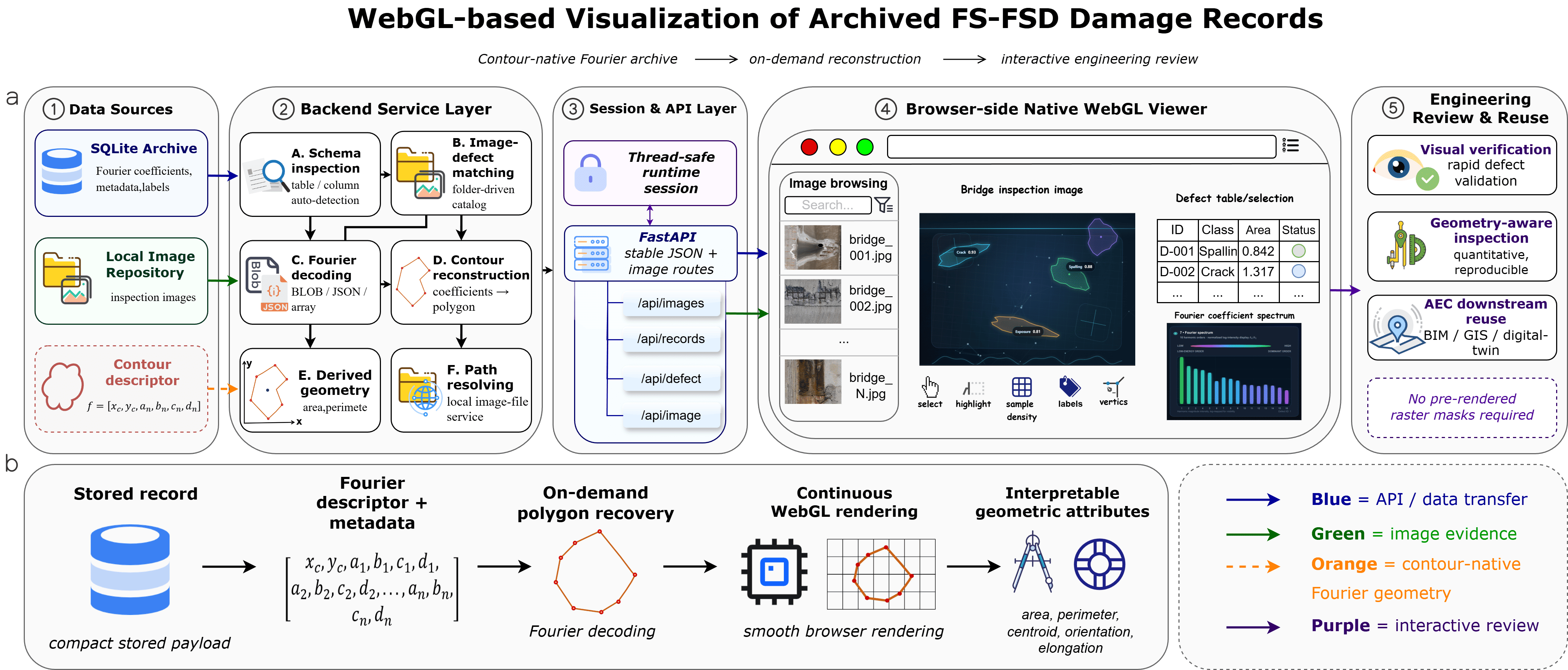}
    \caption{Workflow from archived FS-FSD image-space damage records to browser-side review.}
    \label{fig:webgl_workflow}
\end{figure*}

Fig.~\ref{fig:webgl_workflow} provides an overview of the archive-to-review workflow. The upper part summarizes the prototype architecture, including the SQLite archive, the local image repository, the backend service layer, the session and application programming interface (API) layer, the browser-side native WebGL viewer, and the final browser-side review interface. The lower part illustrates the record-level transformation of a single archived damage instance. A compact stored payload containing Fourier coefficients and metadata is decoded on demand, reconstructed as a polygon in image coordinates, rendered in the browser, and used to compute image-space geometric descriptors when needed. The color-coded flows distinguish API/data transfer, image evidence, contour-native Fourier geometry, and interactive review. This emphasizes that the archived FS-FSD record is not only a static visual label, but a compact record that can be recovered and inspected as an image-space contour.

Instead of relying on pre-generated raster overlays, the prototype adopts a lightweight local service--browser architecture. The backend service layer reads the SQLite archive, inspects the database schema, loads image and defect records, decodes Fourier coefficients, reconstructs polygons on demand, computes image-space descriptors, and resolves local image-file paths. The session and API layer exposes stable image and defect-record interfaces to the browser, while the frontend uses native WebGL for interactive image-and-contour rendering. In practical use, the visible image list is driven by a user-selected local image repository, and the archived defect records stored in the database are attached to those images through record-level matching. The project is available at: \url{https://github.com/wangzai822/Frequency-supervised-Fourier-Series-Detection}.

The key point of the prototype is how WebGL visualization is connected to the contour-native archive. Because FS-FSD stores compact Fourier descriptors rather than dense raster masks, only a small coefficient vector and a limited amount of metadata are required to reconstruct a defect contour for browser-side rendering. Therefore, the viewer does not need to transmit high-resolution binary masks or maintain a separate pre-rendered raster overlay for each defect instance. The same archived descriptor can also be reconstructed at different polygon sampling densities without changing the stored payload. Within the scope of image-space visualization, this provides a lightweight path from archived vector coefficients to interactive contour display.

\begin{figure*}[pos=h]
    \centering
    \includegraphics[width=\textwidth]{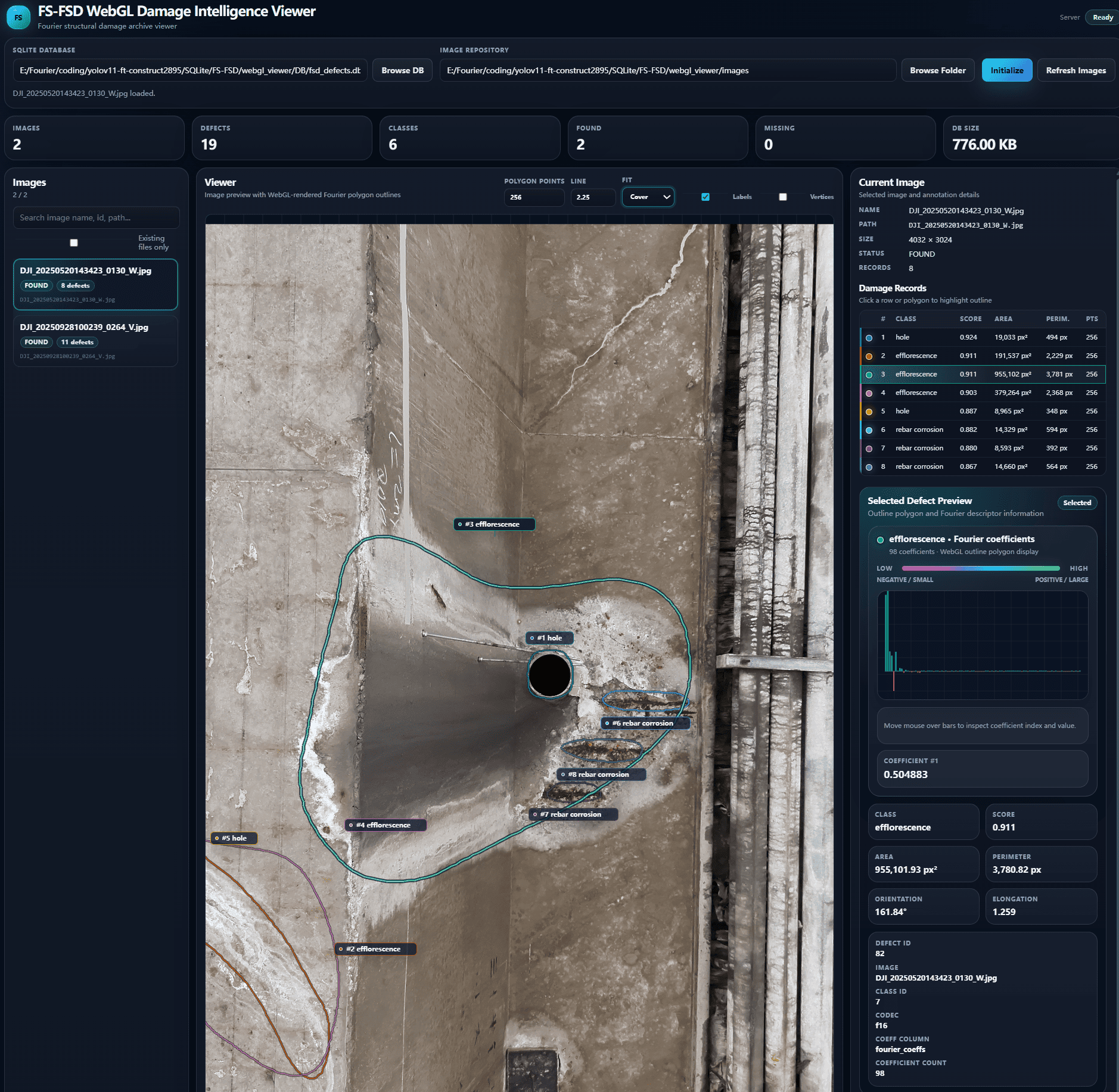}
    \caption{WebGL prototype interface for reviewing archived FS-FSD image-space damage records.}
    \label{fig:webgl_damage_viewer}
\end{figure*}

Fig.~\ref{fig:webgl_damage_viewer} shows the actual prototype interface used for reviewing archived FS-FSD damage records. The interface places image evidence, reconstructed contours, defect categories, confidence scores, image-space geometric descriptors, and Fourier coefficient spectra within a unified visual context. In this way, each archived defect can be inspected at the image level, polygon level, and descriptor level. The viewer supports image browsing, defect selection, linked contour highlighting, coefficient-spectrum inspection, and descriptor-aware review. These functions demonstrate that the archived FS-FSD record can be reopened and examined interactively, rather than serving as a complete inspection management system.

The displayed quantities such as area, perimeter, centroid, orientation, and elongation are derived descriptors computed on demand from the reconstructed polygon. They are included to support visual checking and interpretation of the archived image-space record. They are not additional direct prediction outputs of FS-FSD. They are also not physical-space measurements, because no camera calibration, scale recovery, surface registration, or component-level coordinate mapping is performed in this prototype. This design is consistent with the database benchmark in Section~\ref{subsubsec:database_archive_recovery}, where the native model output remains the Fourier coefficient vector and downstream descriptors are recovered from that representation only when needed.

The prototype also illustrates a practical distinction between contour-native archival and raster-derived archival. Raster-mask records can certainly be visualized, but in an archive-review setting they typically require storing or transmitting mask payloads, converting masks to contours, or preparing raster overlays before display. In contrast, the FS-FSD record directly stores the compact descriptor from which the contour can be reconstructed. The same stored descriptor supports image overlay rendering, polygon resampling, coefficient-spectrum display, and descriptor computation. Thus, the archived defect behaves as a structured image-space record rather than only as a bitmap-like attachment to an image.

Within the limited scope of this proof-of-concept prototype, three observations can be made. First, archived FS-FSD records can be decoded and visualized directly as image-space contours in a browser environment. Second, the compact descriptor format avoids the need to transmit dense masks or maintain separate raster overlays for each defect instance in this prototype setting. Third, semantic information, confidence scores, Fourier spectra, and image-space descriptors can be displayed around the same archived contour representation, which makes the record easier to inspect and verify. These observations support the data-handling usefulness of the contour-native representation, while remaining within image-space visualization rather than validating physical-scale measurement or integration with an external inspection management platform.

Taken together with the route-level archive benchmark, the workflow overview in Fig.~\ref{fig:webgl_workflow} and the prototype interface in Fig.~\ref{fig:webgl_damage_viewer} add a complementary usability check to the evaluation of FS-FSD. The archive benchmark shows that the proposed representation is efficient in storage footprint and route-level recovery cost. The WebGL prototype further shows that the same representation can be reopened and rendered interactively as an image-space polygonal record in a browser. Therefore, the value demonstrated here is not only higher contour prediction accuracy, but also a compact and recoverable representation that is convenient for post-prediction image-space data handling.

\subsection{Edge-side deployment and on-device computational efficiency}
\label{subsec:edge_deployment}

The database-oriented benchmark above evaluates FS-FSD after prediction, focusing on how compactly and efficiently prediction records can be archived and recovered. A complementary question is whether the same contour-native formulation remains practical during prediction, especially on embedded hardware with limited memory and operator support. This subsection therefore evaluates the on-device executability and computational efficiency of FS-FSD on an edge AI platform.

This validation is a deployment-oriented proof of concept that isolates model-side inference behavior on the target device. The reported latency and FPS correspond to steady-state single-image inference after model loading. They do not include camera acquisition, network communication, database insertion, browser visualization, or other application-level operations. The geometric accuracy values referenced in this subsection are the image-space polygon-space mAP values reported in the preceding experiments.

\begin{table}[pos=h]
\centering
\caption{Edge deployment platform used in this study.}
\label{tab:edge_platform}

\fontsize{5.95}{6.85}\selectfont
\setlength{\tabcolsep}{4.2pt}
\renewcommand{\arraystretch}{1.10}
\setlength{\aboverulesep}{0.30ex}
\setlength{\belowrulesep}{0.30ex}

\begin{threeparttable}
\begin{tabularx}{\linewidth}{@{}>{\RaggedRight\arraybackslash}p{0.27\linewidth}>{\RaggedRight\arraybackslash}X@{}}
\toprule
\textbf{Component} & \textbf{Specification} \\
\midrule
Platform & Atlas 200I DK A2 developer kit \\
AI processor & Ascend 310-series AI processor \\
Processor configuration & 1 $\times$ DaVinci V300 AI core; 4 $\times$ Taishan V200M CPU cores \\
Compute capability & FP16: 4 TFLOPS; INT8: 8 TOPS \\
Memory & 4 GB LPDDR4X, 3200 Mbps, 64-bit, ECC \\
Deployment pipeline & ONNX export followed by OM compilation \\
\bottomrule
\end{tabularx}
\end{threeparttable}
\end{table}

The deployment platform used in this study is an \textbf{Atlas 200I DK A2} developer kit. As summarized in Table~\ref{tab:edge_platform}, the platform is built around an \textbf{Ascend 310-series AI processor}, provides \textbf{4 TFLOPS} FP16 and \textbf{8 TOPS} INT8 compute capability, and is equipped with \textbf{4 GB LPDDR4X} memory. Compared with desktop GPU evaluation, this platform provides a more constrained execution environment in terms of memory budget, compilation requirements, supported operators, and runtime scheduling.

To examine deployment behavior under different capacity--order trade-offs, four FS-FSD variants are deployed: \textbf{FS-FSD-S ($n_f=16$)}, \textbf{FS-FSD-M ($n_f=8$)}, \textbf{FS-FSD-M ($n_f=12$)}, and \textbf{FS-FSD-M ($n_f=16$)}. \textit{YOLO11m-seg} and \textit{YOLO26m-seg} are used as strong raster-based baselines. All models follow the same resized input setting used in their main experiments. They are first exported to ONNX and then compiled into the hardware-native OM format, so that runtime comparison is conducted under a unified device execution environment.

For embedded NPUs, desktop-side indicators such as parameter count or GFLOPs do not fully determine practical runtime, because graph compilation, operator scheduling, memory access, and device-specific kernel support can substantially affect latency. Therefore, this subsection reports deployment-oriented metrics that are directly measured on the device: \emph{load time}, \emph{steady-state latency per image}, and \emph{FPS}. In addition, the sizes of the PyTorch checkpoint, ONNX file, and compiled OM artifact are reported to document model conversion and deployment overhead.

\newcommand{\deploysecond}[1]{%
  {\setlength{\fboxsep}{0.18pt}%
   \colorbox{geombestbg}{\textcolor{geombestfg}{\strut #1}}}%
}

\begin{table}[pos=h]
\centering
\caption{On-device deployment results on the Atlas 200I DK A2.}
\label{tab:edge_deploy}

\fontsize{5.72}{6.52}\selectfont
\setlength{\tabcolsep}{2.5pt}
\renewcommand{\arraystretch}{1.10}
\setlength{\aboverulesep}{0.28ex}
\setlength{\belowrulesep}{0.28ex}
\setlength{\cmidrulesep}{0.10ex}

\begin{threeparttable}
\begin{tabular*}{\linewidth}{@{\extracolsep{\fill}}lcccccc@{}}
\toprule
\textbf{Method} &
\makecell[c]{\textbf{Load} \\ \textbf{(s)} $\downarrow$} &
\makecell[c]{\textbf{Latency} \\ \textbf{(s/image)} $\downarrow$} &
\makecell[c]{\textbf{FPS} \\ $\uparrow$} &
\makecell[c]{\textbf{PT} \\ \textbf{(MB)} $\downarrow$} &
\makecell[c]{\textbf{ONNX} \\ \textbf{(MB)} $\downarrow$} &
\makecell[c]{\textbf{OM} \\ \textbf{(MB)} $\downarrow$} \\
\midrule

\routemask\hspace{0.25em}YOLO11m-seg
& \bestnum{1.696}
& \bestnum{0.2123}
& \bestnum{4.706}
& \bestnum{45.3}
& \bestnum{85.8}
& \bestnum{46.8} \\

\routemask\hspace{0.25em}YOLO26m-seg
& 4.131
& 0.2321
& 4.308
& \deploysecond{54.6}
& \deploysecond{94.6}
& \deploysecond{49.4} \\

\midrule

\rowcolor{summarygray}
\routecont\hspace{0.25em}FS-FSD-S ($n_f=16$)
& 2.199
& 0.4778
& 2.093
& 133.5
& 266.4
& 135.6 \\

\rowcolor{summarygray}
\routecont\hspace{0.25em}FS-FSD-M ($n_f=8$)
& \deploysecond{1.849}
& \deploysecond{0.2279}
& \deploysecond{4.388}
& 93.2
& 185.6
& 95.6 \\

\rowcolor{summarygray}
\routecont\hspace{0.25em}FS-FSD-M ($n_f=12$)
& 2.330
& 0.4029
& 2.482
& 130.2
& 259.7
& 132.6 \\

\rowcolor{summarygray}
\routecont\hspace{0.25em}FS-FSD-M ($n_f=16$)
& 2.291
& 0.4969
& 2.013
& 175.2
& 349.5
& 170.0 \\
\bottomrule
\end{tabular*}

\vspace{1.2pt}

\begin{tablenotes}[flushleft]
\fontsize{5.12}{5.86}\selectfont
\item Load time denotes one-off model initialization and graph loading on the device. Latency and FPS denote steady-state single-image inference after model loading. They do not include camera capture, network transfer, database insertion, browser visualization, or other application-level operations. PT, ONNX, and OM report the sizes of the PyTorch checkpoint, exported ONNX file, and hardware-native compiled model, respectively. Blue boxes denote the best values and lavender boxes denote the second-best values under each metric direction. Gray row shading denotes FS-FSD variants. All models are executed in OM format on the same edge platform.
\end{tablenotes}
\end{threeparttable}
\end{table}

Table~\ref{tab:edge_deploy} reports the on-device deployment results. For visual guidance, blue-highlighted cells indicate the best value under the corresponding metric direction, lavender-highlighted cells indicate the second-best value, and gray row shading marks the FS-FSD variants. Overall, the results show that contour-native prediction is feasible on the tested embedded platform. Under an appropriate Fourier-order setting, FS-FSD reaches on-device throughput close to the mask-based baselines while retaining much stronger image-space polygon accuracy in the preceding evaluation.

Among the deployed FS-FSD variants, \textbf{FS-FSD-M ($n_f=8$)} provides the most favorable deployment trade-off. It requires \textbf{0.2279 s/image} and reaches \textbf{4.388 FPS}, ranking second among all deployed models in both latency and FPS. \textbf{YOLO11m-seg} remains the fastest model in absolute terms, with \textbf{0.2123 s/image} and \textbf{4.706 FPS}. However, the gap between YOLO11m-seg and FS-FSD-M ($n_f=8$) is modest in this test: YOLO11m-seg has only \textbf{6.8\%} lower latency and \textbf{7.2\%} higher FPS. This indicates that the contour-native formulation does not necessarily introduce a prohibitive inference-time penalty on the tested embedded hardware.

A more relevant comparison is with \textbf{YOLO26m-seg}, which is the strongest raster-based baseline in the main polygon-space shape evaluation. In this case, \textbf{FS-FSD-M ($n_f=8$)} is slightly faster on the edge device: its latency is \textbf{1.8\%} lower and its FPS is \textbf{1.9\%} higher. This near-parity in runtime is important because the corresponding image-space polygon accuracy differs substantially. As reported earlier in Tables~\ref{tab:overall_main} and~\ref{tab:ablation_ftfsd}, \textit{YOLO26m-seg} reaches \textbf{49.68/24.16} under the Mask$\rightarrow$Polygon route, whereas \textbf{FS-FSD ($n_f=8$)} reaches \textbf{85.48/55.28} under the Contour$\rightarrow$Polygon route. Relative to YOLO26m-seg, this corresponds to improvements of \textbf{72.1\%} in $mAP_{50}$ and \textbf{128.8\%} in $mAP_{50{:}95}$. Thus, on the tested edge platform, FS-FSD-M ($n_f=8$) provides a more favorable accuracy--runtime trade-off than the strongest raster baseline considered here.

The deployment results also show that the Fourier order can be used as a practical control variable for edge deployment. As shown in Table~\ref{tab:ablation_ftfsd}, reducing the order from $n_f=16$ to $n_f=8$ changes $mAP_{50}$ by only \textbf{1.51} points, while $mAP_{50{:}95}$ remains essentially unchanged and is slightly higher at \textbf{55.28}. On the edge device, however, the runtime effect is much larger. Compared with \textbf{FS-FSD-M ($n_f=16$)}, \textbf{FS-FSD-M ($n_f=8$)} reduces latency from \textbf{0.4969} to \textbf{0.2279 s/image}, corresponding to a reduction of approximately \textbf{54.1\%}. Its FPS increases from \textbf{2.013} to \textbf{4.388}, corresponding to an increase of approximately \textbf{118.0\%}. Even compared with \textbf{FS-FSD-M ($n_f=12$)}, the $n_f=8$ variant reduces latency by \textbf{43.4\%} and increases FPS by \textbf{76.8\%}, while changing $mAP_{50}$ by only \textbf{0.84} points and keeping $mAP_{50{:}95}$ at \textbf{55.28}. These results suggest that reducing the harmonic order is an effective way to adapt FS-FSD to edge-side latency constraints without causing a corresponding loss in polygon-space accuracy.

Another observation is that the compiled FS-FSD artifacts are larger than those of the YOLO baselines. For example, \textbf{FS-FSD-M ($n_f=8$)} produces a \textbf{95.6 MB} OM model, compared with \textbf{49.4 MB} for \textbf{YOLO26m-seg}. This is approximately \textbf{93.5\%} larger. Nevertheless, the measured runtime remains comparable on the tested device. This indicates that executable model size alone does not fully determine NPU inference speed. Operator scheduling, graph structure, memory access, and hardware compilation can also affect the realized latency. At the same time, executable artifact size should be distinguished from prediction-record size. Although FS-FSD does not produce the smallest compiled model, the preceding archive benchmark shows that its post-prediction records are much more compact because the model directly outputs Fourier coefficients rather than dense raster masks. Therefore, the deployment result and the archive result should be interpreted as complementary: the model artifact is larger than the YOLO artifacts, but the recovered prediction record is substantially smaller and faster to archive and recover in the tested workflow.

Taken together, the two proof-of-concept validations provide complementary evidence for the practical feasibility of FS-FSD within the tested settings. The database-oriented benchmark shows that FS-FSD is compact and efficient after prediction, at the storage and recovery stage. The edge-device benchmark shows that FS-FSD-M ($n_f=8$) is also executable with competitive steady-state inference speed on the tested embedded platform. These results indicate that FS-FSD can provide both strong image-space polygon accuracy and practical on-device inference behavior under the evaluated conditions, while full field deployment would require additional system-level integration beyond the scope of this component-level validation.

\section{Conclusions}
\label{sec:conclusions}

This study addressed a representation problem in AI-assisted bridge defect detection and recognition. Instead of focusing only on recognition accuracy, it examined which output form is more suitable when predicted defects must remain useful after inference for boundary recovery, geometric evaluation, compact archival, retrieval, transmission, visualization, and repeated access. The proposed \textbf{FS-FSD} addresses this problem through a contour-native formulation that directly regresses Fourier descriptors, compares heterogeneous outputs in a unified polygon space, and evaluates the predicted representation in several post-prediction data-handling settings, including archive-and-recovery benchmarking, browser-side visualization, and on-device inference testing.

The results indicate that, in AI-assisted bridge defect detection and recognition, the output representation should not be treated as a secondary implementation detail. Under the image-space setting considered in this paper, FS-FSD is not merely another detector or segmenter. It provides a compact contour-native defect record in which boundary prediction, polygon-space evaluation, storage, recovery, visualization, and device-side execution can be considered as connected stages of the same information pipeline. At the same time, the workflow-related experiments are component-level proof-of-concept tests. They do not constitute a complete field inspection system, do not provide physical-scale defect measurement, and do not claim bridge-component registration or integration with external asset-management platforms.

The main findings are as follows.

\begin{enumerate}
    \item From the evaluation perspective, box mAP, mask mAP, and contour-native prediction metrics correspond to different geometric semantics and therefore do not directly answer the same question about defect boundary recovery. The experiments show that rankings obtained from task-native metrics can change after box, mask, and contour outputs are converted into a common polygon representation. This is important for bridge defects because many categories, such as cracks, spalling regions, seepage stains, rust regions, and honeycombing areas, have irregular boundaries that cannot be fully characterized by rectangular supports or raster scores alone. The polygon-space protocol used in this paper therefore provides a more representation-consistent basis for comparing box-, mask-, and contour-based methods. It also offers a possible benchmark structure for future studies that evaluate heterogeneous defect representations. Its conclusions should be interpreted together with the specified conversion rules and route assignments used in this paper. In particular, fair cross-paradigm comparison depends on transparent output conversion, and future benchmarks could further report sensitivity to contour extraction, polygon simplification, and annotation styles.

    \item Under the unified polygon-space evaluation, FS-FSD shows an overall image-space geometric advantage over the representative detection and segmentation baselines evaluated in this study, but this advantage should not be interpreted as uniform dominance on every category-level AP or boundary submetric. The importance of this result is that the evaluation no longer asks only whether a model performs well in its native output format. Instead, it asks whether the final polygon record, after conversion if needed, can accurately recover the image-space defect boundary. From this perspective, box-based methods are inherently limited because an irregular defect must be represented by a rectangular support region. Mask-based methods preserve more boundary information, but the final polygon quality still depends on raster resolution, thresholding, contour extraction, and polygon simplification. FS-FSD avoids this additional shape-recovery step by directly predicting a Fourier contour descriptor that can be reconstructed as a polygon.

    The metric-level results should therefore be interpreted according to the type of geometric error being measured. Boundary F-score evaluates whether boundary points fall within a fixed tolerance, and its value can be sensitive to the tolerance setting and to the contour obtained from mask extraction. In the current S2P matched-TP comparison, FS-FSD achieves the highest micro-averaged B-F1, although mask-based baselines remain competitive in several class-wise B-F1 entries, such as crack and honeycombing. Distance-based boundary metrics, in contrast, evaluate how close the recovered boundary is to the reference boundary as a whole, and area error reflects whether the global defect extent is preserved. FS-FSD is particularly strong on these aspects, achieving the lowest S2P matched-TP CD and A-Err both in the micro average and across all defect categories. Perimeter error provides a different and complementary view because it measures boundary length and is sensitive to local oscillations, polygon simplification, contour smoothing, and high-frequency boundary details. In the current S2P matched-TP results, FS-FSD also obtains the lowest P-Err, indicating that its Fourier representation preserves boundary length while maintaining compact and stable shape recovery. Therefore, the main implication is not that FS-FSD removes all category-specific exceptions, but that contour-native and frequency-aware prediction provides a better overall representation-level trade-off for bridge defect detection and recognition when the target output is a recoverable image-space boundary record.

    \item The ablation results indicate that the effectiveness of FS-FSD comes from the coordinated design of representation, supervision, and optimization, rather than from the use of Fourier descriptors alone. U-DFS is the dominant component because it supervises predicted Fourier coefficients directly in the unit-grid frequency-domain space and avoids unstable reliance on inverse reconstruction during training. FT norm improves strict polygon-space accuracy by balancing coefficient errors across harmonic orders. CHPA reduces spurious coefficient discrepancies caused by the arbitrary starting point of a closed contour. These results show that Fourier contour learning is not simply a replacement of one output format with another. It becomes effective only when the output parameterization, supervision space, and treatment of geometric invariance are aligned with the internal structure of the representation.

    This observation has broader methodological relevance for geometry-sensitive detection tasks. When the final target is an irregular defect boundary, performance should not be pursued only through larger backbones, stronger decoders, or more complex post-processing. The representation itself becomes part of the model design. A mismatch among output parameterization, loss space, and contour invariance can reduce both prediction accuracy and the reliability of the recovered record. FS-FSD therefore provides a useful design principle for future defect-detection and recognition models: representation, supervision, and geometric alignment should be optimized jointly when the desired output is a recoverable boundary rather than a coarse support region.

    \item The Fourier-order analysis shows that harmonic order is a practical control variable linking geometric fidelity, descriptor compactness, and device-side cost. Low-to-medium harmonic orders recover most of the global contour structure, whereas higher-order components mainly refine local curvature and small-scale boundary irregularities. This behavior is reflected quantitatively in the saturation of polygon-space accuracy and qualitatively in the gradual convergence of reconstructed contours toward the reference boundary. In the on-device inference test, reducing the order from $n_f=16$ to $n_f=8$ substantially reduces latency on the tested Atlas 200I DK A2 platform while preserving most of the polygon-space accuracy. This result should be interpreted as an implementation- and hardware-specific trade-off rather than a general claim about all edge devices. Nevertheless, it shows that Fourier order provides an explicit and interpretable mechanism for balancing accuracy, compactness, and runtime. It also motivates adaptive-order formulations, in which compact or near-regular defects may use fewer active harmonics, whereas thin, elongated, or locally irregular defects may require a larger harmonic budget.

    \item The post-prediction data-handling experiments show that FS-FSD can produce a compact and recoverable image-space defect record under the tested proof-of-concept conditions. The SQLite archive-and-recovery benchmark shows that Fourier descriptors require much smaller prediction records than raster-derived routes and can be converted back to image-space polygons with low recovery overhead. The WebGL prototype shows that the same archived descriptor can be reopened, reconstructed, rendered, and inspected interactively in a browser without storing dense masks or pre-rendered overlays in the tested setting. The on-device inference experiment further shows that FS-FSD-M with $n_f=8$ achieves competitive steady-state inference speed on the evaluated Atlas 200I DK A2 platform while maintaining stronger polygon-space accuracy than the strongest raster baseline considered here. These experiments should be read as component-level feasibility tests of the representation, not as evidence of a complete field deployment pipeline. Taken together, they indicate that FS-FSD outputs can function as compact, recoverable, and displayable image-space records while also supporting efficient on-device inference under the tested conditions. This is advantageous for post-inference handling of large numbers of predicted bridge defects.
\end{enumerate}

Several methodological boundaries and possible extensions should be considered when interpreting these results.

\begin{enumerate}
    \item The Fourier parameterization used in the current implementation is most naturally suited to simply connected closed contours. Each defect instance is mainly represented by its dominant outer boundary. Internal holes, multiple disconnected components, island-like regions, and other multi-connected topologies are not explicitly modeled. This may affect the complete representation of complex defects, such as large honeycombing or spalling regions that contain void-like interiors or unspalled islands. In such cases, a single outer-boundary descriptor can still provide a compact approximation of the dominant defect extent, but it cannot preserve all topological details. Since the experiments and metrics in this paper mainly focus on image-space recovery of the dominant instance boundary, this topological boundary does not alter the main conclusions about unified polygon-space evaluation, Fourier contour prediction, and compact image-space defect recording under the current setting. Future work could investigate topology-aware multi-contour representations, hierarchical inner-outer boundary modeling, and systematic failure-case analysis for multi-connected damage regions.

    \item In terms of polygon validity, the current framework does not include an explicit self-intersection constraint. Under the moderate Fourier orders and dataset setting used in this study, self-intersection artifacts are uncommon and are not a dominant factor affecting the reported results. Therefore, the absence of an explicit polygon-validity term does not change the main conclusions drawn under the current experimental setting. More precisely, it identifies a robustness issue that may become more relevant under higher harmonic orders, more complex boundary shapes, or more unstable predictions. Local self-intersections may occur when the harmonic order is substantially increased, when high-order coefficients become unusually large, or when the predicted boundary is highly unstable. Future extensions could incorporate self-intersection penalties, curvature regularization, spectral decay constraints, or differentiable polygon-validity checks to improve contour robustness in extreme-shape or high-order configurations.

    \item The experiments use a fixed global Fourier order for each model configuration. The results show that $n_f=16$ provides a strong overall balance in the evaluated setting, but a single order is unlikely to be optimal for every defect instance. Simple compact defects may be over-parameterized by a high-order descriptor, whereas thin, elongated, or locally irregular defects may benefit from additional harmonics. Adaptive Fourier order selection is therefore a promising extension. Such a mechanism could estimate the required harmonic budget from defect appearance, class identity, boundary uncertainty, or predicted geometric complexity, and could store only the active coefficients needed for each instance. This would further improve the trade-off among accuracy, archive size, and device-side latency.

    \item The empirical conclusions are tied to the dataset, baselines, and conversion protocol evaluated in this paper. The study includes representative box-, mask-, and contour-based models and evaluates them through a unified polygon-space protocol. However, the absolute values and relative rankings may vary with additional datasets, alternative segmentation architectures, different contour extraction settings, different polygon simplification parameters, or different defect annotation styles. The conversion rules used in this study are specified to make the comparison transparent. Future benchmark studies could further examine the sensitivity of polygon-space metrics to mask-to-polygon extraction, simplification thresholds, and different defect annotation standards.
\end{enumerate}

Finally, it is important to distinguish the methodological boundaries of FS-FSD from downstream tasks that are outside the stated scope of this paper. All predictions, evaluations, archive records, and visualization examples in this study are defined in image or normalized image coordinates, as clarified in Appendix~\ref{app:physical_mapping}. The recovered polygons and Fourier descriptors should therefore not be interpreted as physical-scale defect measurements. Converting these records into metric quantities, such as crack length in millimeters or spalling area in square centimeters, would require additional information such as Ground Sample Distance (GSD), camera intrinsics, depth or pose estimates, surface geometry, or component-level registration. Under simple scale-consistent imaging conditions, image-coordinate polygons could be scaled using GSD. Under perspective distortion or non-planar surface conditions, reconstructed polygon points would need to be mapped through calibrated geometric transformations.

Similarly, using the predicted contours in downstream inspection information workflows requires additional contextual and system-level metadata, such as structure or component identifiers, coordinate transformations, database schemas, inspection-task definitions, and record-management rules. These requirements belong to downstream information-integration layers rather than to the proposed image-space representation itself. Accordingly, the absence of physical-scale measurement or full information-system integration in this paper should be understood as a scope boundary of the present evaluation, not as a limitation of the Fourier contour descriptor as a compact image-space defect record.

Overall, the practical significance of this study lies in moving AI-generated bridge defect outputs closer to reusable inspection records. In large inspection campaigns, predicted defects are not only viewed once after inference; they often need to be stored, transmitted, reopened, checked by inspectors, and compared in later reviews. FS-FSD addresses this need by representing each predicted defect as a compact and recoverable Fourier contour record. Compared with boxes and dense masks, this representation preserves boundary geometry while reducing the burden of storage, transfer, recovery, and browser-side review. Thus, the immediate value of this work is not to claim a complete field inspection automation system, but to provide a structured image-space representation layer that makes AI-generated defect results easier to retain, revisit, and reuse.

Future research will further investigate topology-aware contour modeling, explicit geometric regularization, adaptive harmonic-order selection, broader cross-dataset validation, and calibrated mappings from image-space contour records to physically grounded inspection information when the required metadata and calibration conditions are available. With reliable scale, camera, pose, surface, and component-registration information, compact contour records could support metric defect quantification, component-level localization, cross-time defect comparison, and asset-level data organization. For the bridge inspection and maintenance sector, this direction could help transform isolated visual predictions into traceable, comparable, and accumulable defect records, providing a stronger data basis for condition assessment and maintenance planning. These future extensions would build on, rather than replace, the central premise of this study: reliable downstream use of AI-generated defect information first requires an output representation that is geometrically meaningful, compact, recoverable, and consistent with the intended evaluation and data-handling tasks.


\section*{CRediT authorship contribution statement}
\noindent \textbf{Jin Liu:} Original Draft, Visualization, Validation, Coding, Funding Acquisition. 
\textbf{Wang Wang:} Conceptualization, Original Draft, Experimental Design, Data Curation, Coding, Supervision.
\textbf{Hongxu Pu:} Validation, Data Curation.
\textbf{Zhen Cao:} Data Curation, Coding, Visualization. 
\textbf{Yasong Wang:} Data Curation, Figure Preparation. 
\textbf{Hu Wang:} Data Curation, Review.
\textbf{Kunming Luo:} Data Curation, Review, Figure Preparation.

\section*{Acknowledgement}
\noindent This work was supported by the Major Science and Technology Special Plan of Shanxi Province ("Open Bidding for Selecting the Best" Project), Key Technologies and Equipment Development for Digital Intelligence of In-service Highway Transportation Infrastructure (Grant No. 202201150401020).

\section*{Declaration of generative AI and AI-assisted technologies in the 
writing process}
\noindent During the preparation of this work the authors used GPT5.4 pro in 
order to improve language. After using this tool, the authors reviewed 
and edited the content as needed and take full responsibility for the 
content of the publication.

\section*{Declaration of competing interest}
\noindent We declare that we do not have any commercial or associative interest that represents a conflict of interest in connection with the work submitted.

\section*{Data availability}
\noindent Data will be made available on request.

\bibliographystyle{my-elsarticle}
\bibliography{rmfile}

\appendix

\section{Image-space Evaluation Scope and Optional Metric Mapping}
\label{app:physical_mapping}

This appendix clarifies the scope of the geometric evaluation used in this paper and summarizes the external information that would be required to convert archived FS-FSD contours into physical or asset-level coordinates. The main experiments are conducted in image space. Therefore, the reported polygon-space mAP, boundary metrics, perimeter error, and area error should be interpreted as representation-level geometric measures in normalized image coordinates, rather than as physical-scale engineering measurements.

FS-FSD provides a compact and recoverable image-space contour representation. It does not by itself recover camera scale, camera pose, surface depth, local Ground Sample Distance (GSD), or BIM/GIS registration. Physical-scale measurement would require additional calibrated information, such as local surface GSD, known-size references, camera calibration and camera-to-surface distance, planar homography, depth information, or asset-registration metadata. Because the dataset used in this study does not provide calibrated physical ground truth or verified GSD metadata for every defect instance, no physical-scale accuracy claim is made in this paper.

\subsection{Coordinate spaces}

Table~\ref{tab:app_coordinate_spaces} summarizes the coordinate spaces relevant to FS-FSD contour recovery and optional metric mapping.

\begin{table}[pos=htbp]
\centering
\caption{Coordinate spaces involved in image-space evaluation and optional metric mapping.}
\label{tab:app_coordinate_spaces}
\small
\renewcommand{\arraystretch}{1.18}
\begin{tabular}{p{0.22\linewidth}p{0.25\linewidth}p{0.44\linewidth}}
\hline
\textbf{Space} & \textbf{Coordinate form} & \textbf{Role} \\
\hline
Normalized image space &
$\Omega_N=[0,1]^2$ &
Used for the unified polygon-space evaluation in this paper. Distances, perimeters, and areas are dimensionless. \\
\hline
Original pixel space &
$\mathbf{p}=(u,v)^\top$ &
Used after reversing image resizing, padding, or letterbox preprocessing. Coordinates are measured in pixels. \\
\hline
Local physical surface space &
$\mathbf{s}=(X,Y)^\top$ &
Defined on a local bridge-component surface. Coordinates may be expressed in $\mathrm{mm}$, $\mathrm{cm}$, or $\mathrm{m}$ if reliable metric information is available. \\
\hline
Asset-level coordinate space &
BIM/GIS/component coordinates &
Requires additional registration to a bridge component, BIM model, GIS reference frame, or asset-management database. \\
\hline
\end{tabular}
\end{table}

Let the recovered FS-FSD contour in normalized image coordinates be
\begin{equation}
\mathbf{r}_N(t)=
\begin{bmatrix}
x_N(t)\\
y_N(t)
\end{bmatrix},
\qquad
t\in[0,2\pi).
\label{eq:app_norm_contour}
\end{equation}
For an original image with width $W$ and height $H$, the corresponding pixel-space point is
\begin{equation}
\mathbf{p}(t)=
\begin{bmatrix}
u(t)\\
v(t)
\end{bmatrix}
=
\begin{bmatrix}
W x_N(t)\\
H y_N(t)
\end{bmatrix}.
\label{eq:app_norm_to_pixel}
\end{equation}
If the image has been resized, padded, or letterboxed before inference, the inverse preprocessing transform should be applied before any optional metric mapping. For example, with scale factors $r_x,r_y$ and padding $(p_x,p_y)$,
\begin{equation}
u=\frac{u_{\mathrm{net}}-p_x}{r_x},
\qquad
v=\frac{v_{\mathrm{net}}-p_y}{r_y}.
\label{eq:app_inverse_preprocess}
\end{equation}

\subsection{External information required for physical mapping}

Optional conversion from image-space contours to local physical coordinates requires external metric information. Typical sources are summarized in Table~\ref{tab:app_metric_metadata}.

\begin{table}[pos=htbp]
\centering
\caption{External information required for optional physical or asset-level mapping.}
\label{tab:app_metric_metadata}
\small
\renewcommand{\arraystretch}{1.18}
\begin{tabular}{p{0.27\linewidth}p{0.62\linewidth}}
\hline
\textbf{Information source} & \textbf{Use and requirement} \\
\hline
Local surface GSD &
Provides the physical length represented by one image pixel on the inspected bridge surface. It must correspond to the same local surface as the defect. \\
\hline
Known-size references &
Scale bars, markers, bolt spacing, plate width, or known component dimensions can be used to estimate local scale if they lie on the same surface as the defect or after perspective correction. \\
\hline
Camera calibration and distance &
If camera intrinsics and camera-to-surface distance are known, local scale can be estimated under suitable geometric assumptions, such as approximate fronto-parallel viewing. \\
\hline
Planar homography &
For a locally planar surface observed from an oblique view, an image-to-surface homography can map pixel coordinates to local physical coordinates. \\
\hline
Depth, multi-view reconstruction, or LiDAR &
Required when the inspected region is non-planar, strongly perspective-distorted, or when three-dimensional localization is needed. \\
\hline
BIM/GIS or survey registration &
Required to associate a local physical contour with a bridge component, BIM object, GIS reference frame, or asset-management record. \\
\hline
\end{tabular}
\end{table}

When a locally planar or approximately planar surface is available, an optional affine image-to-surface mapping may be written as
\begin{equation}
\mathbf{s}(t)
=
\mathbf{s}_{\mathrm{ref}}
+
M\left(\mathbf{p}(t)-\mathbf{p}_{\mathrm{ref}}\right),
\label{eq:app_optional_affine_mapping}
\end{equation}
where $\mathbf{p}_{\mathrm{ref}}$ is a reference pixel point, $\mathbf{s}_{\mathrm{ref}}$ is its corresponding local physical coordinate, and $M$ is a metric mapping matrix determined by local scale, axis orientation, and coordinate convention. For a locally planar surface observed from an oblique view, a homography may be used instead:
\begin{equation}
\tilde{\mathbf{s}}
\sim
H_{I\rightarrow S}\tilde{\mathbf{p}} .
\label{eq:app_optional_homography}
\end{equation}
These mappings are not estimated by FS-FSD. They must be provided by external calibration, survey, photogrammetry, image registration, or asset-registration procedures.

\subsection{Archive fields for optional metric mapping}

Table~\ref{tab:app_archive_fields} lists the database fields that are sufficient for image-space recovery and those that would be needed for optional physical or asset-level mapping.

\begin{table}[pos=htbp]
\centering
\caption{Archive fields for image-space recovery and optional metric mapping.}
\label{tab:app_archive_fields}
\small
\renewcommand{\arraystretch}{1.18}
\begin{tabular}{p{0.23\linewidth}p{0.64\linewidth}}
\hline
\textbf{Field type} & \textbf{Examples} \\
\hline
Required for image-space recovery &
Image ID, defect instance ID, defect category, confidence score, Fourier coefficients, original image width $W$, original image height $H$, and preprocessing metadata if the stored coefficients are not already expressed in original-image coordinates. \\
\hline
Optional for local metric mapping &
Local GSD, known-size reference metadata, reference pixel $\mathbf{p}_{\mathrm{ref}}$, reference physical coordinate $\mathbf{s}_{\mathrm{ref}}$, affine mapping matrix $M$, or image-to-surface homography $H_{I\rightarrow S}$. \\
\hline
Optional for asset-level association &
Bridge ID, component ID, span/pier/girder/deck region, BIM/GIS registration metadata, survey control points, or digital-twin asset identifiers. \\
\hline
\end{tabular}
\end{table}

If the optional metric metadata are available, the recovered FS-FSD contour can be transformed into a local physical polygon and standard geometric quantities such as area, perimeter, centroid, orientation, and bounding extent can then be computed from that mapped polygon. These quantities depend on the accuracy of the external metric mapping and should not be attributed to FS-FSD alone.

For crack-like defects, additional care is required. FS-FSD predicts the visible boundary of the crack region, not the engineering centerline. Therefore, crack centerline length, maximum width, average width, and related crack-specific quantities generally require additional post-processing, such as skeletonization, medial-axis extraction, distance-transform-based width estimation, or centerline fitting after the image-space contour has been mapped to a calibrated physical space.

\subsection{Scope limitation}

The optional mapping described above is intended to clarify how archived image-space contours could be connected to physical or asset-level records when the required external metadata are available. It is not used as an experimental claim in this paper. A rigorous physical-scale validation would require images with verified GSD, known scale references, camera calibration, depth or pose information, and independently measured physical annotations for the defects. Constructing such a calibrated dataset is outside the scope of the present work and is left for future studies on physically grounded defect quantification.

\section{Mask-to-Polygon Conversion for Unified Polygon-space Evaluation}
\label{sec:mask_to_polygon_yolo_mask2former}

This appendix specifies the conversion rules used to evaluate mask-based baselines in the same polygon space as contour-native methods. The goal is to compare heterogeneous outputs under a common geometric protocol without changing the original category, confidence score, or model-side post-processing result.

\subsection{Common polygon construction}

All polygons used for unified evaluation are represented in normalized image coordinates:
\begin{equation}
\Omega=[0,1]\times[0,1].
\label{eq:app_eval_domain}
\end{equation}
For a contour vertex in pixel coordinates $(u_k,v_k)$, the normalized coordinate is
\begin{equation}
(x_k,y_k)
=
\left(
\frac{u_k}{W},
\frac{v_k}{H}
\right).
\label{eq:app_pixel_to_norm}
\end{equation}
For model interfaces that already provide normalized coordinates, such as the Ultralytics \texttt{xyn} fields, the coordinates are used directly.

Given the contour rings $\mathcal{X}_i$ associated with the $i$-th predicted instance, the evaluator removes consecutive duplicated vertices, closes each ring if necessary, discards rings with fewer than three non-collinear vertices, and constructs the polygonal prediction as
\begin{equation}
\hat{P}_i
=
\operatorname{MakeValid}
\left[
\left(
\bigcup_{\Gamma\in\mathcal{X}_i}
\operatorname{Polygon}(\Gamma)
\right)
\cap
\Omega
\right].
\label{eq:app_common_polygon}
\end{equation}
Shapely polygon operations are used for validity checking, clipping, union, intersection, and area computation. If validity repair returns a geometry collection, only polygonal components are retained. Empty geometries and zero-area polygons are discarded.

No Douglas--Peucker simplification, contour smoothing, morphological opening or closing, evaluator-side mask refinement, confidence recalibration, or additional non-maximum suppression is applied during this conversion. Therefore, S2P-Mask evaluation preserves the boundary-bearing output exposed by each model interface as directly as possible.

\subsection{Conversion rules for different model types}

Table~\ref{tab:app_route_assignment} summarizes the route assignment and conversion rules used in the unified polygon-space evaluation.

\begin{table}[pos=htbp]
\centering
\caption{Route assignment and conversion rules for polygon-space evaluation.}
\label{tab:app_route_assignment}
\small
\renewcommand{\arraystretch}{1.18}
\begin{tabular}{p{0.23\linewidth}p{0.18\linewidth}p{0.49\linewidth}}
\hline
\textbf{Native output type} & \textbf{Route} & \textbf{Geometry used for polygon-space evaluation} \\
\hline
Box-producing detectors, such as Faster R-CNN and RT-DETR-R50 &
R2P &
The predicted box is clipped to $\Omega$ and converted into a rectangular polygon. \\
\hline
YOLO segmentation baselines, including YOLOv5m-seg, YOLO11m-seg, and YOLO26m-seg &
S2P-Mask &
The mask-derived contours from \texttt{result.masks.xyn} are converted to polygons using Eq.~\eqref{eq:app_common_polygon}. Category labels and confidence scores are inherited from \texttt{result.boxes.cls} and \texttt{result.boxes.conf}. \\
\hline
Mask2Former-style segmentation outputs &
S2P-Mask &
Instance masks are converted to binary masks if needed and then vectorized using \texttt{cv2.findContours}. The resulting contours are normalized and converted to polygons using Eq.~\eqref{eq:app_common_polygon}. \\
\hline
Contour-native methods, including Deep Snake, FSD, and FS-FSD &
S2P-Contour &
The predicted contour or reconstructed Fourier contour is sampled into a polygon. Boundary and geometry metrics are computed from the contour-derived polygon. \\
\hline
\end{tabular}
\end{table}

For YOLO segmentation baselines, the official Ultralytics inference output is used. The S2P-Mask branch uses

\[
\texttt{result.masks.xyn},
\]
while the category and confidence score of the $i$-th instance are taken from the corresponding box fields:

\[
\texttt{result.boxes.cls},
\qquad
\texttt{result.boxes.conf}.
\]
The instance order returned by the Ultralytics interface is preserved. When high-resolution mask decoding is enabled by the model interface, the evaluator uses the contours exposed by that interface and does not intentionally downgrade the mask into a coarser representation.

For Mask2Former-style outputs, if the instance mask is already binary, it is used directly. If a probability or logit mask is returned, it is binarized with
\begin{equation}
\tau_{\mathrm{mask}}=0.5.
\label{eq:app_mask_threshold}
\end{equation}
Contours are then extracted using

\[
\texttt{cv2.findContours}
\left(
\texttt{RETR\_CCOMP},
\texttt{CHAIN\_APPROX\_NONE}
\right).
\]
The \texttt{RETR\_CCOMP} hierarchy is used to preserve interior rings as holes when holes are present. The \texttt{CHAIN\_APPROX\_NONE} mode is used to avoid evaluator-side contour simplification.

For completeness, the R2P conversion of a normalized box $\hat{B}_i=(x_1,y_1,x_2,y_2)$ is
\begin{equation}
\mathcal{C}_{\mathrm{R2P}}(\hat{B}_i)
=
\operatorname{Polygon}
\left(
[
(x_1,y_1),
(x_2,y_1),
(x_2,y_2),
(x_1,y_2)
]
\right),
\label{eq:app_r2p}
\end{equation}
after clipping the coordinates to $\Omega$ if necessary. For mask-producing models, the principal shape comparison uses S2P-Mask rather than R2P, so that YOLO-seg and Mask2Former are evaluated through their native boundary-bearing outputs rather than weakened rectangular approximations.

\subsection{Algorithmic summary and degenerate cases}

Table~\ref{tab:app_s2p_algorithm} summarizes the S2P-Mask conversion procedure.

\begin{table}[pos=htbp]
\centering
\caption{Algorithmic summary of S2P-Mask conversion.}
\label{tab:app_s2p_algorithm}
\small
\renewcommand{\arraystretch}{1.18}
\begin{tabular}{p{0.08\linewidth}p{0.82\linewidth}}
\hline
\textbf{Step} & \textbf{Operation} \\
\hline
1 & Read the native mask prediction, category label, and confidence score from the model output. \\
2 & Obtain contour rings directly from the model interface, or extract them from the binary instance mask. \\
3 & Normalize contour vertices to $[0,1]^2$ if they are provided in pixel coordinates. \\
4 & Remove duplicated vertices, close rings, and discard invalid rings with fewer than three non-collinear vertices. \\
5 & Construct polygons, clip them to $\Omega$, repair validity, and retain only polygonal components. \\
6 & Attach the original category and confidence score to the converted polygon and use it for S2P-Mask evaluation. \\
\hline
\end{tabular}
\end{table}

A converted prediction is discarded from S2P-Mask evaluation if no valid polygon can be obtained. This includes cases where no mask or contour is returned, fewer than three non-collinear vertices remain, the clipped geometry is empty, the repaired geometry has zero area, or no polygonal component remains after validity repair.

If a valid prediction is represented as a \texttt{MultiPolygon}, polygon IoU and polygon-space mAP are computed using the full polygonal geometry. For boundary or geometry metrics that require a single closed exterior boundary, the exterior boundary of the largest-area polygonal component is used. This rule is applied consistently to all S2P-Mask predictions and is used only to make single-boundary metrics well defined.

The evaluator does not perform additional score calibration, class relabeling, model-specific confidence filtering beyond the globally specified evaluation threshold, or additional non-maximum suppression. Consequently, differences in the reported polygon-space metrics reflect differences in the native geometric outputs of the compared models rather than evaluator-side simplification or degradation.

\end{document}